%% file: acl_latex.tex
\documentclass[11pt]{article}

\usepackage[final]{acl}

\usepackage{times}
\usepackage{latexsym}
\usepackage{multirow}
\usepackage{graphicx}
\usepackage{rotating}
\usepackage{arydshln}
\usepackage{float}
\usepackage{placeins}
\usepackage{array}
\usepackage{amsmath} 
\usepackage{booktabs}
\usepackage{siunitx}    
\usepackage{float}
\usepackage[normalem]{ulem} 
\usepackage{amssymb}
\usepackage{makecell}
\usepackage{longtable}
\usepackage{listings}
\usepackage{comment}
\usepackage{listingsutf8}
\useunder{\uline}{\ul}{}

\usepackage{tcolorbox}
\tcbuselibrary{listings, breakable}

\usepackage{booktabs}

\usepackage[table,xcdraw]{xcolor}

\usepackage[T1]{fontenc}

\usepackage[utf8]{inputenc}

\usepackage{microtype}

\usepackage{inconsolata}

\usepackage{graphicx}

\title{Trade-offs in Medical LLM Adaptation: An Empirical Study in French QA}

\author{
  \textbf{Ikram Belmadani\textsuperscript{1,2}}
  \textbf{Oumaima El Khettari\textsuperscript{2}} 
  \textbf{Carlos Ramisch\textsuperscript{1}} \\
  \textbf{Frederic Bechet\textsuperscript{1}}
  \textbf{Richard Dufour\textsuperscript{2}}
  \textbf{Benoit Favre\textsuperscript{1,3}}
\\
\\
  \textsuperscript{1}Aix-Marseille Univ., CNRS, LIS UMR 7020, 13000 Marseille, France, \\
  \textsuperscript{2}Nantes Univ., École Centrale Nantes, CNRS, LS2N UMR 6004, 44000 Nantes, France,\\
  \textsuperscript{3}Grenoble Alpes Univ., CNRS, INRIA, Grenoble INP, LIG UMR 5217, 38000 Grenoble, France
\\
  \small{
    \textbf{Correspondence:} \href{mailto:email@domain}{{first.last}@{univ-amu.fr, univ-nantes.fr}}
  }
}

\usepackage{xcolor} 

\newcommand{\CPTtext}{\textcolor{blue}{CPT }}
\newcommand{\SFTtext}{\textcolor{green!60!black}{SFT }}
\newcommand{\CPTSFTtext}{\textcolor{red}{CPT+SFT }}

\begin{document} 
\maketitle
\begin{abstract}
The development of large language models (LLMs) has led to an increased focus on their adaptation to specialized domains and languages, yet the effectiveness of domain adaptation strategies remains unclear. We present a study of medical domain adaptation using French medical question-answering (QA) as a case study. We compare continual pretraining (CPT), supervised fine-tuning (SFT), and their combination across three model families, multiple sizes, and three initialization types, explicitly disentangling adaptation effects from base model choice.
We evaluate both multiple-choice (MCQA) and open-ended QA (OEQA) under greedy and constrained decoding using automatic metrics and LLM-as-a-Judge evaluation. For MCQA, CPT+SFT most often achieves the best scores, but gains over SFT are small and frequently not statistically significant, making SFT a strong and cost-effective default. For OEQA, CPT consistently improves overlap-based metrics, while SFT often degrades generation quality; instruction tuning and CPT+SFT are preferred by LLM-based evaluation. Cross-lingual experiments further show effective transfer from French adaptation to English benchmarks. Overall, we provide practical guidelines for selecting adaptation strategies under computational constraints.

\end{abstract}
\begin{figure*}[!t]
\centering
\includegraphics[width=0.9\textwidth]{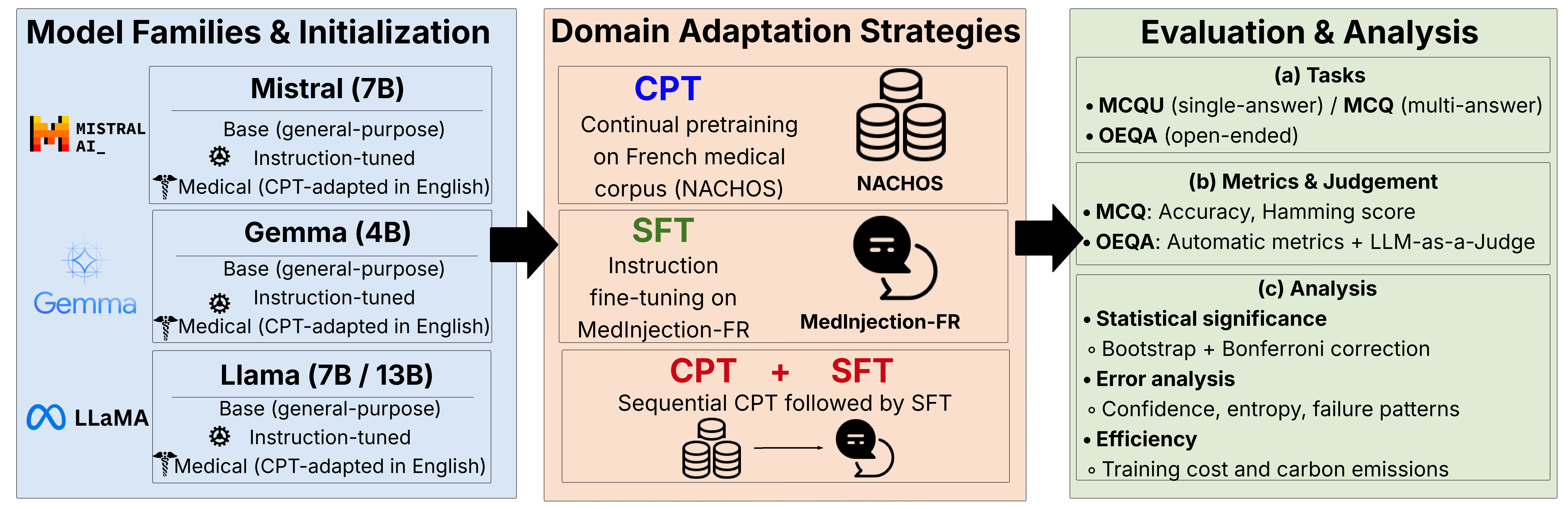}
\caption{Overview of the experimental pipeline for evaluating medical domain adaptation strategies.}
\label{fig:overview}
\end{figure*}

\section{Introduction}
LLMs are increasingly applied to medical question-answering (QA) and clinical reasoning, where accuracy, robustness, and domain-specific knowledge are critical~\citep{huang2024comprehensive}. However, most high-performing LLMs are trained on general-domain data, making domain adaptation essential for safe medical deployment. In practice, this adaptation relies on continual pretraining (CPT) on domain corpora and supervised fine-tuning (SFT) on task-specific data.

Despite their widespread use, the relative effectiveness of these strategies remains unclear. Their impact depends on training scale, data composition, and optimization choices~\citep{christophe-etal-2024-beyond,lu2025fine}, and even combined approaches yield inconsistent or statistically insignificant gains~\citep{jeong-etal-2024-medical}. 

Most prior work fixes base model initialization, making it difficult to disentangle adaptation effects~\citep{lu2025fine, christophe-etal-2024-beyond}. Evaluations are also predominantly English-centric, and largely focus on MCQA, limiting interpretation and generalization, especially given recent evidence of memorization in medical LLMs~\citep{li2025memorization}.

More broadly, this work is motivated by a practical constraint often overlooked in the literature. In many real-world settings, especially for non-English medical NLP, both domain-specific data and computational resources are limited. As a result, the key question is not whether one strategy can theoretically outperform another, but how to allocate limited resources effectively. In this context, practitioners face a concrete and unresolved question: \textit{given limited data and computational resources, which adaptation strategy should be prioritized?} Existing studies provide partial answers, but heterogeneous setups make it difficult to derive actionable guidelines.

To address these limitations, we conduct a controlled study of medical domain adaptation using French medical QA as a case study. We compare CPT, SFT, and their combination across model families and sizes while varying base initialization, and evaluate models in both French and English to isolate domain and cross-lingual effects.

We include OEQA as a complementary evaluation of generative behavior, but note that its assessment remains challenging; our conclusions are therefore primarily grounded in MCQA.

Our goal is to provide practical guidance on when and why CPT and SFT are effective under realistic constraints.
Our study is guided by the following research questions:

\begin{itemize}
\item \textbf{RQ1:} \textit{What are the performance and efficiency trade-offs between CPT and SFT across model families and sizes?}
\item \textbf{RQ2:} \textit{How does base model initialization influence the effectiveness of CPT and SFT for medical domain adaptation?}
\item \textbf{RQ3:} \textit{How does French medical adaptation affect cross-lingual transfer to English?}
\end{itemize}

Our contributions are: 
(i) we introduce a controlled and reproducible framework to compare medical domain adaptation strategies across model families, sizes, initialization types, and decoding settings; (ii) we provide a statistically grounded analysis of CPT and SFT for medical QA, covering performance trade-offs, error patterns, and cross-lingual transfer to English benchmarks. All resources are publicly available: {\url{https://github.com/ikram28/MedAdapt}}.

\section{Related Work}

Medical LLM adaptation primarily relies on CPT on domain-specific corpora and SFT on instruction–response data, both shown to support domain transfer~\citep{gururangan-etal-2020-dont,gema-etal-2024-parameter}. CPT has been adopted in models such as MediTron~\citep{chen2023meditron70bscalingmedicalpretraining}, BioMistral~\citep{labrak2024biomistralcollectionopensourcepretrained}, PMC-Llama~\citep{wu2023pmcLlamabuildingopensourcelanguage}, and MedGemma~\citep{sellergren2025medgemma}. However, recent analyses question the robustness and consistency of CPT gains under stricter evaluation protocols~\citep{jeong-etal-2024-medical}. In parallel, SFT-based models such as ChatDoctor~\citep{li2023chatdoctor} and MedAlpaca~\citep{han2023medalpaca} report great task-level improvements, though evaluations remain largely in English.

Medical domain adaptation is further challenged in non-English settings due to limited domain-specific resources. Several multilingual medical LLMs have been proposed, including Medical mT5~\citep{garcíaferrero2024medicalmt5opensourcemultilingual}, BiMediX~\citep{Pieri_2024}, Apollo~\citep{wang2024apollo}, and MMedLM~\citep{qiu2024buildingmultilinguallanguagemodel}. However, these models are mostly evaluated on translated benchmarks, with limited validation on native-language medical tasks, leaving their language- and cultural-specificities underexplored.
Evaluation practices also pose challenges. Widely used benchmarks such as PubMedQA~\citep{jin2019pubmedqa}, MedQA~\citep{jin2019pubmedqa}, and MedMCQA~\citep{pal2022medmcqa} primarily target English.

Beyond proposing individual models, recent work has compared adaptation strategies in controlled settings.~\citet{christophe-etal-2024-beyond} analyze CPT, SFT, and related techniques for clinical LLMs, finding that CPT alone yields limited gains but can amplify performance when combined with instruction tuning. Similarly,~\citet{lu2025fine} study CPT, SFT, and preference-based optimization across domains, highlighting complex interactions between adaptation methods. However, these studies focus on English and fix the base model initialization. In contrast, in this work, we systematically compare CPT, SFT, and their combination across multiple model families and initialization points for French medical QA, while also evaluating cross-lingual performance and analyzing adaptation behavior.

\section{Experimental Framework}
\label{sec:exp-setup}

We propose a controlled experimental framework to evaluate medical domain adaptation strategies across architectures, initialization points, and task formats, as illustrated in Figure~\ref{fig:overview}. Our setup explicitly varies (i) the base model and its prior training, (ii) the adaptation strategy, and (iii) the evaluation task and language in order to isolate the factors that drive adaptation effectiveness.

\subsection{Base Models and Adaptation Approaches}

Our study focuses on model families with three complementary initialization states: (i) a general-purpose base model, (ii) an instruction-tuned variant, and (iii) a medically adapted version obtained via CPT. This constraint is central to our experimental design, as it enables a controlled comparison that isolates the effect of adaptation strategy from that of the starting point. As a result, model selection is restricted to families providing these aligned variants, rather than to the most recent model releases.

We consider three model families spanning different sizes, pretraining regimes, and linguistic exposure. Specifically, we include Mistral-7B, Gemma-4B, and Llama models at the 7B and 13B scales. For Mistral-7B, we use Mistral-7B-v0.1 and its instruction-tuned version, and BioMistral-7B, a model adapted to the biomedical domain via CPT~\citep{jiang2023mistral7b,labrak-etal-2024-biomistral}. For Gemma, we rely on the Gemma-3-4B pretrained and instruction-tuned models, together with MedGemma-3-4B, which incorporates medical pretraining~\citep{gemmateam2025gemma3technicalreport,sellergren2025medgemma}. Finally, for Llama, we include both 7B and 13B variants, using the base and chat versions of Llama-2, as well as their medically adapted counterparts, MediTron-7B and MedLlama-13B~\citep{touvron2023Llama,chen2023meditron,wu2024pmc}.

These families differ not only in scale but also in pretraining data and exposure to French. Mistral and Gemma are explicitly multilingual, whereas Llama models are primarily English-centric, although exact language proportions are not disclosed. Except for MedGemma, whose medical pretraining corpus is not fully documented, all medical variants rely on PubMed Central as their primary biomedical source\footnote{\url{https://pmc.ncbi.nlm.nih.gov/}}. 

Across all model families and initialization points, we investigate three adaptation strategies: (i) \CPTtext on domain-specific corpora, (ii) \SFTtext on instruction-response pairs, and (iii) a sequential \CPTSFTtext pipeline.

\subsection{Training Data}
\label{sec:trainingdata}

\paragraph{CPT.} We use NACHOS corpus~\citep{labrak2023drbertrobustpretrainedmodel}, an open-source French medical dataset comprising 4\,GB of text collected from French medical websites; full details are provided in Appendix~\ref{app:nachos}.

\paragraph{SFT.} We use the train and validation sets of the MedInjection-FR corpus~\citep{belmadani-etal-2026-medinjection}, which contains 543\,505 instruction-response pairs. The dataset includes multiple-choice questions with a single \textbf{unique} correct answer (MCQU, $\sim$83\%), \textbf{multiple} correct answers (MCQ, $\sim$6\%), and OEQAs ($\sim$11\%). This mixture allows us to evaluate adaptation effects across both discriminative and generative medical reasoning tasks. Additional dataset details are provided in Appendix~\ref{app:medinjection}.

\subsection{Training Process}  

To explore the trade-off between computational cost and model plasticity, we adopt contrasting fine-tuning regimes for CPT and SFT. CPT is performed using full-parameter fine-tuning, while SFT relies on parameter-efficient adaptation. This choice is supported by preliminary experiments, as explained in Appendix~\ref{app:peft_vs_full}.

\paragraph{CPT.} CPT is performed for three epochs following the setup of~\citet{labrak-etal-2024-biomistral}. Full hyperparameter details are provided in Appendix~\ref{app:CPT-hyperparams}.

\paragraph{SFT.} We employ DoRA (Weight-Decomposed Low-Rank Adaptation)~\citep{mao-etal-2024-dora}, an extension of LoRA~\citep{hu2022lora} that decouples magnitude and directional updates. We select DoRA after preliminary experiments, as detailed in Appendix~\ref{app:peft_vs_full}. SFT is run for ten epochs, with hyperparameters reported in Appendix~\ref{app:SFT-hyperparams}.

\subsection{Evaluation Protocol}

\paragraph{Benchmarks.}
We evaluate all models on MedInjection-FR test set, which consists of 14\,533 native French medical examples and 13\,293 translated examples derived from established English benchmarks. The test set covers MCQU, MCQ, and OEQA tasks, enabling evaluation of both answer selection and free-form answers. Benchmark sources and translation procedure are detailed in Appendix~\ref{app:benchmarks}.

\paragraph{Prompting Strategy.}
All evaluations are conducted in a zero-shot setting using greedy, deterministic decoding to ensure reproducibility. For MCQU and MCQ tasks, following~\citet{liang2022holistic,beeching2023open,chen2023meditron}, we restrict the output vocabulary to valid answer options to prevent hallucinated responses. To mitigate position bias, we randomly shuffle answer choices three times and report aggregated results, following best practices for MCQ evaluation~\citep{pezeshkpour-hruschka-2024-large}. Prompt templates are provided in Appendix~\ref{app:prompts}.

\paragraph{Evaluation Metrics.}
For MCQU, we report \textit{Exact Match (EM)}, which measures the proportion of questions for which the predicted answer exactly matches the gold answer. For MCQ, we additionally report the \textit{Hamming score}, which accounts for partial overlap between predicted and reference answer sets and is therefore more informative for multi-answer questions. Formal definitions of both metrics are provided in Appendix~\ref{app:metrics}.

For OEQA, we rely on both automatic text-based metrics and model-based judgments. We report BLEU~\citep{papineni-etal-2002-bleu}, ROUGE~\citep{lin-2004-rouge}, METEOR~\citep{banerjee-lavie-2005-meteor}, and BERTScore~\citep{zhang2019bertscore} as automatic baselines. Their reliability was assessed through agreement with senior physician annotations on a held-out subset of 500 OEQA instances in \citet{belmadani-etal-2026-judges}, where MedGemma-27B was identified as the most stable and best-performing LLM judge, and is therefore used in the present work.

\paragraph{Statistical Significance and Error Analysis.}
We assess statistical significance using a percentile bootstrap procedure with 10\,000 resamples, following~\citet{jeong2024limitedimpactmedicaladaptation}. Differences between paired model configurations are considered significant when the associated two-sided $p$-value is below a predefined threshold $\alpha$. To control for multiple comparisons, we apply the Bonferroni correction, yielding a corrected $\alpha$ as specified in Appendix~\ref{app:significance}. In addition, we conduct an error analysis by examining output probabilities, confidence scores, and entropy, enabling us to characterize how CPT and SFT affect uncertainty and error patterns across different base model initializations.

\section{Results and Discussion}
\input{main-results-bis}

\subsection{MCQA Evaluation}

Table~\ref{tab:main-results} reports performance on MCQA across three model families (Gemma-4B, Mistral-7B, Llama-7B-13B), three initialization types (General, Instruct, Medical), and three adaptation strategies (CPT, SFT, CPT+SFT). Results are shown for both MCQs and MCQUs, using EM and Hamming scores for MCQs. All results reported in this table are obtained using constrained decoding. The corresponding results under greedy decoding are provided in Appendix~\ref{app:greedy-results}.


\paragraph{Effectiveness of Adaptation Strategy:}

A recurring pattern observed throughout the results is:
\vspace{-0.7em}
\[ \textsc{Base} \ll \textsc{\CPTtext} < \textsc{\SFTtext} \lesssim \textsc{\CPTSFTtext} \]
The strongest performance is most frequently achieved by the \textsc{CPT+SFT} adaptation. Across model families and initialization types, \textsc{CPT+SFT} yields the highest scores in aggregated EM as well as in MCQ and MCQU EM more often than any other strategy. 

However, a closer inspection of the results indicates that the gains brought by \textsc{CPT+SFT} over \textsc{SFT} alone are generally limited. When \textsc{CPT+SFT} attains the highest score, the margin over \textsc{SFT} rarely exceeds 1.3 points. In contrast, in configurations where \textsc{SFT} outperforms \textsc{CPT+SFT}, the performance gap is larger. For example, on Llama-7B Instruct, \textsc{SFT} exceeds \textsc{CPT+SFT} by 3.12 points, and a similar pattern is observed for Mistral-7B Instruct, with a gap of 1.44 points in favor of \textsc{SFT}.

Furthermore, the statistical analysis reported in Appendix~\ref{app:significance} shows that, when comparing each adapted model to its corresponding base model, the observed improvements are not always statistically significant. In particular, for Gemma Instruct, neither \textsc{SFT} nor \textsc{CPT+SFT} yields statistically significant gains over the base model. Likewise, for Llama-7B Instruct, the improvement brought by \textsc{CPT+SFT} is not statistically significant. These constitute the only cases in which the improvements of \textsc{SFT} or \textsc{CPT+SFT} over the base model fail to reach statistical significance.
Consequently, although \textsc{CPT+SFT} most frequently ranks first, its advantage over \textsc{SFT} is not consistently substantial.

By contrast, \textsc{CPT} alone exhibits less stable behavior. Although it can improve performance in some rare cases, it can also occasionally degrade performance compared to the base model. Moreover, it is the strategy that most often fails to produce statistically significant improvements over the base. This is the case for 8 models: MedGemma, all Llama-13B variants, Llama-7B GENERAL and INSTRUCT, and Mistral-7B GENERAL and MEDICAL. This suggests that representation-level domain adaptation is most effective when paired with task-specific supervision.


Overall, while \textsc{CPT+SFT} ranks first most often, its limited and inconsistent gains over \textsc{SFT}, together with a substantially higher computational cost (see Appendix~\ref{app:conso}), make \textsc{SFT} a strong default for medical MCQA. For example, on 7B models, \textsc{CPT+SFT} costs over \$1\,500 versus \$360 for \textsc{SFT}, with a fourfold increase in carbon emissions.

\paragraph{Impact of Model Initialization:} The impact of model initialization (General / Instruct / Medical) varies across MCQA metrics and question formats. Considering the overall best scores across all model families, instruction-tuned models dominate the most demanding EM settings: the highest MCQ EM and aggregated MCQA EM scores are both achieved by Llama-13B Instruct, while the best MCQ Hamming score is obtained by Gemma-4B Instruct. In contrast, the best MCQU EM score is achieved by a general Mistral model. 

At the family level, the patterns differ. For MCQ EM, the best score within each model family is always obtained by an instruction-tuned variant, confirming that instruction alignment is particularly beneficial for exact multi-label prediction; this advantage is further supported by statistically significant gains when compared to general or medical initializations (see Appendix~\ref{app:significance}). For MCQ Hamming, results are more balanced, with the best scores split across initialization types (two instruction-tuned, one general, and one medical).

For MCQU EM, general models most frequently achieve the best performance (three cases), followed by medical models, while instruction-tuned models do not dominate. 
This indicates that when only a single answer must be selected, performance is driven primarily by answer plausibility ranking, favoring strong language modeling and domain knowledge, while explicit instruction alignment, which mainly benefits structured or multi-label outputs, provides less advantage.
Finally, for the aggregated MCQA score, no single initialization consistently dominates: instruction-tuned and general models each obtain the best result in two configurations (with ties between them), while medical models lead in one case, and differences across initializations are often not statistically significant.

\subsection{OEQA Evaluation}

The right side of Table~\ref{tab:main-results} reports OEQA across model families using ROUGE-L, BERTScore, and LLM-as-a-Judge. We additionally report BLEU and METEOR in Appendix~\ref{app:greedy-results}, as they reflect similar information to ROUGE-L. Overall, absolute scores remain moderate, reflecting the difficulty of evaluating free-form answer generation. ROUGE-L scores should be interpreted with caution, as they measure surface-level lexical overlap and penalize semantically correct answers that differ in formulation~\citep{yim2025morqa,zhu2025judgelm}.

Moreover, OEQA represents only 11\% of the training data, resulting in a strongly imbalanced supervision signal. Models are therefore adapted to generate short, structured outputs (answer letters in MCQA), which limits OEQA performance. 

\paragraph{Effect of Adaptation Strategy:} Across model families, SFT often degrades ROUGE-L and BERTScore-F1 compared to base or CPT-adapted models, particularly for instruction-tuned and medical variants. This suggests that SFT can overly constrain generation, reducing lexical diversity and semantic overlap in an open-ended setting.

By contrast, CPT is the most consistently beneficial strategy for OEQA. CPT improves ROUGE-L and BERTScore-F1 across most general, instruct, and medical models, with especially strong gains for Mistral and Llama families. These results suggest that domain-adaptive language modeling supports better medical generation than instruction-level supervision alone. Combining CPT with SFT rarely outperforms CPT alone and often leads to intermediate or degraded performance, reflecting the same instability observed in MCQA, but with more pronounced negative effects in OEQA. 

In contrast to overlap-based metrics, LLM-as-a-Judge favors \textsc{CPT+SFT} in half of the configurations (6/12), compared to three cases each for the base and \textsc{CPT} models. Gains are most pronounced for Llama-7B, where \textsc{CPT+SFT} consistently outperforms \textsc{SFT} across initializations, and for medical models, where it yields the best or near-best qualitative scores. However, despite these trends, statistically significant improvements over the base model are rare: CPT is significant in only three cases, SFT in two (all involving smaller 4B models), and \textsc{CPT+SFT} never yields statistically significant gains over the base model in OEQA.

\paragraph{Effect of Model Initialization.} Initialization effects on OEQA depend strongly on the evaluation metric. For overlap-based metrics, no initialization consistently dominates: ROUGE-L is split between general and instruction-tuned models, while medical models never achieve the top score; BERTScore-F1 is mostly dominated by instruction-tuned models, with a single exception (Gemma).

LLM-as-a-Judge reveals clearer and statistically grounded patterns. When differences are significant, medical models are consistently outperformed across families, particularly under SFT and CPT (Table~\ref{tab:stat-oeqa}). Comparisons between instruction-tuned and general models are mixed and direction-dependent, with some significant gains under SFT and CPT (notably for Mistral and Llama), but these effects largely disappear under CPT+SFT. 

Overall, medical initialization alone does not improve OEQA, while instruction-tuned initialization yields more reliable, yet limited, gains when significant.



\section{Cross-Lingual Transfer After French Medical Adaptation}
\label{sec:en-vs-fr}
\begin{figure*}[!t]
\centering
\includegraphics[width=0.9\textwidth]{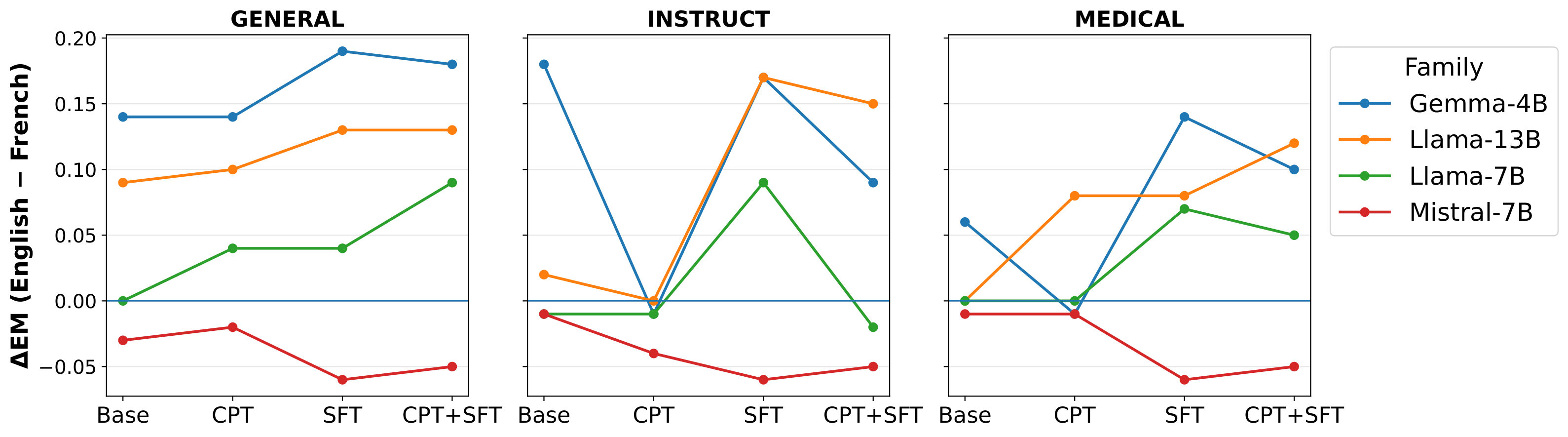}
\caption{Difference in EM accuracy ($\Delta EM$) between native English MCQU test benchmarks and their French translations across model families and adaptation strategies (constrained decoding).}
\label{fig:en-vs-fr}
\end{figure*}
To analyze whether models perform better in English prior to adaptation and how cross-lingual adaptation affects performance, we compute the EM accuracy difference for MCQU benchmarks as the score on French translations minus the score on the corresponding native English datasets. Figure~\ref{fig:en-vs-fr} reports averaged results across datasets using constrained decoding; full results for both greedy and constrained decoding are provided in Appendix~\ref{app:en-vs-fr}.

For the Mistral family, base models consistently perform better on the translated French benchmarks than on the original English data. This trend persists after adaptation with CPT, SFT, and CPT+SFT, with French performance systematically exceeding English, the differences being statistically significant (Table~\ref{tab:stat-en-vs-fr}).

In contrast, Gemma and Llama families show higher performance on native English benchmarks at the base level, and this advantage remains after adaptation on French data. Moreover, adaptation gains are often larger in English than in French (Table~\ref{tab:en-vs-fr}), despite all adaptation data being in French.

These results suggest that Mistral models encode French more effectively, whereas Gemma and Llama have stronger English representations. Notably, the improvements observed in both languages indicate effective cross-lingual transfer of medical knowledge: adapting with French medical data improves performance on the original English benchmarks, sometimes more than on their French translations. This supports the complementarity of multilingual medical data, in line with~\citet{wang2024apollo}.

A salient exception is Llama-7B: before adaptation, the base model shows slightly higher performance on French translations than on English, but this difference is not statistically significant (Table~\ref{tab:stat-en-vs-fr}). After adaptation, English performance surpasses French, suggesting that adaptation amplifies the model’s dominant English representations.

\section{Effect of Translated Benchmarks on Performance and Confidence}
\begin{figure}[!t]
\centering
\includegraphics[width=0.9\columnwidth]{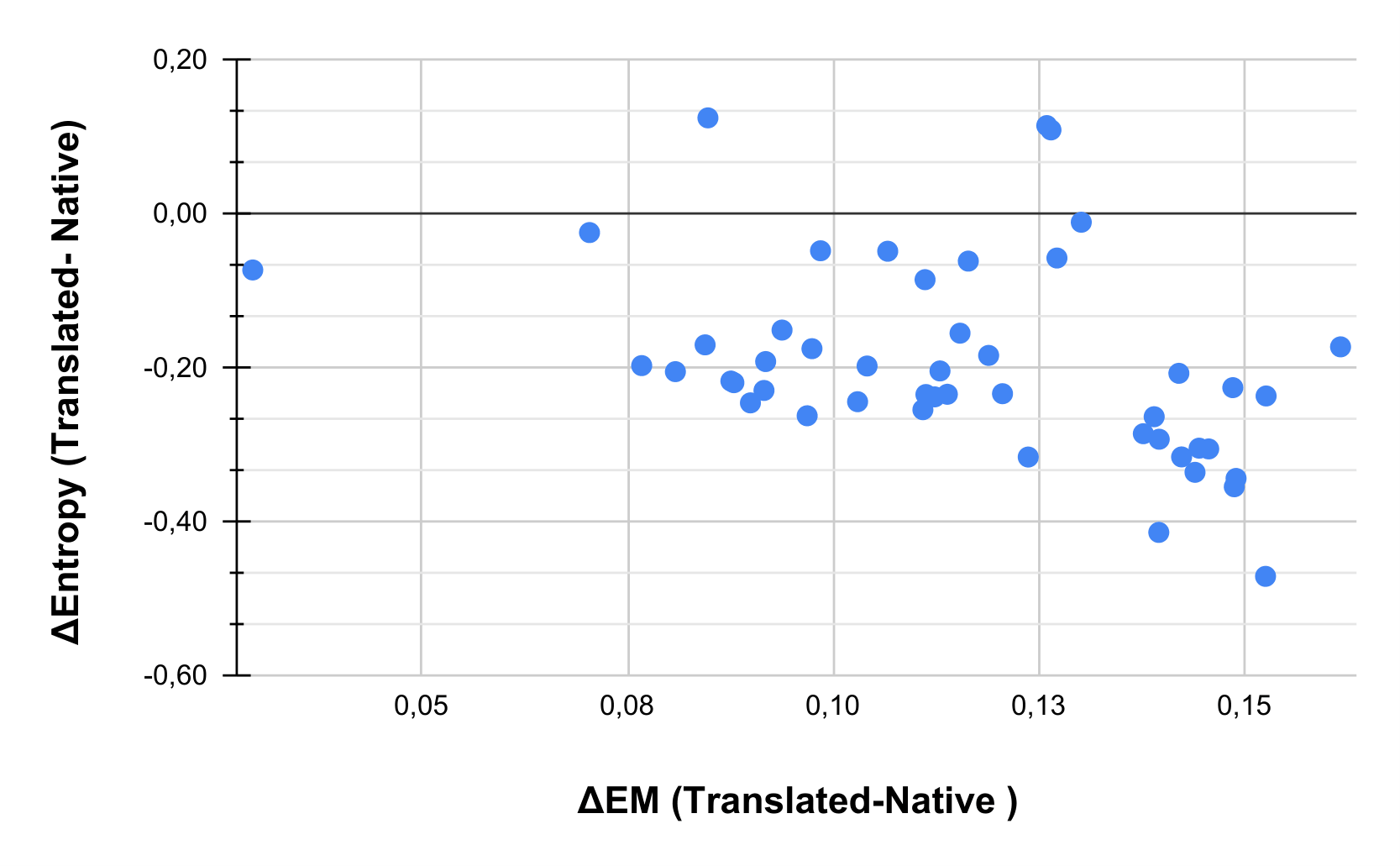}
\caption{Relationship between accuracy gain ($\Delta EM$) and change in predictive entropy ($\Delta Entropy$) when moving from the translated to native benchmarks. Each point corresponds to a model configuration.}
\label{fig:nat-vs-trad-entropy}
\end{figure}
We compare model behavior on a native benchmark, MediQAl~\citep{bazoge2025mediqal}, and a translated benchmark, MedMCQA~\citep{pal2022medmcqa}, using accuracy and confidence-based metrics. Both benchmarks consist of MCQUs of comparable size, for fair comparison. Although instances are not shared, consistent differences across models are observed.

As shown in Figure~\ref{fig:nat-vs-trad-entropy}, all models achieve higher EM accuracy on the translated benchmark. This gain is systematically accompanied by a reduction in predictive entropy, indicating that translated benchmarks induce more confident and less uncertain predictions. The concentration of models in the bottom-right quadrant suggests that translated benchmarks operate in a different evaluation regime, characterized by both higher performance and reduced uncertainty.

Figure~\ref{fig:nat-vs-tad-pmax} further reveals that accuracy gains are often associated with increased confidence in incorrect predictions. Most models exhibit positive shifts in confidence even when wrong, indicating a systematic overconfidence effect induced by the translated benchmark.

Overall, these results show that translated benchmarks are not neutral substitutes for native ones: they tend to inflate performance while also altering model confidence calibration, potentially leading to over-optimistic evaluations.

\section{Error Analysis}
\subsection{Probability-Level Analysis of MCQA}

\begin{figure}[!t]
    \centering
    \includegraphics[width=0.9\columnwidth]{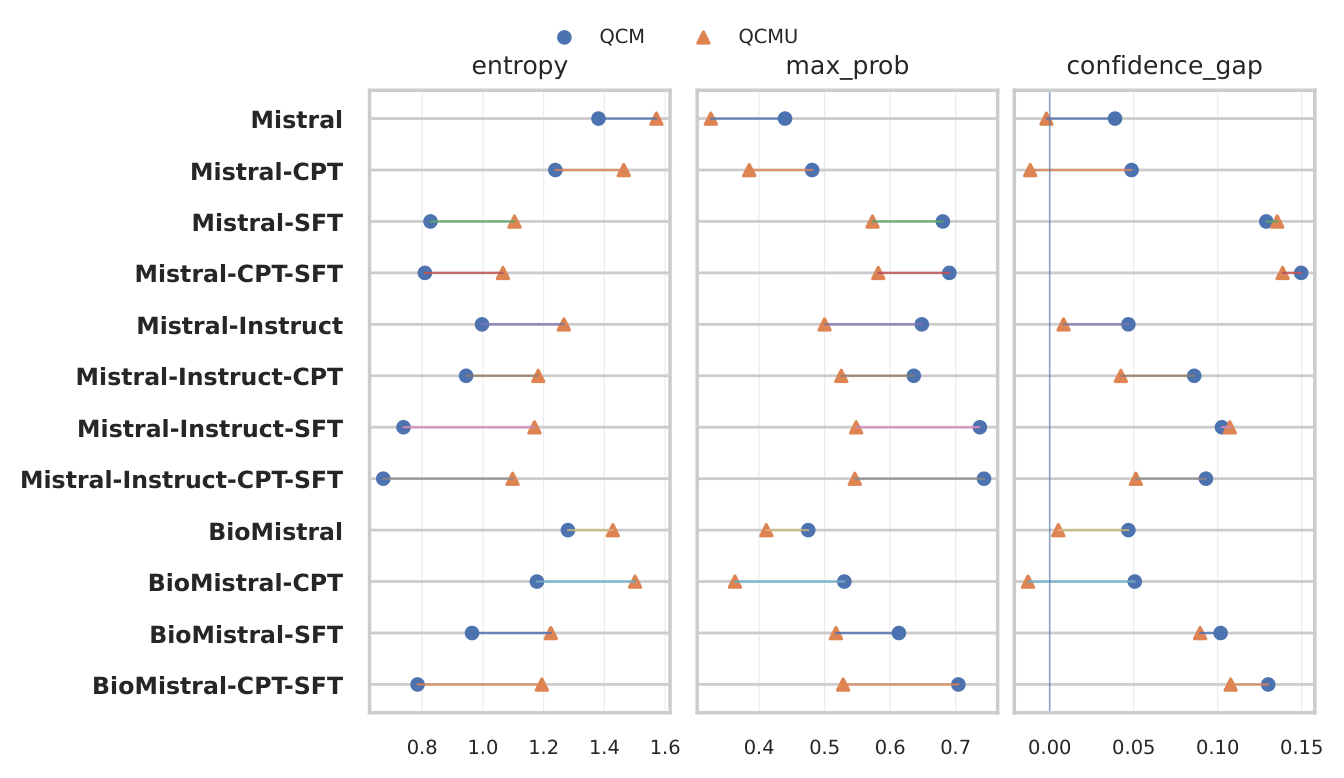}
    \caption{Probability-level metrics for MCQ and MCQU across Mistral variants.}
    \label{fig:mcqa-error-analysis}
\end{figure}

To explain why MCQ is harder than MCQU, we analyze class probability distributions from the Mistral family, selected for its high variance across models and adaptation settings. For each item, we compute entropy, maximum probability, and a confidence gap measuring gold/non-gold separation (mean gold vs. non-gold probability in MCQ; margin to the second-best option in MCQU). We also report a near-miss rate, defined as cases where all gold answers are ranked in the top-k but the predicted set is incorrect (Figure~\ref{fig:mcqa-error-analysis}, Appendix~\ref{app:nearmiss}).

Across all variants, MCQ predictions are not more uncertain than MCQU: MCQ exhibits lower entropy and higher maximum probability, indicating confident local rankings. The confidence gap is consistently positive and increases with adaptation, but remains insufficient for exact multi-label generation under greedy decoding, leading to omissions or over-generation.

Adaptation clarifies this effect: SFT strongly improves MCQU, while gains on MCQ remain limited. CPT+SFT primarily increases ranking confidence rather than exact set match, yielding larger confidence gaps without reducing near-miss rates.
\subsection{Verbosity Bias in OEQA}

To better understand the differences observed between overlap-based metrics and LLM-as-a-Judge evaluations in OEQA, we analyze the length of generated answers across models. The results are reported in Appendix~\ref{app:oeqa-verbosity}. We find that CPT-adapted models systematically produce longer responses, with higher mean and median word counts across all model families. This increased verbosity provides a plausible explanation for their strong performance on ROUGE-L and BERTScore-F1, which reward lexical recall and content coverage.

In contrast, instruction-tuned models generate substantially shorter and more controlled answers, particularly under SFT, often producing concise responses with low variance. While this behavior negatively impacts overlap-based metrics, it aligns with higher LLM-as-a-Judge scores, suggesting that concise answers are preferred under LLM evaluation. Finally, SFT exhibits unstable behavior in OEQA, leading either to excessively short outputs or overly long responses depending on model initialization. Overall, these results indicate that OEQA performance is strongly influenced by length biases, and that improvements in automatic metrics may partially reflect increased verbosity rather than improved answer quality.

\section{Conclusion}

We presented a controlled and statistically grounded study of medical domain adaptation for LLMs using French medical QA, isolating the effects of model initialization, adaptation strategy, decoding, and evaluation. Our results show that adaptation effectiveness is task-dependent and that stronger strategies are not always more cost-effective. We therefore distill practical guidelines for selecting adaptation strategies based on data availability and computational constraints. Given the limited reliability of current OEQA metrics and the small proportion of OEQA supervision, our recommendations primarily emphasize MCQA, with OEQA trends interpreted cautiously.

\paragraph{Unlabeled data only.}
When only unlabeled medical text is available, CPT yields modest and unstable gains for MCQA and should not be used in isolation. Its benefits mainly appear on OEQA overlap-based metrics, which are sensitive to verbosity and should be interpreted with caution.

\paragraph{Labeled data only.}
With labeled QA data, SFT provides the best performance–efficiency trade-off for MCQA across all model families. It frequently matches or exceeds CPT+SFT while requiring substantially fewer computational resources, making it the most practical default in this setting.

\paragraph{Labeled and unlabeled data.}
When both data types are available, CPT+SFT most often achieves the highest MCQA scores, but improvements over SFT are typically small and not consistently statistically significant. Consequently, CPT+SFT is justified only when maximal performance outweighs computational cost.

\paragraph{Initialization and compute considerations.}
Instruction-tuned models constitute the strongest baseline for French medical MCQA. Medical initialization alone does not reliably improve downstream performance. From a resource perspective, parameter-efficient SFT is by far the most cost-effective strategy, whereas CPT incurs high computational and environmental costs for limited MCQA gains, and CPT+SFT compounds these costs for marginal improvements.

\paragraph{Evaluation and transfer considerations.}
Finally, we observe strong evaluation effects: adaptation on French medical data transfers to English benchmarks, translated datasets inflate both accuracy and confidence, and OEQA metrics are sensitive to verbosity. These findings highlight the need for task-aware adaptation choices and cautious metric interpretation in medical LLM evaluation.

\section{Limitations}
Our evaluation of adaptation strategies faces several limitations. First, we perform an exploratory contamination study to assess possible exposure to \textsc{NACHOS} during pretraining (Appendix~\ref{app:contamination}). Although no direct evidence of memorization is observed, likelihood-based tests remain inconclusive due to the lack of a reliable non-member biomedical control corpus, requiring the use of synthetic controls. We therefore treat these results as indicative only and avoid causal conclusions about pretraining inclusion.

 Second, our evaluation of OEQA relies on overlap-based metrics, BERTScore, and LLM-as-a-Judge. While these measures capture complementary aspects of answer quality, they do not fully characterize semantic equivalence, clinical correctness, or reasoning validity, and may therefore overlook qualitative differences between correct answers~\citep{yim2025morqa,zhu2025judgelm}.

Third, while we demonstrate the efficiency of SFT compared to CPT in terms of computational resources, our analysis does not account for the human effort required to create high-quality instruction-tuning datasets. This consideration is particularly relevant for low-resource settings where creating domain-specific instruction data may be costly.

Fourth, we do not include few-shot prompting as an evaluation setting. Our objective is to isolate the effects of parameter-level adaptation strategies under controlled and reproducible conditions. Few-shot prompting introduces additional sources of variance related to example selection, ordering, and prompt design, which would complicate statistical comparison and obscure the interpretation of adaptation gains. Moreover, few-shot prompting assumes access to curated task-specific examples at inference time, which may be unrealistic in medical deployment scenarios. For these reasons, we focus on zero-shot evaluation to ensure fair and stable comparisons across adaptation strategies.

Finally, our study focuses exclusively on CPT and SFT. We do not explore reinforcement learning–based adaptation strategies, such as preference optimization or reward-driven fine-tuning, which may better align models with clinical judgment or evaluation criteria. Investigating how such methods interact with CPT and SFT, particularly under multilingual and domain-specific constraints, constitutes an important direction for future work.

In addition, our findings about the effectiveness of adaptation strategies are specific to the medical domain and French language. The generalizability of these results to other domains or languages, particularly those with different resource constraints or linguistic characteristics, requires further investigation.

\section{Ethical Considerations}

This work is intended for research purposes only and not for direct clinical use. All experiments rely on publicly available biomedical datasets without identifiable patient data. We acknowledge that medical LLMs may generate inaccurate or overconfident outputs, particularly in open-ended settings, and therefore include physician-based evaluation protocols. We also report computational cost and estimated carbon emissions for transparency.

\section{Acknowledgements}
This work was financially supported by ANR MALADES (ANR-23-IAS1-0005). It was provided with computing HPC and storage resources by GENCI at IDRIS thanks to the grants 2025-AD011015256R1 and 2025-AD011016540 on the supercomputer Jean Zay’s H100 partition.

\newpage
\bibliography{custom}
\newpage

\appendix

\section{CPT Training Corpus : NACHOS Description}
\label{app:nachos}
The NACHOS corpus is a French medical open-source dataset compiled through extensive web crawling and text collection. While the full corpus spans 7.4~GB of data and contains over one billion words sourced from 24 French-speaking high-quality websites~\citep{labrak2023drbertrobustpretrainedmodel}, we use in this work its \emph{small} variant, \textsc{NACHOS$_{small}$}. This version consists of approximately 4~GB of data and was obtained by shuffling the full corpus and randomly selecting 25.3 million sentences to ensure homogeneous coverage of data sources.

\textbf{Note:} Full details of the corpus compilation and processing are available in the original paper~\citep{labrak2023drbertrobustpretrainedmodel}.

\subsection{Corpus Composition}

The NACHOS corpus encompasses a diverse range of medical textual sources, including:
\begin{itemize}
    \item Descriptions of diseases and conditions
    \item Treatment and medication information
    \item General health-related advice
    \item Official scientific meeting reports
    \item Anonymized clinical cases
    \item Scientific literature
    \item Theses
    \item French translation pairs
    \item University health courses
\end{itemize}

\subsection{Data Sources}

The corpus integrates data from multiple sources, with the most significant contributions coming from:
\begin{itemize}
    \item HAL (638,508,261 words)
    \item Haute Autorité de Santé (HAS) (113,394,539 words)
    \item Drug leaflets (74,770,229 words)
    \item Medical Websites Scraping (60,561,495 words)
    \item ANSES SAISINE (51,372,932 words)
    \item Public Drug Database (BDPM) (48,302,695 words)
\end{itemize}

\subsection{Corpus Preparation}

The researchers employed several preprocessing steps:
\begin{enumerate}
    \item Text collection through web scraping, raw textual sources, and optical character recognition (OCR)
    \item Sentence splitting using heuristic methods
    \item Aggressive filtering to remove short or low-quality sentences
    \item Language classification using a custom classifier trained on multilingual corpora
\end{enumerate}

\section{SFT Training Corpus : MedInjection-Fr Description}
\label{app:medinjection}
\subsection{Overview}
MedInjection-FR~\citep{belmadani-etal-2026-medinjection} is a large-scale French biomedical instruction dataset composed of native, translated, and synthetic instruction--response pairs. The dataset comprises 571\,436 examples spanning MCQUs, MCQs, and OEQAs.

\subsection{Data Composition}
The dataset consists of 77\,247 native examples, 417\,674 translated examples, and 76\,506 synthetic examples. All data are formatted as instruction--response pairs and normalized to a unified schema, ensuring consistency across heterogeneous sources and supervision types.

\subsection{Quality Control for Translated Data}
\label{sec:translation-quality}
The translated subset was obtained by translating English biomedical instruction datasets into French using two LLMs: \texttt{GPT-4o-mini}~\citep{hurst2024gpt} and \texttt{Gemini~2.0~Flash}~\footnote{\url{https://cloud.google.com/vertex-ai/generative-ai/docs/models/gemini/2-0-flash}}. Translation quality was evaluated on the WMT~2024 Biomedical Translation Task benchmark~\citep{neves-etal-2024-findings} using BLEU and COMET metrics. \texttt{GPT-4o-mini} achieved a BLEU score of 51.01 and a COMET score of 0.8751, while \texttt{Gemini~2.0~Flash} achieved a BLEU score of 53.72 and a COMET score of 0.8783. These results are comparable to the best-performing system reported in the shared task (BLEU 53.54, COMET 0.8760), suggesting high semantic fidelity and robust preservation of biomedical terminology in the translated subset.

\subsection{Quality Control for Synthetic Data}
The synthetic subset was generated using \texttt{GPT-4o} from source documents including clinical cases and biomedical abstracts. Each source document was used to generate multiple instructional tasks covering a broad range of biomedical reasoning, such as clinical summarization, factual QA, diagnostic reasoning, treatment suggestion, and classification.

To control generation quality, each synthetic instruction--response pair was evaluated using four independent large language models acting as automatic judges: \texttt{GPT-4.1-mini}~\footnote{\url{https://openai.com/index/gpt-4-1/}}, \texttt{Gemini~2.0~Flash}, \texttt{MedGemma-27B}~\citep{sellergren2025medgemma}, and \texttt{Qwen3-Next-80B-A3B-Instruct}~\citep{qwen3technicalreport}. For MCQAs, evaluators assigned scores on a three-point scale reflecting answer correctness and contextual coherence. For OEQAs, a five-point scale was used to capture varying degrees of factual accuracy and completeness. Only examples meeting predefined minimum quality thresholds across evaluators were retained in the final dataset.

\section{CPT hyperparameters}
\label{app:CPT-hyperparams}
\begin{table}[H]
\centering
\resizebox{\columnwidth}{!}{%
\begin{tabular}{cc}
\hline
\textbf{Parameter} & \textbf{Value} \\ \hline
Learning rate & 2e-05 (1e-04 for gemma family) \\ 
Train batch size & 2 (4 for gema family) \\ 
Seed & 42 \\ 
Gradient accumulation steps & 2 (16 for gemma family) \\ 
Optimizer & AdamW \\ 
Weight Decay & 0.01 \\ 
Scheduler & Cosine \\ 
Number of epochs & 3 \\ \hline
\end{tabular}%
}
\caption{Hyperparameters used in CPT training}
\label{cpt-params}
\end{table}

\section{SFT hyperparameters}
\label{app:SFT-hyperparams}
\begin{table}[H]
\centering
\resizebox{\columnwidth}{!}{%
\begin{tabular}{cc}
\hline
\textbf{Parameter} & \textbf{Value} \\ \hline
Rank & 16 \\
LoRA Aplha & 16 \\
LoRA Dropout & 0.05 \\
use\_dora & True \\
Learning rate & 2e-05 (1e-04 for gemma family) \\
Train batch size & 4 \\
Evaluation batch size & train\_batch\_size * 2 \\
Seed & 42 \\
WarmUp\_ratio & 0.05 \\
Gradient accumulation steps & 8 \\
Optimizer & AdamW \\
Scheduler & Cosine \\
Number of epochs & 10 \\
Target Modules & QKVOGUD \\ \hline
\end{tabular}%
}
\caption{Hyperparameters used in SFT training}
\label{tab:sft-params}
\end{table}

\section{Preliminary Comparison of Full Fine-Tuning and PEFT}
\label{app:peft_vs_full}

To justify our choice of parameter-efficient fine-tuning (PEFT) for SFT, we conducted preliminary experiments comparing full fine-tuning with several PEFT methods on the FrenchMedMCQA dataset~\citep{labrak:hal-03824241}.

We evaluated LoRA~\citep{hu2022lora}, DoRA~\citep{mao-etal-2024-dora}, and VeRA~\citep{kopiczko2024veravectorbasedrandommatrix} against full-parameter fine-tuning using identical training configurations. Results are reported in Table~\ref{tab:peft_vs_full}.

\begin{table*}[h]
\centering
\begin{tabular}{lcccc}
\toprule
 & LoRA & DoRA & VeRA & Full FT \\
\midrule
Exact Match & 0.2211 & \textbf{0.2435} & 0.1153 & 0.1121 \\
Hamming Distance & 0.4325 & \textbf{0.4627} & 0.3482 & 0.3143 \\
Trainable Params (\%) & 0.583 & 0.602 & 0.0037 & 100 \\
\bottomrule
\end{tabular}
\caption{Comparison of full fine-tuning and parameter-efficient methods on FrenchMedMCQA.}
\label{tab:peft_vs_full}
\end{table*}

We observe that PEFT methods, particularly DoRA, outperform full fine-tuning while requiring significantly fewer trainable parameters. In addition, full fine-tuning exhibited higher overfitting tendencies, with faster training loss convergence but weaker generalization performance on validation data.

These results support the use of parameter-efficient methods for SFT in our main experiments, as they provide a better trade-off between performance, efficiency, and generalization.

\section{Evaluation Metrics}
\label{app:metrics}

We provide here the formal definitions of the evaluation metrics used for MCQU and MCQ evaluation.

\paragraph{Exact Match (EM).}
Exact Match measures the proportion of predictions that exactly match the gold answer:
\[
\text{EM} = \frac{1}{N} \sum_{i=1}^{N} [\hat{y}_i = y_i],
\]
where $N$ denotes the number of questions, $y_i$ the gold answer, $\hat{y}_i$ the model prediction, and $[\cdot]$ is the indicator function.

\paragraph{Hamming Score.}
For multi-answer MCQ, we additionally report the Hamming score, which captures partial agreement between predicted and reference label sets:
\[
\text{Hamming Score} = \frac{1}{N} \sum_{i=1}^{N} \frac{|y_i \cap \hat{y}_i|}{|y_i \cup \hat{y}_i|}.
\]
This metric rewards partial correctness and is therefore better suited for evaluating multi-label predictions than Exact Match alone.

\section{Evaluation Benchmarks}
\label{app:benchmarks}
The adapted models are evaluated against their corresponding base models using benchmark datasets drawn from the \emph{test split} of MedInjection-FR. The evaluation suite includes both \emph{native French} benchmarks and \emph{translated English} benchmarks. For the translated benchmarks, English test sets were translated into French following the procedure described in section~\ref{sec:translation-quality}.

The benchmarks cover multiple task formats, including MCQU, MCQ and OEQA. This setup enables a controlled comparison of adaptation effects across both discriminative and generative biomedical reasoning tasks. Table~\ref{tab:benchmarks} summarizes the datasets used for evaluation and their respective sizes.

\begin{table}[H]
\centering
\resizebox{\columnwidth}{!}{%
\begin{tabular}{rrl}
\hline
\textbf{Dataset} & \textbf{\# Items} & \textbf{Task} \\ \hline
\multicolumn{3}{c}{\cellcolor[HTML]{DAE8FC}\textbf{NATIVE}} \\ \hline
 & 3\,384 & MCQ \\
 & 4\,343 & MCQU \\
\multirow{-3}{*}{MediQAl~\citep{bazoge2025mediqal}} & 4\,969 & OEQA \\
FrenchMedMCQA~\citep{labrak:hal-03824241} & 622 & MCQ \\
mlabonne/medical-mcqa-fr\footnote{\url{https://huggingface.co/datasets/mlabonne/medical-cases-fr}} & 150 & MCQ \\
mlabonne/medical-cases-fr\footnote{\url{https://huggingface.co/datasets/mlabonne/medical-mqca-fr}} & 352 & MCQ \\
FrBMedQA~\citep{kaddari2022frbmedqa} & 187 & MCQU \\
 & 343 & OEQA \\
\multirow{-2}{*}{S-Editions\footnote{\url{https://s-editions.fr/}}} & 183 & MCQ \\ \hline
\multicolumn{3}{c}{\cellcolor[HTML]{DAE8FC}\textbf{TRANSLATED}} \\ \hline
MedQA\_4options~\citep{jin2021disease} & 1\,273 & MCQU \\
MedQA\_5options~\citep{jin2021disease} & 1\,273 & MCQU\\
PubMedQA~\citep{jin2019pubmedqa} & 500 & MCQU \\
MedMCQA~\citep{pal2022medmcqa} & 4\,183 & MCQU \\
MMLU~\citep{hendrycks2021measuringmassivemultitasklanguage} & 1\,080 & MCQU \\
K-QA~\citep{manes-etal-2024-k} & 201 & OEQA \\
MMLU-PRO~\citep{wang2024mmlu} & 2\,333 & MCQU \\
MedXpertQA~\citep{zuo2025medxpertqa} & 2\,450 & MCQU \\ \hline
\end{tabular}%
}
\caption{Evaluation benchmarks used to compare adapted models with their base counterparts. All datasets correspond to the test split of MedInjection-FR.}
\label{tab:benchmarks}
\end{table}

\section{Prompt Templates}
\label{app:prompts}

\paragraph{Overview.}
We use a unified instruction format across all task types, both for supervised fine-tuning and for zero-shot evaluation. When available, we rely on the native chat templates provided by instruction-tuned models; otherwise, prompts are formatted as plain-text instruction--response pairs.

\paragraph{Shared Structure.}
All prompts begin with a high-level medical instruction, optionally followed by a contextual passage. The core components are:
\begin{enumerate}
    \item an instruction describing the task,
    \item the question (and answer options when applicable),
    \item an optional context section, and
    \item a response header indicating where the model output should begin.
\end{enumerate}

\paragraph{Task-Specific Constraints.}
The only variation across task types lies in the expected response format, which is explicitly stated in the instruction. Table~\ref{tab:prompt-templates} summarizes the templates used for each task.

\begin{table*}[t]
\centering
\small
\begin{tabular}{p{2.2cm} p{5.2cm} p{3.8cm}}
\toprule
\textbf{Task} & \textbf{Instruction Constraint} & \textbf{Expected Output} \\
\midrule
MCQU &
Respond only with the letter corresponding to the single correct answer. &
Single letter (e.g., \texttt{A}) \\
MCQ &
Respond only with the letters corresponding to all correct answers, separated by commas. &
Comma-separated letters (e.g., \texttt{A, C, D}) \\
OEQA &
Provide a free-form medical answer based on the instruction and context. &
Unconstrained text \\
\bottomrule
\end{tabular}
\caption{Summary of task-specific prompt templates and output constraints.}
\label{tab:prompt-templates}
\end{table*}

\paragraph{Canonical Prompt Format.}
The following abstract template illustrates the prompt structure shared across all tasks:
\newtcolorbox{promptbox}[1]{
  colback=black!7,        
  colframe=gray!90,      
  boxrule=0.6pt,
  arc=3pt,
  left=6pt,
  right=6pt,
  top=6pt,
  bottom=6pt,
  title=\textbf{#1},
  fonttitle=\small,
}
\begin{promptbox}{System prompt (training and evaluation) for MCQ}
\begin{lstlisting}
Lire l'instruction médicale suivante et fournir une réponse adaptée à la situation décrite.
Répondre uniquement avec la lettre correspondant à la ou les bonnes réponses séparées par des virgules. Exemple : A, C, D.

\end{lstlisting}
\end{promptbox}

\begin{promptbox}{System prompt (training and evaluation) for MCQU}
\begin{lstlisting}
Lire l'instruction médicale suivante et fournir une réponse adaptée à la situation décrite.
Répondre uniquement avec la lettre correspondant à la bonne réponse. Exemple : A.

\end{lstlisting}
\end{promptbox}

\begin{promptbox}{System prompt (training and evaluation) for OEQA}
\begin{lstlisting}
Lire l'instruction médicale suivante et fournir une réponse adaptée à  la situation décrite.

\end{lstlisting}
\end{promptbox}
\begin{promptbox}{User prompt (task-dependent)}
\begin{lstlisting}
### Instruction:
[Question (+ options for MCQ tasks)]

### Contexte:
[Context, if available]

### Réponse:
\end{lstlisting}
\end{promptbox}

\paragraph{Chat-Based Formatting.}
For instruction-tuned models providing an explicit chat interface, the same content is mapped to role-based messages as follows:
\begin{itemize}
    \item \textbf{System}: high-level medical instruction (shared across tasks),
    \item \textbf{User}: task instruction, question, and optional context,
    \item \textbf{Assistant}: model-generated answer.
\end{itemize}

This formulation ensures consistent supervision and evaluation across models with different input formatting requirements.

\section{MCQA and OEQA Results}
\label{app:greedy-results}
\input{greedy-main-results}
\subsection{MCQA Greedy Decoding}
Table~\ref{tab:main-result-greedy} reports performance on MCQA across the studied model families  (Gemma-4B, Mistral-7B, Llama-7B, and Llama-13B), three initialization types (General, Instruct, Medical), and four adaptation strategies (Base, CPT, SFT, CPT+SFT). Results are shown for both standard multiple-answer MCQs (MCQ) and single-answer MCQs (MCQU), using Exact Match (EM), Hamming score for MCQ, and aggregated EM. The reported results here are obtained using greedy decoding.

\paragraph{Effectiveness of Adaptation Strategy:} Under greedy decoding, SFT clearly dominates all other adaptation strategies across MCQ, MCQU, and aggregated metrics. Unlike constrained decoding, where CPT+SFT often ranks first, greedy decoding exposes a much sharper separation between strategies:

\[ \textsc{Base} \ll \textsc{\CPTtext} \ll \textsc{\CPTSFTtext} < \textsc{\SFTtext}\]

Across nearly all model families and initializations, SFT yields the highest MCQ EM, MCQ Hamming, MCQU EM, and aggregated EM. This trend is particularly strong for instruction-tuned models (Mistral, Llama-7B, Llama-13B), where SFT consistently delivers large absolute gains, often by wide margins. As in the constrained decoding setting, when \textsc{CPT+SFT} outperforms \textsc{SFT}, the performance gap is generally smaller than in configurations where \textsc{SFT} outperforms \textsc{CPT+SFT}.

CPT alone remains unstable under greedy decoding. While it sometimes improves over the base model, it frequently underperforms SFT and can even degrade MCQU and aggregated scores. Importantly, CPT+SFT does not systematically improve over SFT in greedy decoding and often performs worse, indicating that the benefits of CPT are largely redundant once task supervision is introduced and decoding constraints are removed.

Overall, greedy decoding amplifies the advantages of task-aligned supervision, making SFT the best adaptation strategy when decoding is unconstrained.

\paragraph{Impact of Model Initialization:} Model initialization plays a stronger and more consistent role under greedy decoding than under constrained decoding. Across all families and metrics, instruction-tuned models dominate.
General models benefit from SFT but remain consistently below instruction-tuned counterparts. Medical models, while improving with SFT, never achieve the best greedy decoding performance, confirming that domain pretraining alone is insufficient without strong instruction alignment.

This contrasts with constrained decoding, where general and medical models occasionally remain competitive. Under greedy decoding, instruction tuning becomes a necessary condition for strong performance.
\paragraph{MCQA Greedy Decoding Guidelines:} For greedy decoding in medical MCQA, the optimal configuration is to start from an instruction-tuned model and apply SFT only. CPT and CPT+SFT offer no consistent benefit in this setting and can be safely avoided unless constrained decoding is explicitly required.
\subsection{OEQA Overlap-based Evaluation}

The right part of Table~\ref{tab:main-result-greedy} reports the overlap-based metrics BLEU and METEOR. Both metrics exhibit trends consistent with ROUGE-L, with improvements primarily driven by CPT. In a few isolated cases, CPT+SFT yields additional gains on METEOR, but with small differences with when compared with CPT. Regarding model initialization, BLEU and METEOR consistently favor instruction-tuned models as the strongest starting point.

\section{Statistical Significance}
\label{app:significance}
\input{significance-MCQA}
\input{sig-stat-oeqa}

We assess whether observed differences between adaptation strategies and initialization choices are statistically significant using paired bootstrap significance testing. For each comparison, we compute the per-instance score difference (EM for MCQ/MCQU; judge-based correctness for OEQA) and report a two-sided $p$-value (\texttt{p\_two\_sided}). Statistical significance is determined by comparing this $p$-value against a predefined threshold $\alpha$. We report results using a Bonferroni-corrected threshold to control for multiple comparisons.

For comparisons between adaptation strategies within a model family, we perform 12 pairwise tests per family, yielding a corrected threshold of $\alpha_{\text{Bonferroni}} = 0.05 / 12$. For comparisons between model initialization types under a fixed adaptation strategy, we perform 9 pairwise tests per family, yielding $\alpha_{\text{Bonferroni}} = 0.05 / 9$. The applied threshold (\texttt{alpha\_Bonferoni}) and the resulting significance decision (\texttt{significant\_Bonferroni}) are reported explicitly in Tables~\ref{tab:stat-mcq} and~\ref{tab:stat-oeqa}.

We define the mean difference as $\Delta=\mathrm{score}(\texttt{model\_a})-\mathrm{score}(\texttt{model\_b})$ (not shown in the tables). Therefore, if the confidence interval is entirely above zero, \texttt{model\_a} performs better; if it is entirely below zero, \texttt{model\_b} performs better. A comparison is considered statistically significant if the corrected decision is TRUE.

\subsection{Interpretation of comparison IDs}
Each row in Tables~\ref{tab:stat-mcq} and~\ref{tab:stat-oeqa} corresponds to a specific pairwise comparison between two models (\texttt{model\_a} vs.\ \texttt{model\_b}). The \texttt{id} field encodes the purpose of the comparison:
(i) \textbf{IDs A--C} compare models \emph{within the same model type} (GENERAL, INSTRUCT, or MEDICAL) in order to quantify the effect of adaptation strategies (CPT, SFT, CPT+SFT) relative to a fixed initialization.
(ii) \textbf{IDs D--F} compare models \emph{across model types} under a fixed adaptation strategy, in order to identify the most effective initialization point (GENERAL vs.\ INSTRUCT vs.\ MEDICAL) for downstream adaptation.

\subsection{Decoding conditions}
For MCQ/MCQU, Table~\ref{tab:stat-mcq} reports significance results separately for greedy and constrained decoding. For OEQA, Table~\ref{tab:stat-oeqa} reports strategy-level comparisons under the evaluation setting used for the main experiments.

\section{Near-Miss Rates in MCQA}
\label{app:nearmiss}
\input{near-miss-rate}
To better characterize model behavior beyond EM accuracy in MCQA, we analyze \emph{near-miss} predictions. Table~\ref{tab:near-miss-MCQ-MCQu} shows the near-miss rates obtained for MCQ and MCQU  across different Mistral-based model variants and adaptation strategies.

\section{OEQA Evaluation: Verbosity Bias}
\label{app:oeqa-verbosity}
\input{oeqa-verbosity}
To investigate verbosity bias in OEQA, we compute descriptive statistics of generated answer lengths across all models. Table~\ref{tab:oeqa-verb} reports mean, median, and standard deviation of word and character counts over greedy OEQA outputs.

\section{English vs.\ French Benchmarks: Full Numeric Results}
\label{app:en-vs-fr}
\input{en-vs-fr}
The main paper reports averaged results using \emph{constrained decoding} (Figure~\ref{fig:en-vs-fr}). Table~\ref{tab:en-vs-fr} provides the complete \emph{numeric} EM results for both \emph{greedy} and \emph{constrained} decoding on the native English MCQU benchmarks (MCQU-EN) and their French translations (MCQU-FR).

\subsection{Greedy decoding analysis}
The greedy decoding results reported in Table~\ref{tab:en-vs-fr} exhibit the same overall tendencies as those observed under constrained decoding in section~\ref{sec:en-vs-fr}. For the Mistral family, greedy decoding consistently yields higher performance on the French translations than on the original English benchmarks, both before and after adaptation. Conversely, Gemma and Llama models generally perform better on native English benchmarks under greedy decoding, and this advantage is preserved after French medical adaptation.

As with constrained decoding, adaptation on French medical data improves performance in both languages under greedy decoding, indicating effective cross-lingual transfer. While absolute EM scores differ between decoding strategies, greedy decoding generally producing lower scores, the relative ordering between English and French benchmarks and the direction of adaptation effects remain consistent. These results suggest that the cross-lingual patterns reported in the main paper are robust to the choice of decoding strategy.

\subsection{Significance testing (English vs.\ French)}
\label{app:stat-en-vs-fr}
To assess whether the English--French performance gaps are statistically significant, we perform paired significance testing \emph{separately for each model configuration}, i.e., for each combination of (model family/type, adaptation strategy, decoding type). For each configuration, we compute the per-item EM difference between MCQU-EN and MCQU-FR on matched translated instances, and estimate a 95\% confidence interval for the mean difference together with a two-sided $p$-value. Because each test compares a model strictly with itself across languages and each English--French pair is independent of the others, we do not apply a Bonferroni correction. Table~\ref{tab:stat-en-vs-fr} reports the resulting confidence intervals and significance decisions.

\input{stat-en-vs-fr}

\section{Effect of Translated Benchmarks on Performance and Confidence}
\begin{figure}[H]
\centering
\includegraphics[width=\columnwidth]{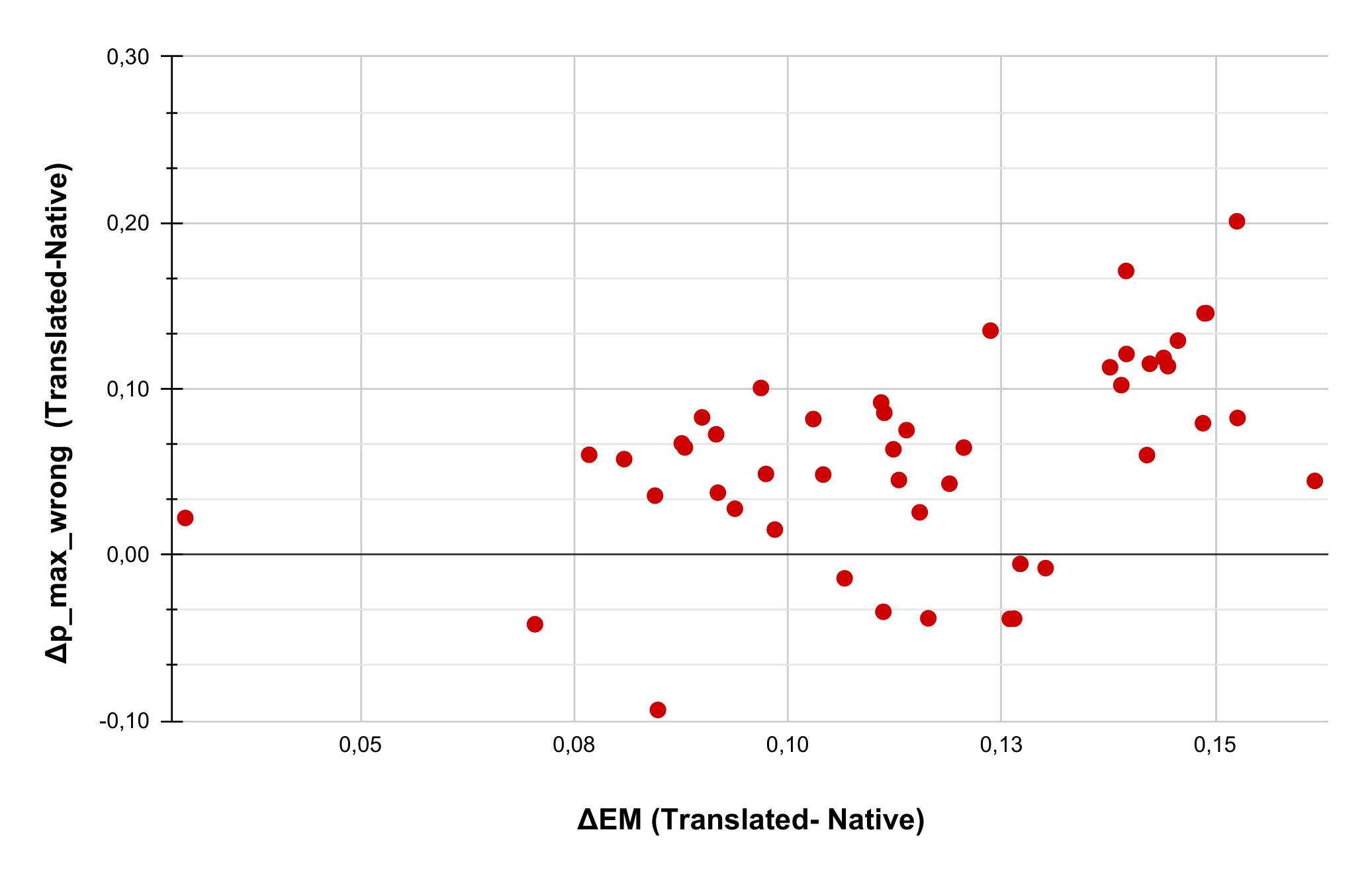}
\caption{Relationship between accuracy gain ($\Delta EM$) and change in confidence on incorrect predictions ($\Delta p_{\max,\mathrm{wrong}}$) between the translated and native benchmarks. Positive values of $\Delta p_{\max,\mathrm{wrong}}$ indicate increased confidence on incorrect predictions.}
\label{fig:nat-vs-tad-pmax}
\end{figure}

\section{Computational Resources and Environmental Impact}
\label{app:conso}

Table~\ref{tab:conso} summarizes the computational resources and environmental impact associated with each adaptation strategy, aggregated by model size. For clarity and conciseness, we do not report the consumption of each individual training run. Instead, we provide a representative summary per model size and per adaptation strategy. In total, 36 training runs were performed across all experiments.
\input{gpu-conso}
The CPT+SFT strategy is not reported as a separate entry in the table, as its computational cost and environmental impact correspond to the sum of the CPT and SFT phases. Reporting CPT and SFT independently therefore fully characterizes the overall resource usage of the combined strategy.

We report, for each configuration, the dataset size, number of epochs, batch size, GPU type, GPU memory, number of GPUs, total training time, estimated carbon emissions (in gCO$_2$e), and estimated monetary cost (in USD). Carbon emissions and cost estimates are derived from documented power consumption profiles and usage costs of the underlying high-performance computing infrastructure. All experiments were conducted on the Jean Zay supercomputer operated by GENCI-IDRIS\footnote{\url{http://www.idris.fr/docs/jean-zay/jean-zay/jean-zay-presentation/}}.

\subsection{Analysis}
Overall, the results highlight a clear contrast between CPT and SFT in terms of computational cost and environmental impact. CPT is consistently the most resource-intensive strategy, driven by large-scale datasets, longer effective compute time, and high degrees of GPU parallelism. 

In contrast, SFT incurs substantially lower emissions and monetary costs across all model sizes. This difference is not only due to the smaller dataset size, but also to the use of parameter-efficient fine-tuning: SFT is implemented with DoRA adapters rather than full weight updates, significantly reducing both memory usage and energy consumption. Despite longer wall-clock durations in some configurations, the overall compute footprint of SFT remains markedly lower than that of CPT.

As model size increases, CPT costs grow rapidly, particularly for the 13B setting, where energy consumption and carbon emissions increase sharply. SFT, while also scaling with model size, remains comparatively efficient due to its parameter-efficient design. These findings underscore the importance of adaptation strategies that balance performance gains with computational and environmental sustainability.

\section{Pretraining Data Contamination Study: Was \textsc{NACHOS} Seen During Pretraining?}
\label{app:contamination}

Because most of the base models we evaluate (Gemma, MedGemma, Mistral, and Llama) do not disclose their full pretraining mixtures, we conducted a small contamination study to probe whether the French biomedical \textsc{NACHOS} corpus may have been included (or partially included) in their pretraining data. This appendix reports two complementary, lightweight detection protocols inspired by the broader literature on memorization and pretraining-data detection in large language models~\citep{ravaut2024comprehensive}

\subsection{Protocol 1: Prefix--Continuation Reproduction + Likelihood Heuristics}
\label{app:contamination:proto1}

\paragraph{Idea.}
If a model has memorized (or near-memorized) training documents, conditioning on a prefix may lead it to reproduce the exact continuation, or to assign a noticeably higher likelihood to the true continuation than to a perturbed version. This is conceptually related to training-data extraction / memorization diagnostics used in prior work~\citep{carlini2021extracting}. 

\paragraph{Implementation.}
Using a sample of $n{=}1915$ \textsc{NACHOS} documents, we split each document into a \emph{prefix} (first 400 characters) and a \emph{continuation} (rest). For each sampled document, we:
(i) generate up to 200 new tokens from the prefix (greedy decoding), and compute ROUGE-L between the generated continuation and the gold continuation;
(ii) compute the length of the longest common prefix (LCP) between generated and gold continuations;
(iii) compute the perplexity of the gold continuation conditioned on the prefix, and compare it to the perplexity of a lightly perturbed continuation (character swaps + whitespace noise), reporting the ratio $\mathrm{PPL}(\text{gold})/\mathrm{PPL}(\text{perturbed})$.
We flag a case as ``suspicious'' if any of the following holds: ROUGE-L $\ge 0.7$, LCP $\ge 200$ characters, or $\mathrm{PPL}(\text{gold})/\mathrm{PPL}(\text{perturbed}) \le 0.85$.

\paragraph{Results.}
Across models, ROUGE-L remained very low and we observed no exact continuation matches, which does \emph{not} support verbatim memorization of long continuations under this setup. However, the fraction of items flagged as ``suspicious'' is extremely high (0.82--0.96), which indicates that our heuristic is likely over-sensitive (in particular, the perturbation and/or the chosen ratio threshold may dominate the flagging decision).

\begin{itemize}
    \item Llama-2-7B: ROUGE-L $=0.031$, exact matches $=0/1915$, suspicious fraction $=0.959$.
    \item Llama-2-13B: ROUGE-L $=0.020$, exact matches $=0/1915$, suspicious fraction $=0.964$.
    \item Mistral-7B: ROUGE-L $=0.014$, exact matches $=0/1915$, suspicious fraction $=0.944$.
    \item MedGemma-4B: ROUGE-L $=0.018$, exact matches $=0/1915$, suspicious fraction $=0.821$.
    \item Gemma-3-4B: ROUGE-L $=0.019$, exact matches $=0/1915$, suspicious fraction $=0.835$.
\end{itemize}

\paragraph{Interpretation.}
Given the near-zero reproduction scores (ROUGE-L, exact match) but massive ``suspicious'' rates, this first protocol is inconclusive as a contamination detector in our setting: it does not show direct copying, and the likelihood-based heuristic is too unstable without a careful calibration procedure and stronger perturbations/controls. This is consistent with known difficulties of turning likelihood signals into reliable membership decisions without explicit calibration~\citep{yeom2018privacy}.

\subsection{Protocol 2: DC-PDD (Divergence-based Calibration Pretraining Data Detection)}
\label{app:contamination:proto2}

\paragraph{Idea.}
We also tested a dedicated pretraining-data detection score, DC-PDD, which estimates a per-text statistic $\beta(x)$ combining (i) the model probability of next tokens and (ii) reference token frequencies estimated from a large background corpus $D'$ (here, French OSCAR\footnote{\url{https://oscar-project.org/}}). The method is designed to be more robust than raw perplexity by incorporating a calibration term from $D'$~\citep{zhang-etal-2024-pretraining}. 

\paragraph{Implementation.}
For each model, we first build a tokenizer-specific unigram table $p(v;D')$ from OSCAR-FR (streaming counts, capped number of documents), then compute DC-PDD $\beta(x)$ on:
(i) 1,000 \textsc{NACHOS} samples, and
(ii) a \emph{synthetic control} set (``non-member'') of biomedical texts generated to be \emph{unlikely} to appear in any public pretraining mixture.
We report distributional statistics (median, p75/p90/p95, mean, std) for both sets and a separation diagnostic $\Delta_\text{median} = \mathrm{median}(\beta_\text{nachos})-\mathrm{median}(\beta_\text{control})$.

\paragraph{Why synthetic controls? (Major limitation)}
Gemma-family models were released recently (June 2025), and we could not reliably curate a sufficiently large set of \emph{web-native biomedical French texts written after the model release date} to serve as a credible ``definitely-non-member'' control. As a consequence, we used synthetic biomedical controls, which weakens the study: synthetic controls differ from natural corpora in style and token statistics, and thus may artificially inflate separation (or mask it), independently of membership. We therefore treat DC-PDD results as \emph{indicative only}, not as evidence of true pretraining inclusion.

\paragraph{Results.}
DC-PDD yields consistently \emph{lower} scores on \textsc{NACHOS} than on the synthetic controls (negative $\Delta$), suggesting the models assign \emph{slightly} more ``in-distribution'' likelihood structure to \textsc{NACHOS} than to the synthetic texts. The separation is small for Mistral/Llama and somewhat larger for Gemma/MedGemma:

\begin{itemize}
    \item Mistral-7B: $\Delta_\text{median} \approx -3.23\times 10^{-4}$.
    \item Llama-2-7B: $\Delta_\text{median} \approx -3.58\times 10^{-4}$.
    \item Llama-2-13B: $\Delta_\text{median} \approx -2.94\times 10^{-4}$.
    \item Gemma-3-4B: $\Delta_\text{median} \approx -8.20\times 10^{-4}$.
    \item MedGemma-4B: $\Delta_\text{median} \approx -6.77\times 10^{-4}$.
\end{itemize}

\paragraph{Interpretation.}
While DC-PDD produces a consistent ordering (Nachos $<$ Control), this cannot be confidently attributed to pretraining membership because our control set is synthetic and therefore not distribution-matched. In other words, the observed separation may reflect \emph{domain/style differences} rather than exposure during pretraining. As prior work emphasizes, robust pretraining-data detection typically requires carefully constructed controls and/or calibrated baselines (e.g., Min-K\% variants, calibrated likelihood tests), which we could not fully satisfy here~\citep{zhang-etal-2024-pretraining}.

\subsection{Summary and Takeaways}
Overall, these experiments do not provide strong evidence for (or against) \textsc{NACHOS} being included in the undisclosed pretraining mixtures:
(i) we do not observe continuation copying under our greedy prefix--continuation setup;
(ii) DC-PDD shows a small but consistent separation between \textsc{NACHOS} and synthetic controls, but the lack of a reliable post-release, naturally occurring biomedical control corpus makes the conclusion weak.
We therefore report these results for transparency, but we do not use them to support any causal claim about pretraining contamination in the main analysis.

\end{document}

%% file: main-results-bis.tex

\begin{table*}[!t]
\centering

\begingroup
\setlength{\tabcolsep}{3pt}       
\renewcommand{\arraystretch}{1}
\footnotesize                     

\scalebox{0.8}{ 
\begin{tabular}{ll S S S S | S S S}
\toprule
\multirow{2}{*}{\textbf{Model Type}} &
\multirow{2}{*}{\textbf{Strategy}} &
\multicolumn{2}{c}{\textbf{MCQ}} &
\textbf{MCQU} &
\textbf{Aggregation} &
\multicolumn{3}{c}{\textbf{OEQA}} \\
\cmidrule(lr){3-4}\cmidrule(lr){5-6}\cmidrule(lr){7-9}
& &
{\textbf{EM}} & {\textbf{Hamming}} &
{\textbf{EM}} & {\textbf{EM}} &
{\textbf{ROUGE-L}} & {\textbf{BERT-F1}} & {\textbf{Judge}} \\
\midrule

\multicolumn{9}{c}{\cellcolor[HTML]{CFE2F3}\textbf{Gemma-4B}}\\ \hline

\addlinespace[1pt]

\multirow{4}{*}{\textbf{GENERAL}} & Base      & 2.24 & 30.17 & 27.11 & 14.68 & 7.11  & 46.62 & 25.09 \\
                & {\color[HTML]{0000FF}CPT}     & 0.73 & 11.71 & 25.83 & 13.28 & \uline{\textbf{10.18}} & \uline{\textbf{49.07}} & \textbf{25.71} \\
                & {\color[HTML]{6AA84F}SFT}     & 3.73 & \textbf{43.57} & 32.36 & 18.05 & 8.23  & 47.76 & 21.60 \\
                & {\color[HTML]{FF0000}CPT+SFT} & \textbf{3.90} & 42.81 & \uline{\textbf{32.59}} & \uline{\textbf{18.25}} & 6.01  & 45.78 & 24.80 \\
\midrule
\multirow{4}{*}{\textbf{INSTRUCT}} & Base      & \uline{\textbf{4.83}} & 44.01 & 29.30 & 17.06 & \textbf{7.38}  & \textbf{48.94} & \uline{\textbf{47.71}} \\
                 & {\color[HTML]{0000FF}CPT}     & 3.47 & 46.00 & 25.05 & 14.26 & 4.57  & 42.77 & 13.35 \\
                 & {\color[HTML]{6AA84F}SFT}     & 3.68 & \uline{\textbf{48.22}} & \textbf{31.95} & \textbf{17.81} & 2.20  & 38.26 & 20.21 \\
                 & {\color[HTML]{FF0000}CPT+SFT} & 3.42 & 48.14 & 30.73 & 17.07 & 3.12  & 40.90 & 20.76 \\
\midrule
\multirow{4}{*}{\textbf{MEDICAL}} & Base      & 1.98 & 31.46 & 26.41 & 14.19 & \textbf{7.38}  & \textbf{48.94} & \textbf{22.01} \\
                & {\color[HTML]{0000FF}CPT}     & 1.68 & 24.04 & 25.13 & 13.41 & 6.77  & 45.18 & 1.95  \\
                & {\color[HTML]{6AA84F}SFT}     & \textbf{3.54} & \textbf{46.22} & 30.66 & 17.10 & 5.70  & 45.66 & 14.23 \\
                & {\color[HTML]{FF0000}CPT+SFT} & 3.38 & 43.28 & \textbf{30.86} & \textbf{17.12} & 5.04  & 43.05 & 11.41 \\

\hline
\multicolumn{9}{c}{\cellcolor[HTML]{CFE2F3}\textbf{Mistral-7B}}\\ \hline

\addlinespace[1pt]

\multirow{4}{*}{\textbf{GENERAL}} & Base      & 0.37 & 5.40  & 28.52 & 14.44 & 5.82  & 44.21 & \textbf{27.23} \\
                & {\color[HTML]{0000FF}CPT}     & 3.54 & 30.50 & 27.21 & 15.37 & 7.22  & 46.32 & 24.57 \\
                & {\color[HTML]{6AA84F}SFT}     & 5.24 & 21.62 & \uline{\textbf{32.88}} & 19.06 & \textbf{8.83}  & \textbf{48.02} & 22.93 \\
                & {\color[HTML]{FF0000}CPT+SFT} & \textbf{6.13} & \textbf{30.86} & 32.29 & \uline{\textbf{19.21}} & 6.62  & 46.35 & 24.89 \\
\midrule
\multirow{4}{*}{\textbf{INSTRUCT}} & Base      & 4.86 & 23.53 & 24.92 & 14.89 & 7.34  & 49.66 & 30.14 \\
                 & {\color[HTML]{0000FF}CPT}     & \uline{\textbf{7.32}} & \uline{\textbf{36.18}} & 28.79 & 18.06 & \uline{\textbf{13.51}} & \uline{\textbf{53.87}} & \uline{\textbf{37.59}} \\
                 & {\color[HTML]{6AA84F}SFT}     & 6.80 & 23.42 & \textbf{31.61} & \uline{\textbf{19.21}} & 12.41 & 52.72 & 17.63 \\
                 & {\color[HTML]{FF0000}CPT+SFT} & 5.45 & 32.47 & 30.09 & 17.77 & 9.02  & 48.99 & 32.13 \\
\midrule
\multirow{4}{*}{\textbf{MEDICAL}} & Base      & 2.80 & 17.47 & 26.69 & 14.74 & 11.34 & \textbf{51.58} & 20.76 \\
                & {\color[HTML]{0000FF}CPT}     & 3.57 & 24.43 & 25.73 & 14.65 & \textbf{12.41} & 51.45 & 17.89 \\
                & {\color[HTML]{6AA84F}SFT}     & 3.36 & 26.37 & 31.62 & 17.49 & 8.75  & 47.86 & 16.72 \\
                & {\color[HTML]{FF0000}CPT+SFT} & \textbf{4.94} & \textbf{27.27} & \textbf{32.58} & \textbf{18.76} & 9.15  & 48.63 & \textbf{24.96} \\

\hline
\multicolumn{9}{c}{\cellcolor[HTML]{CFE2F3}\textbf{Llama-7B}}\\ \hline
\addlinespace[1pt]

\multirow{4}{*}{\textbf{GENERAL}} & Base      & 1.33 & 12.01 & 25.72 & 13.53 & 5.05  & 41.27 & 9.39  \\
                & {\color[HTML]{0000FF}CPT}     & 1.12 & 12.86 & 25.59 & 13.36 & \textbf{10.58} & \textbf{47.85} & 3.78  \\
                & {\color[HTML]{6AA84F}SFT}     & 2.66 & 28.41 & 28.93 & 15.80 & 6.02  & 44.49 & 7.67  \\
                & {\color[HTML]{FF0000}CPT+SFT} & \textbf{3.17} & \uline{\textbf{46.00}} & \textbf{29.89} & \textbf{16.53} & 5.85  & 44.67 & \textbf{12.26} \\
\midrule
\multirow{4}{*}{\textbf{INSTRUCT}} & Base      & 3.95 & 34.43 & 25.08 & 14.51 & 2.57  & 43.85 & 25.35 \\
                 & {\color[HTML]{0000FF}CPT}     & 3.93 & \textbf{42.67} & 25.07 & 14.50 & 11.16 & \uline{\textbf{51.37}} & 26.06 \\
                 & {\color[HTML]{6AA84F}SFT}     & \uline{\textbf{5.12}} & 21.06 & \textbf{29.32} & \textbf{17.22} & \uline{\textbf{11.44}} & 51.28 & 12.92 \\
                 & {\color[HTML]{FF0000}CPT+SFT} & 3.13 & 25.98 & 25.07 & 14.10 & 9.92  & 50.99 & \uline{\textbf{27.84}} \\
\midrule
\multirow{4}{*}{\textbf{MEDICAL}} & Base      & 0.23 & 2.90  & 24.43 & 12.33 & 5.61  & 43.25 & 12.50 \\
                & {\color[HTML]{0000FF}CPT}     & 2.37 & 28.06 & 25.60 & 13.99 & \textbf{8.00}  & \textbf{45.79} & 13.14 \\
                & {\color[HTML]{6AA84F}SFT}     & 3.24 & 29.10 & 30.44 & 16.84 & 5.40  & 42.25 & 9.50  \\
                & {\color[HTML]{FF0000}CPT+SFT} & \textbf{3.80} & \textbf{44.95} & \uline{\textbf{31.52}} & \uline{\textbf{17.66}} & 5.87  & 44.34 & \textbf{17.39} \\

\hline
\multicolumn{9}{c}{\cellcolor[HTML]{CFE2F3}\textbf{Llama-13B}}\\ \hline

\addlinespace[1pt]

\multirow{4}{*}{\textbf{GENERAL}} & Base      & 2.14 & 21.20 & 26.11 & 14.13 & 2.10  & 33.25 & 11.79 \\
                & {\color[HTML]{0000FF}CPT}     & 2.53 & 19.17 & 26.99 & 14.76 & \uline{\textbf{14.12}} & \textbf{50.36} & 5.66  \\
                & {\color[HTML]{6AA84F}SFT}     & \textbf{3.54} & \textbf{40.49} & 30.95 & 17.24 & 5.60  & 43.19 & 14.88 \\
                & {\color[HTML]{FF0000}CPT+SFT} & 3.34 & 29.59 & \uline{\textbf{32.36}} & \textbf{17.85} & 6.30  & 45.45 & \textbf{20.38} \\
\midrule
\multirow{4}{*}{\textbf{INSTRUCT}} & Base      & 0.09 & 29.74 & 21.52 & 10.81 & 3.40  & 45.31 & 30.02 \\
                 & {\color[HTML]{0000FF}CPT}     & 5.63 & \textbf{37.40} & 25.01 & 15.32 & 12.34 & \uline{\textbf{53.07}} & \uline{\textbf{36.19}} \\
                 & {\color[HTML]{6AA84F}SFT}     & 6.58 & 23.96 & 30.20 & 18.39 & 11.54 & 50.94 & 11.81 \\
                 & {\color[HTML]{FF0000}CPT+SFT} & \uline{\textbf{7.77}} & 25.26 & \textbf{31.58} & \uline{\textbf{19.68}} & \textbf{12.86} & 52.46 & 20.22 \\
\midrule
\multirow{4}{*}{\textbf{MEDICAL}} & Base      & 1.77 & 11.82 & 24.62 & 13.19 & 5.00  & 42.11 & 10.86 \\
                & {\color[HTML]{0000FF}CPT}     & 2.26 & 30.87 & 24.10 & 13.18 & 8.00  & 45.79 & 13.39 \\
                & {\color[HTML]{6AA84F}SFT}     & 3.12 & 41.24 & 30.62 & 16.87 & 6.85  & 45.22 & 13.77 \\
                & {\color[HTML]{FF0000}CPT+SFT} & \textbf{3.24} & \uline{\textbf{45.59}} & \textbf{32.25} & \textbf{17.74} & \textbf{8.38}  & \textbf{46.29} & \textbf{19.55} \\
\bottomrule
\end{tabular}
}
\caption{Constrained decoding results (\%) for MCQ/MCQU and OEQA. Aggregation corresponds to average EM over MCQ and MCQU. \textbf{Bold} denotes the best strategy, and \uline{underlining} the best initialization.}
\label{tab:main-results}
\endgroup
\end{table*}

%% file: greedy-main-results.tex
\begin{table*}[]
\centering
\resizebox{0.7\textwidth}{!}{%
\begin{tabular}{cccccc|cc}
\hline
 &  & \multicolumn{2}{c}{\textbf{MCQ}} & \textbf{MCQU} & \textbf{Aggregation} & \multicolumn{2}{c}{\textbf{OEQA}} \\ \cline{3-8} 
\multirow{-2}{*}{\textbf{Model Type}} & \multirow{-2}{*}{\textbf{Strategy}} & \textbf{EM} & \textbf{Hamming} & \textbf{EM} & \textbf{EM} & \multicolumn{1}{l}{\textit{\textbf{BLEU}}} & \multicolumn{1}{l}{\textit{\textbf{METEOR}}} \\ \hline
\multicolumn{8}{c}{\cellcolor[HTML]{CFE2F3}\textit{\textbf{Gemma-4B}}} \\ \hline
 & Base & 0.56 & 5.15 & 5.88 & 3.22 & 0.92 & 8.09 \\ 
 & {\color[HTML]{0000FF} CPT} & 0.03 & 0.68 & 8.53 & 4.28 & {\ul\textbf{1.68}} & \textbf{8.35} \\ 
 & {\color[HTML]{6AA84F} SFT} & \textbf{1.73} & \textbf{19.48} & \textbf{19.81} & \textbf{10.77} & 1.03 & 7.42 \\ 
\multirow{-4}{*}{\textbf{GENERAL}} & {\color[HTML]{FF0000} CPT+SFT} & 1.67 & 15.34 & 19.52 & 10.60 & 0.56 & 6.99 \\ \hline
 & Base & {\ul \textbf{6.81}} & {\ul \textbf{40.75}} & 28.88 & {\ul \textbf{17.85}} & \textbf{0.76} & {\ul \textbf{10.41}} \\ 
 & {\color[HTML]{0000FF} CPT} & 0.07 & 1.72 & 1.37 & 0.72 & 0.46 & 5.89 \\ 
 & {\color[HTML]{6AA84F} SFT} & 1.38 & 10.17 & {\ul \textbf{31.87}} & 16.63 & 0.09 & 5.32 \\ 
\multirow{-4}{*}{\textbf{INSTRUCT}} & {\color[HTML]{FF0000} CPT+SFT} & 1.22 & 8.61 & 30.66 & 15.94 & 0.21 & 6.93 \\ \hline
 & Base & 1.22 & 11.28 & 11.47 & 6.34 & 0.76 & {\ul \textbf{10.41}} \\ 
 & {\color[HTML]{0000FF} CPT} & 1.62 & \textbf{18.11} & 11.18 & 6.40 & \textbf{1.04} & 5.26 \\ 
 & {\color[HTML]{6AA84F} SFT} & 1.15 & 11.51 & \textbf{17.91} & 9.53 & 0.56 & 6.98 \\ 
\multirow{-4}{*}{\textbf{MEDICAL}} & {\color[HTML]{FF0000} CPT+SFT} & \textbf{1.67} & 16.54 & 17.63 & \textbf{9.65} & 0.34 & 5.60 \\ \hline
\multicolumn{8}{c}{\cellcolor[HTML]{CFE2F3}\textit{\textbf{Mistral-7B}}} \\ \hline
 & Base & 0.15 & 2.90 & 4.04 & 2.09 & 0.66 & 7.32 \\ 
 & {\color[HTML]{0000FF} CPT} & 0.44 & 3.75 & 13.71 & 7.07 & 0.99 & 7.69 \\ 
 & {\color[HTML]{6AA84F} SFT} & \textbf{1.79} & \textbf{20.62} & \textbf{19.88} & \textbf{10.84} & \textbf{1.04} & 7.59 \\ 
\multirow{-4}{*}{\textbf{GENERAL}} & {\color[HTML]{FF0000} CPT+SFT} & 1.42 & 17.95 & 19.57 & 10.49 & 0.63 & \textbf{8.66} \\ \hline
 & Base & 3.42 & 26.94 & 21.46 & 12.44 & 1.12 & 7.70 \\ 
 & {\color[HTML]{0000FF} CPT} & 4.10 & 29.16 & 27.63 & 15.87 & {\ul \textbf{2.34}} & {\ul \textbf{10.90}} \\ 
 & {\color[HTML]{6AA84F} SFT} & {\ul \textbf{11.94}} & {\ul \textbf{47.39}} & {\ul \textbf{31.52}} & {\ul \textbf{21.73}} & 1.65 & 7.08 \\ 
\multirow{-4}{*}{\textbf{INSTRUCT}} & {\color[HTML]{FF0000} CPT+SFT} & 1.85 & 17.56 & 29.84 & 15.85 & 1.09 & 9.46 \\ \hline
 & Base & 2.24 & 19.09 & 13.39 & 7.81 & 1.73 & 8.39 \\ 
 & {\color[HTML]{0000FF} CPT} & \textbf{2.25} & 18.39 & 12.19 & 7.22 & \textbf{1.93} & \textbf{9.54} \\ 
 & {\color[HTML]{6AA84F} SFT} & 1.95 & \textbf{20.68} & 18.47 & 10.21 & 1.06 & 6.75 \\ 
\multirow{-4}{*}{\textbf{MEDICAL}} & {\color[HTML]{FF0000} CPT+SFT} & 1.44 & 16.43 & \textbf{19.33} & \textbf{10.38} & 1.05 & 8.02 \\ \hline
\multicolumn{8}{c}{\cellcolor[HTML]{CFE2F3}\textit{\textbf{Llama-7B}}} \\ \hline
 & Base & 0.17 & 2.69 & 9.46 & 4.82 & 0.49 & 5.91 \\ 
 & {\color[HTML]{0000FF} CPT} & 1.13 & 10.16 & 5.83 & 3.48 & \textbf{1.39} & 5.90 \\ 
 & {\color[HTML]{6AA84F} SFT} & \textbf{1.82} & \textbf{24.64} & 16.08 & 8.95 & 0.51 & \textbf{7.26} \\ 
\multirow{-4}{*}{\textbf{GENERAL}} & {\color[HTML]{FF0000} CPT+SFT} & 1.61 & 17.50 & \textbf{17.11} & \textbf{9.36} & 0.52 & 6.75 \\ \hline
 & Base & 0.03 & 0.64 & 0.04 & 0.03 & 0.40 & 2.51 \\ 
 & {\color[HTML]{0000FF} CPT} & 4.92 & 40.87 & 0.04 & 2.48 & {\ul \textbf{1.72}} & 8.64 \\ 
 & {\color[HTML]{6AA84F} SFT} & {\ul \textbf{7.72}} & {\ul \textbf{42.70}} & {\ul \textbf{29.26}} & {\ul \textbf{18.49}} & 1.42 & 6.22 \\ 
\multirow{-4}{*}{\textbf{INSTRUCT}} & {\color[HTML]{FF0000} CPT+SFT} & 1.03 & 6.36 & 0.07 & 0.55 & 1.41 & {\ul \textbf{10.07}} \\ \hline
 & Base & 0.14 & 1.98 & 0.57 & 0.35 & 0.51 & \textbf{7.06} \\ 
 & {\color[HTML]{0000FF} CPT} & \textbf{2.11} & \textbf{24.01} & 12.20 & 7.16 & \textbf{1.12} & 5.69 \\ 
 & {\color[HTML]{6AA84F} SFT} & 1.12 & 20.40 & 17.49 & 9.30 & 0.40 & 5.50 \\ 
\multirow{-4}{*}{\textbf{MEDICAL}} & {\color[HTML]{FF0000} CPT+SFT} & 1.67 & 15.87 & \textbf{18.29} & \textbf{9.98} & 0.49 & 6.70 \\ \hline
\multicolumn{8}{c}{\cellcolor[HTML]{CFE2F3}\textit{\textbf{Llama-13B}}} \\ \hline
 & Base & 0.29 & 0.84 & 11.77 & 6.03 & 0.19 & 1.98 \\ 
 & {\color[HTML]{0000FF} CPT} & \textbf{2.81} & 21.03 & 10.49 & 6.65 & \textbf{1.85} & 7.85 \\ 
 & {\color[HTML]{6AA84F} SFT} & 2.18 & 17.08 & 18.67 & 10.42 & 0.42 & 5.85 \\ 
\multirow{-4}{*}{\textbf{GENERAL}} & {\color[HTML]{FF0000} CPT+SFT} & 1.65 & \textbf{21.56} & \textbf{19.84} & \textbf{10.74} & 0.59 & \textbf{8.07} \\ \hline
 & Base & 0.00 & 4.87 & 0.04 & 0.02 & 0.50 & 3.85 \\ 
 & {\color[HTML]{0000FF} CPT} & 4.82 & 34.09 & 0.04 & 2.43 & {\ul \textbf{2.09}} & {\ul \textbf{10.78}} \\ 
 & {\color[HTML]{6AA84F} SFT} & 10.92 & 42.98 & 30.13 & 20.53 & 1.52 & 6.52 \\ 
\multirow{-4}{*}{\textbf{INSTRUCT}} & {\color[HTML]{FF0000} CPT+SFT} & {\ul \textbf{12.57}} & {\ul \textbf{44.02}} & {\ul \textbf{31.48}} & {\ul \textbf{22.02}} & 1.74 & 7.32 \\ \hline
 & Base & 0.56 & 9.41 & 11.01 & 5.78 & 0.48 & 5.60 \\ 
 & {\color[HTML]{0000FF} CPT} & 2.02 & \textbf{22.25} & 11.28 & 6.65 & \textbf{1.12} & 5.69 \\ 
 & {\color[HTML]{6AA84F} SFT} & \textbf{2.07} & 21.50 & 18.11 & 10.09 & 0.59 & \textbf{7.37} \\ 
\multirow{-4}{*}{\textbf{MEDICAL}} & {\color[HTML]{FF0000} CPT+SFT} & 1.61 & 17.64 & \textbf{19.75} & \textbf{10.68} & 0.67 & 6.53 \\ \hline
\end{tabular}%
}
\caption{Greedy decoding results for MCQ and MCQU and BLEU/METEOR scores for OEQA across model families and adaptation strategies. \textbf{Bold} denotes the best strategy and {\ul underlining} the best initialization.}
\label{tab:main-result-greedy}
\end{table*}

%% file: significance-MCQA.tex
\begin{table*}[]
\centering
\resizebox{0.5\textwidth}{!}{%
\begin{tabular}{lllllllllll}
\hline
 & \textbf{id} & \textbf{strategy} & \textbf{model\_a} & \textbf{model\_b} & \textbf{Decoding type} & \textbf{ci95\_low} & \textbf{ci95\_high} & \textbf{p\_two\_sided} & \textbf{alpha\_Bonferoni} & \textbf{significant\_Bonferroni} \\ \hline
\multicolumn{11}{c}{\cellcolor[HTML]{CFE2F3}\textbf{Gemma 4B}} \\ \hline
 & \textbf{A1} & {\color[HTML]{0000FF} CPT} & gemma-3-4b-pt-CPT & gemma-3-4b-pt & constrained & -2.60E-02 & -3.48E-03 & 2.00E-04 & 4.17E-03 & \textbf{TRUE} \\  
 & \textbf{A1} & {\color[HTML]{0000FF} CPT} & gemma-3-4b-pt-CPT & gemma-3-4b-pt & greedy & -6.03E-03 & 2.71E-02 & 1.75E-01 & 4.17E-03 & FALSE \\  
 & \textbf{A2} & {\color[HTML]{CC0000} CPT+SFT} & gemma-3-4b-pt-CPT-SFT & gemma-3-4b-pt & constrained & 1.57E-02 & 5.29E-02 & 4.00E-04 & 4.17E-03 & \textbf{TRUE} \\  
 & \textbf{A2} & {\color[HTML]{CC0000} CPT+SFT} & gemma-3-4b-pt-CPT-SFT & gemma-3-4b-pt & greedy & 6.03E-02 & 9.37E-02 & 2.00E-04 & 4.17E-03 & \textbf{TRUE} \\  
 & \textbf{A3} & {\color[HTML]{CC0000} CPT+SFT} & gemma-3-4b-pt-CPT-SFT & gemma-3-4b-pt-CPT & constrained & 2.34E-02 & 7.52E-02 & 2.00E-04 & 4.17E-03 & \textbf{TRUE} \\  
 & \textbf{A3} & {\color[HTML]{CC0000} CPT+SFT} & gemma-3-4b-pt-CPT-SFT & gemma-3-4b-pt-CPT & greedy & 4.06E-02 & 9.63E-02 & 2.00E-04 & 4.17E-03 & \textbf{TRUE} \\  
 & \textbf{A4} & {\color[HTML]{38761D} SFT} & gemma-3-4b-pt-SFT & gemma-3-4b-pt & constrained & 1.32E-02 & 4.90E-02 & 2.00E-04 & 4.17E-03 & \textbf{TRUE} \\  
\multirow{-8}{*}{\textbf{GENERAL}} & \textbf{A4} & {\color[HTML]{38761D} SFT} & gemma-3-4b-pt-SFT & gemma-3-4b-pt & greedy & 6.17E-02 & 9.54E-02 & 2.00E-04 & 4.17E-03 & \textbf{TRUE} \\ \hline
 & \textbf{B1} & {\color[HTML]{0000FF} CPT} & gemma-3-4b-it-CPT & gemma-3-4b-it & constrained & -4.69E-02 & -8.61E-03 & 4.00E-03 & 4.17E-03 & \textbf{TRUE} \\  
 & \textbf{B1} & {\color[HTML]{0000FF} CPT} & gemma-3-4b-it-CPT & gemma-3-4b-it & greedy & -1.94E-01 & -1.47E-01 & 2.00E-04 & 4.17E-03 & \textbf{TRUE} \\  
 & \textbf{B2} & {\color[HTML]{CC0000} CPT+SFT} & gemma-3-4b-it-CPT-SFT & gemma-3-4b-it & constrained & -1.25E-02 & 1.41E-02 & 9.27E-01 & 4.17E-03 & FALSE \\  
 & \textbf{B2} & {\color[HTML]{CC0000} CPT+SFT} & gemma-3-4b-it-CPT-SFT & gemma-3-4b-it & greedy & -3.54E-02 & -2.16E-03 & 2.66E-02 & 4.17E-03 & FALSE \\  
 & \textbf{B3} & {\color[HTML]{CC0000} CPT+SFT} & gemma-3-4b-it-CPT-SFT & gemma-3-4b-it-CPT & constrained & 1.49E-02 & 4.29E-02 & 2.00E-04 & 4.17E-03 & \textbf{TRUE} \\  
 & \textbf{B3} & {\color[HTML]{CC0000} CPT+SFT} & gemma-3-4b-it-CPT-SFT & gemma-3-4b-it-CPT & greedy & 1.26E-01 & 1.78E-01 & 2.00E-04 & 4.17E-03 & \textbf{TRUE} \\  
 & \textbf{B4} & {\color[HTML]{38761D} SFT} & gemma-3-4b-it-SFT & gemma-3-4b-it & constrained & -5.48E-03 & 2.13E-02 & 2.77E-01 & 4.17E-03 & FALSE \\  
\multirow{-8}{*}{\textbf{INSTRUCT}} & \textbf{B4} & {\color[HTML]{38761D} SFT} & gemma-3-4b-it-SFT & gemma-3-4b-it & greedy & -2.62E-02 & 3.58E-03 & 1.23E-01 & 4.17E-03 & FALSE \\ \hline
 & \textbf{C1} & {\color[HTML]{0000FF} CPT} & medgemma-4b-pt-CPT & medgemma-4b-pt & constrained & -1.68E-02 & 1.67E-04 & 5.60E-02 & 4.17E-03 & FALSE \\  
 & \textbf{C1} & {\color[HTML]{0000FF} CPT} & medgemma-4b-pt-CPT & medgemma-4b-pt & greedy & -4.32E-03 & 9.75E-03 & 4.60E-01 & 4.17E-03 & FALSE \\  
 & \textbf{C2} & {\color[HTML]{CC0000} CPT+SFT} & medgemma-4b-pt-CPT-SFT & medgemma-4b-pt-CPT & constrained & 2.64E-02 & 4.73E-02 & 2.00E-04 & 4.17E-03 & \textbf{TRUE} \\  
 & \textbf{C2} & {\color[HTML]{CC0000} CPT+SFT} & medgemma-4b-pt-CPT-SFT & medgemma-4b-pt-CPT & greedy & 2.44E-02 & 3.98E-02 & 2.00E-04 & 4.17E-03 & \textbf{TRUE} \\  
 & \textbf{C3} & {\color[HTML]{CC0000} CPT+SFT} & medgemma-4b-pt-CPT-SFT & medgemma-4b-pt & constrained & 1.72E-02 & 4.13E-02 & 2.00E-04 & 4.17E-03 & \textbf{TRUE} \\  
 & \textbf{C3} & {\color[HTML]{CC0000} CPT+SFT} & medgemma-4b-pt-CPT-SFT & medgemma-4b-pt & greedy & 2.80E-02 & 4.25E-02 & 2.00E-04 & 4.17E-03 & \textbf{TRUE} \\  
 & \textbf{C4} & {\color[HTML]{38761D} SFT} & medgemma-4b-pt-SFT & medgemma-4b-pt & constrained & 1.74E-02 & 4.04E-02 & 2.00E-04 & 4.17E-03 & \textbf{TRUE} \\  
\multirow{-8}{*}{\textbf{MEDICAL}} & \textbf{C4} & {\color[HTML]{38761D} SFT} & medgemma-4b-pt-SFT & medgemma-4b-pt & greedy & 2.58E-02 & 4.14E-02 & 2.00E-04 & 4.17E-03 & \textbf{TRUE} \\ \hline
 & \textbf{D1} & {\color[HTML]{38761D} SFT} & gemma-3-4b-pt-SFT & gemma-3-4b-it-SFT & constrained & -7.13E-03 & 1.15E-02 & 6.83E-01 & 5.56E-03 & FALSE \\  
 & \textbf{D1} & {\color[HTML]{38761D} SFT} & gemma-3-4b-pt-SFT & gemma-3-4b-it-SFT & greedy & -6.92E-02 & -4.32E-02 & 2.00E-04 & 5.56E-03 & \textbf{TRUE} \\  
 & \textbf{D2} & {\color[HTML]{38761D} SFT} & gemma-3-4b-pt-SFT & medgemma-4b-pt-SFT & constrained & 2.60E-03 & 1.27E-02 & 4.80E-03 & 5.56E-03 & \textbf{TRUE} \\  
 & \textbf{D2} & {\color[HTML]{38761D} SFT} & gemma-3-4b-pt-SFT & medgemma-4b-pt-SFT & greedy & 8.83E-03 & 1.61E-02 & 2.00E-04 & 5.56E-03 & \textbf{TRUE} \\  
 & \textbf{D3} & {\color[HTML]{38761D} SFT} & gemma-3-4b-it-SFT & medgemma-4b-pt-SFT & constrained & -5.57E-03 & 1.57E-02 & 3.13E-01 & 5.56E-03 & FALSE \\  
\multirow{-6}{*}{\textbf{SFT}} & \textbf{D3} & {\color[HTML]{38761D} SFT} & gemma-3-4b-it-SFT & medgemma-4b-pt-SFT & greedy & 5.63E-02 & 8.12E-02 & 2.00E-04 & 5.56E-03 & \textbf{TRUE} \\ \hline
 & \textbf{E1} & {\color[HTML]{0000FF} CPT} & gemma-3-4b-pt-CPT & gemma-3-4b-it-CPT & constrained & -3.76E-02 & 2.01E-02 & 7.90E-01 & 5.56E-03 & FALSE \\  
 & \textbf{E1} & {\color[HTML]{0000FF} CPT} & gemma-3-4b-pt-CPT & gemma-3-4b-it-CPT & greedy & 2.58E-02 & 4.95E-02 & 2.00E-04 & 5.56E-03 & \textbf{TRUE} \\  
 & \textbf{E2} & {\color[HTML]{0000FF} CPT} & gemma-3-4b-pt-CPT & medgemma-4b-pt-CPT & constrained & -2.87E-02 & 2.19E-02 & 8.35E-01 & 5.56E-03 & FALSE \\  
 & \textbf{E2} & {\color[HTML]{0000FF} CPT} & gemma-3-4b-pt-CPT & medgemma-4b-pt-CPT & greedy & -5.66E-02 & 1.81E-03 & 8.56E-02 & 5.56E-03 & FALSE \\  
 & \textbf{E3} & {\color[HTML]{0000FF} CPT} & gemma-3-4b-it-CPT & medgemma-4b-pt-CPT & constrained & -2.21E-03 & 1.59E-02 & 1.83E-01 & 5.56E-03 & FALSE \\  
\multirow{-6}{*}{\textbf{CPT}} & \textbf{E3} & {\color[HTML]{0000FF} CPT} & gemma-3-4b-it-CPT & medgemma-4b-pt-CPT & greedy & -9.23E-02 & -3.78E-02 & 2.00E-04 & 5.56E-03 & \textbf{TRUE} \\ \hline
 & \textbf{F1} & {\color[HTML]{CC0000} CPT+SFT} & gemma-3-4b-pt-CPT-SFT & gemma-3-4b-it-CPT-SFT & constrained & 2.15E-03 & 2.27E-02 & 1.92E-02 & 5.56E-03 & FALSE \\  
 & \textbf{F1} & {\color[HTML]{CC0000} CPT+SFT} & gemma-3-4b-pt-CPT-SFT & gemma-3-4b-it-CPT-SFT & greedy & -6.81E-02 & -3.28E-02 & 2.00E-04 & 5.56E-03 & \textbf{TRUE} \\  
 & \textbf{F2} & {\color[HTML]{CC0000} CPT+SFT} & gemma-3-4b-pt-CPT-SFT & medgemma-4b-pt-CPT-SFT & constrained & 2.64E-03 & 1.80E-02 & 1.10E-02 & 5.56E-03 & FALSE \\  
 & \textbf{F2} & {\color[HTML]{CC0000} CPT+SFT} & gemma-3-4b-pt-CPT-SFT & medgemma-4b-pt-CPT-SFT & greedy & 3.47E-03 & 1.52E-02 & 1.60E-03 & 5.56E-03 & \textbf{TRUE} \\  
 & \textbf{F3} & {\color[HTML]{CC0000} CPT+SFT} & gemma-3-4b-it-CPT-SFT & medgemma-4b-pt-CPT-SFT & constrained & -1.27E-02 & 9.11E-03 & 7.58E-01 & 5.56E-03 & FALSE \\  
\multirow{-6}{*}{\textbf{CPT+SFT}} & \textbf{F3} & {\color[HTML]{CC0000} CPT+SFT} & gemma-3-4b-it-CPT-SFT & medgemma-4b-pt-CPT-SFT & greedy & 3.84E-02 & 7.95E-02 & 2.00E-04 & 5.56E-03 & \textbf{TRUE} \\ \hline
\multicolumn{11}{c}{\cellcolor[HTML]{CFE2F3}\textbf{Mistral 7B}} \\ \hline
 & \textbf{A1} & {\color[HTML]{0000FF} CPT} & Mistral-7B-v0.1-CPT & Mistral-7B-v0.1 & constrained & 1.63E-03 & 2.19E-02 & 2.26E-02 & 4.17E-03 & FALSE \\  
 & \textbf{A1} & {\color[HTML]{0000FF} CPT} & Mistral-7B-v0.1-CPT & Mistral-7B-v0.1 & greedy & 3.64E-02 & 6.00E-02 & 2.00E-04 & 4.17E-03 & \textbf{TRUE} \\  
 & \textbf{A2} & {\color[HTML]{CC0000} CPT+SFT} & Mistral-7B-v0.1-CPT-SFT & Mistral-7B-v0.1 & constrained & 3.71E-02 & 6.85E-02 & 2.00E-04 & 4.17E-03 & \textbf{TRUE} \\  
 & \textbf{A2} & {\color[HTML]{CC0000} CPT+SFT} & Mistral-7B-v0.1-CPT-SFT & Mistral-7B-v0.1 & greedy & 6.72E-02 & 1.01E-01 & 2.00E-04 & 4.17E-03 & \textbf{TRUE} \\  
 & \textbf{A3} & {\color[HTML]{CC0000} CPT+SFT} & Mistral-7B-v0.1-CPT-SFT & Mistral-7B-v0.1-CPT & constrained & 2.80E-02 & 5.41E-02 & 2.00E-04 & 4.17E-03 & \textbf{TRUE} \\  
 & \textbf{A3} & {\color[HTML]{CC0000} CPT+SFT} & Mistral-7B-v0.1-CPT-SFT & Mistral-7B-v0.1-CPT & greedy & 2.82E-02 & 4.38E-02 & 2.00E-04 & 4.17E-03 & \textbf{TRUE} \\  
 & \textbf{A4} & {\color[HTML]{38761D} SFT} & Mistral-7B-v0.1-SFT & Mistral-7B-v0.1 & constrained & 3.25E-02 & 6.88E-02 & 2.00E-04 & 4.17E-03 & \textbf{TRUE} \\  
\multirow{-8}{*}{\textbf{GENERAL}} & \textbf{A4} & {\color[HTML]{38761D} SFT} & Mistral-7B-v0.1-SFT & Mistral-7B-v0.1 & greedy & 7.41E-02 & 1.06E-01 & 2.00E-04 & 4.17E-03 & \textbf{TRUE} \\ \hline
 & \textbf{B1} & {\color[HTML]{0000FF} CPT} & Mistral-7B-Instruct-v0.1-CPT & Mistral-7B-Instruct-v0.1 & constrained & 1.76E-02 & 4.88E-02 & 2.00E-04 & 4.17E-03 & \textbf{TRUE} \\  
 & \textbf{B1} & {\color[HTML]{0000FF} CPT} & Mistral-7B-Instruct-v0.1-CPT & Mistral-7B-Instruct-v0.1 & greedy & 1.67E-02 & 5.01E-02 & 2.00E-04 & 4.17E-03 & \textbf{TRUE} \\  
 & \textbf{B2} & {\color[HTML]{CC0000} CPT+SFT} & Mistral-7B-Instruct-v0.1-CPT-SFT & Mistral-7B-Instruct-v0.1-CPT & constrained & -2.10E-02 & 1.47E-02 & 7.64E-01 & 4.17E-03 & FALSE \\  
 & \textbf{B2} & {\color[HTML]{CC0000} CPT+SFT} & Mistral-7B-Instruct-v0.1-CPT-SFT & Mistral-7B-Instruct-v0.1-CPT & greedy & -1.59E-02 & 1.76E-02 & 9.83E-01 & 4.17E-03 & FALSE \\  
 & \textbf{B3} & {\color[HTML]{CC0000} CPT+SFT} & Mistral-7B-Instruct-v0.1-CPT-SFT & Mistral-7B-Instruct-v0.1 & constrained & 1.29E-02 & 5.00E-02 & 2.00E-04 & 4.17E-03 & \textbf{TRUE} \\  
 & \textbf{B3} & {\color[HTML]{CC0000} CPT+SFT} & Mistral-7B-Instruct-v0.1-CPT-SFT & Mistral-7B-Instruct-v0.1 & greedy & 1.68E-02 & 5.21E-02 & 2.00E-04 & 4.17E-03 & \textbf{TRUE} \\  
 & \textbf{B4} & {\color[HTML]{38761D} SFT} & Mistral-7B-Instruct-v0.1-SFT & Mistral-7B-Instruct-v0.1 & constrained & 2.41E-02 & 6.41E-02 & 2.00E-04 & 4.17E-03 & \textbf{TRUE} \\  
\multirow{-8}{*}{\textbf{INSTRUCT}} & \textbf{B4} & {\color[HTML]{38761D} SFT} & Mistral-7B-Instruct-v0.1-SFT & Mistral-7B-Instruct-v0.1 & greedy & 6.71E-02 & 1.18E-01 & 2.00E-04 & 4.17E-03 & \textbf{TRUE} \\ \hline
 & \textbf{C1} & {\color[HTML]{0000FF} CPT} & BioMistral-7B-CPT & BioMistral-7B & constrained & -1.12E-02 & 1.00E-02 & 8.13E-01 & 4.17E-03 & FALSE \\  
 & \textbf{C1} & {\color[HTML]{0000FF} CPT} & BioMistral-7B-CPT & BioMistral-7B & greedy & -1.61E-02 & 2.51E-03 & 2.85E-01 & 4.17E-03 & FALSE \\  
 & \textbf{C2} & {\color[HTML]{CC0000} CPT+SFT} & BioMistral-7B-CPT-SFT & BioMistral-7B & constrained & 2.63E-02 & 6.50E-02 & 2.00E-04 & 4.17E-03 & \textbf{TRUE} \\  
 & \textbf{C2} & {\color[HTML]{CC0000} CPT+SFT} & BioMistral-7B-CPT-SFT & BioMistral-7B & greedy & 1.33E-02 & 4.54E-02 & 2.00E-04 & 4.17E-03 & \textbf{TRUE} \\  
 & \textbf{C3} & {\color[HTML]{CC0000} CPT+SFT} & BioMistral-7B-CPT-SFT & BioMistral-7B-CPT & constrained & 2.23E-02 & 7.24E-02 & 2.00E-04 & 4.17E-03 & \textbf{TRUE} \\  
 & \textbf{C3} & {\color[HTML]{CC0000} CPT+SFT} & BioMistral-7B-CPT-SFT & BioMistral-7B-CPT & greedy & 1.27E-02 & 6.02E-02 & 2.00E-04 & 4.17E-03 & \textbf{TRUE} \\  
 & \textbf{C4} & {\color[HTML]{38761D} SFT} & BioMistral-7B-SFT & BioMistral-7B & constrained & 1.33E-02 & 4.67E-02 & 2.00E-04 & 4.17E-03 & \textbf{TRUE} \\  
\multirow{-8}{*}{\textbf{MEDICAL}} & \textbf{C4} & {\color[HTML]{38761D} SFT} & BioMistral-7B-SFT & BioMistral-7B & greedy & 1.34E-02 & 4.18E-02 & 2.00E-04 & 4.17E-03 & \textbf{TRUE} \\ \hline
 & \textbf{D1} & {\color[HTML]{38761D} SFT} & Mistral-7B-v0.1-SFT & Mistral-7B-Instruct-v0.1-SFT & constrained & -1.89E-02 & 2.68E-02 & 7.28E-01 & 5.56E-03 & FALSE \\  
 & \textbf{D1} & {\color[HTML]{38761D} SFT} & Mistral-7B-v0.1-SFT & Mistral-7B-Instruct-v0.1-SFT & greedy & -1.32E-01 & -8.16E-02 & 2.00E-04 & 5.56E-03 & \textbf{TRUE} \\  
 & \textbf{D2} & {\color[HTML]{38761D} SFT} & Mistral-7B-Instruct-v0.1-SFT & BioMistral-7B-SFT & constrained & -1.41E-02 & 4.27E-02 & 4.35E-01 & 5.56E-03 & FALSE \\  
 & \textbf{D2} & {\color[HTML]{38761D} SFT} & Mistral-7B-Instruct-v0.1-SFT & BioMistral-7B-SFT & greedy & 8.72E-02 & 1.40E-01 & 2.00E-04 & 5.56E-03 & \textbf{TRUE} \\  
 & \textbf{D3} & {\color[HTML]{38761D} SFT} & Mistral-7B-v0.1-SFT & BioMistral-7B-SFT & constrained & 5.20E-03 & 2.93E-02 & 2.00E-03 & 5.56E-03 & \textbf{TRUE} \\  
\multirow{-6}{*}{\textbf{SFT}} & \textbf{D3} & {\color[HTML]{38761D} SFT} & Mistral-7B-v0.1-SFT & BioMistral-7B-SFT & greedy & 1.57E-03 & 1.27E-02 & 1.66E-02 & 5.56E-03 & FALSE \\ \hline
 & \textbf{E1} & {\color[HTML]{0000FF} CPT} & Mistral-7B-v0.1-CPT & Mistral-7B-Instruct-v0.1-CPT & constrained & -5.11E-02 & -3.56E-04 & 4.62E-02 & 5.56E-03 & FALSE \\  
 & \textbf{E1} & {\color[HTML]{0000FF} CPT} & Mistral-7B-v0.1-CPT & Mistral-7B-Instruct-v0.1-CPT & greedy & -1.07E-01 & -6.53E-02 & 2.00E-04 & 5.56E-03 & \textbf{TRUE} \\  
 & \textbf{E2} & {\color[HTML]{0000FF} CPT} & Mistral-7B-v0.1-CPT & BioMistral-7B-CPT & constrained & -9.26E-03 & 2.57E-02 & 6.18E-01 & 5.56E-03 & FALSE \\  
 & \textbf{E2} & {\color[HTML]{0000FF} CPT} & Mistral-7B-v0.1-CPT & BioMistral-7B-CPT & greedy & -1.44E-02 & 1.70E-02 & 6.45E-01 & 5.56E-03 & FALSE \\  
 & \textbf{E3} & {\color[HTML]{0000FF} CPT} & Mistral-7B-Instruct-v0.1-CPT & BioMistral-7B-CPT & constrained & 4.02E-03 & 5.94E-02 & 1.80E-02 & 5.56E-03 & FALSE \\  
\multirow{-6}{*}{\textbf{CPT}} & \textbf{E3} & {\color[HTML]{0000FF} CPT} & Mistral-7B-Instruct-v0.1-CPT & BioMistral-7B-CPT & greedy & 6.80E-02 & 1.01E-01 & 2.00E-04 & 5.56E-03 & \textbf{TRUE} \\ \hline
 & \textbf{F1} & {\color[HTML]{CC0000} CPT+SFT} & Mistral-7B-v0.1-CPT-SFT & Mistral-7B-Instruct-v0.1-CPT-SFT & constrained & 7.69E-03 & 3.25E-02 & 1.20E-03 & 5.56E-03 & \textbf{TRUE} \\  
 & \textbf{F1} & {\color[HTML]{CC0000} CPT+SFT} & Mistral-7B-v0.1-CPT-SFT & Mistral-7B-Instruct-v0.1-CPT-SFT & greedy & -6.20E-02 & -4.21E-02 & 2.00E-04 & 5.56E-03 & \textbf{TRUE} \\  
 & \textbf{F2} & {\color[HTML]{CC0000} CPT+SFT} & Mistral-7B-v0.1-CPT-SFT & BioMistral-7B-CPT-SFT & constrained & -9.97E-03 & 1.59E-02 & 5.02E-01 & 5.56E-03 & FALSE \\  
 & \textbf{F2} & {\color[HTML]{CC0000} CPT+SFT} & Mistral-7B-v0.1-CPT-SFT & BioMistral-7B-CPT-SFT & greedy & -1.08E-02 & 9.42E-03 & 6.69E-01 & 5.56E-03 & FALSE \\  
 & \textbf{F3} & {\color[HTML]{CC0000} CPT+SFT} & Mistral-7B-Instruct-v0.1-CPT-SFT & BioMistral-7B-CPT-SFT & constrained & -3.06E-02 & -1.28E-03 & 3.42E-02 & 5.56E-03 & FALSE \\  
\multirow{-6}{*}{\textbf{CPT+SFT}} & \textbf{F3} & {\color[HTML]{CC0000} CPT+SFT} & Mistral-7B-Instruct-v0.1-CPT-SFT & BioMistral-7B-CPT-SFT & greedy & 3.56E-02 & 6.81E-02 & 2.00E-04 & 5.56E-03 & \textbf{TRUE} \\ \hline
\multicolumn{11}{c}{\cellcolor[HTML]{CFE2F3}\textbf{Llama 7B}} \\ \hline
 & \textbf{A1} & {\color[HTML]{0000FF} CPT} & Llama-2-7b-hf-CPT & Llama-2-7b-hf & constrained & -8.99E-03 & 4.70E-03 & 6.17E-01 & 4.17E-03 & FALSE \\  
 & \textbf{A1} & {\color[HTML]{0000FF} CPT} & Llama-2-7b-hf-CPT & Llama-2-7b-hf & greedy & -2.29E-02 & -4.41E-03 & 3.00E-03 & 4.17E-03 & \textbf{TRUE} \\  
 & \textbf{A2} & {\color[HTML]{CC0000} CPT+SFT} & LLama-2-7b-hf-CPT-SFT & Llama-2-7b-hf-CPT & constrained & 2.06E-02 & 4.76E-02 & 2.00E-04 & 4.17E-03 & \textbf{TRUE} \\  
 & \textbf{A2} & {\color[HTML]{CC0000} CPT+SFT} & LLama-2-7b-hf-CPT-SFT & Llama-2-7b-hf-CPT & greedy & 4.28E-02 & 8.52E-02 & 2.00E-04 & 4.17E-03 & \textbf{TRUE} \\  
 & \textbf{A3} & {\color[HTML]{CC0000} CPT+SFT} & LLama-2-7b-hf-CPT-SFT & Llama-2-7b-hf & constrained & 1.81E-02 & 4.46E-02 & 2.00E-04 & 4.17E-03 & \textbf{TRUE} \\  
 & \textbf{A3} & {\color[HTML]{CC0000} CPT+SFT} & LLama-2-7b-hf-CPT-SFT & Llama-2-7b-hf & greedy & 3.37E-02 & 6.48E-02 & 2.00E-04 & 4.17E-03 & \textbf{TRUE} \\  
 & \textbf{A4} & {\color[HTML]{38761D} SFT} & LLama-2-7b-hf-SFT & Llama-2-7b-hf & constrained & 1.58E-02 & 3.86E-02 & 2.00E-04 & 4.17E-03 & \textbf{TRUE} \\  
\multirow{-8}{*}{\textbf{GENERAL}} & \textbf{A4} & {\color[HTML]{38761D} SFT} & LLama-2-7b-hf-SFT & Llama-2-7b-hf & greedy & 3.08E-02 & 6.12E-02 & 2.00E-04 & 4.17E-03 & \textbf{TRUE} \\ \hline
 & \textbf{B1} & {\color[HTML]{0000FF} CPT} & Llama-2-7b-chat-hf-CPT & Llama-2-7b-chat-hf & constrained & -7.23E-03 & 4.04E-03 & 8.20E-01 & 4.17E-03 & FALSE \\  
 & \textbf{B1} & {\color[HTML]{0000FF} CPT} & Llama-2-7b-chat-hf-CPT & Llama-2-7b-chat-hf & greedy & 1.86E-02 & 3.22E-02 & 2.00E-04 & 4.17E-03 & \textbf{TRUE} \\  
 & \textbf{B2} & {\color[HTML]{CC0000} CPT+SFT} & Llama-2-7b-chat-hf-CPT-SFT & Llama-2-7b-chat-hf & constrained & -1.32E-02 & 2.43E-03 & 2.78E-01 & 4.17E-03 & FALSE \\  
 & \textbf{B2} & {\color[HTML]{CC0000} CPT+SFT} & Llama-2-7b-chat-hf-CPT-SFT & Llama-2-7b-chat-hf & greedy & 2.64E-03 & 7.96E-03 & 2.00E-04 & 4.17E-03 & \textbf{TRUE} \\  
 & \textbf{B3} & {\color[HTML]{CC0000} CPT+SFT} & Llama-2-7b-chat-hf-CPT-SFT & Llama-2-7b-chat-hf-CPT & constrained & -7.62E-03 & 2.54E-04 & 7.10E-02 & 4.17E-03 & FALSE \\  
 & \textbf{B3} & {\color[HTML]{CC0000} CPT+SFT} & Llama-2-7b-chat-hf-CPT-SFT & Llama-2-7b-chat-hf-CPT & greedy & -2.69E-02 & -1.17E-02 & 2.00E-04 & 4.17E-03 & \textbf{TRUE} \\  
 & \textbf{B4} & {\color[HTML]{38761D} SFT} & Llama-2-7b-chat-hf-SFT & Llama-2-7b-chat-hf & constrained & 9.28E-03 & 4.49E-02 & 2.00E-03 & 4.17E-03 & \textbf{TRUE} \\  
\multirow{-8}{*}{\textbf{INSTRUCT}} & \textbf{B4} & {\color[HTML]{38761D} SFT} & Llama-2-7b-chat-hf-SFT & Llama-2-7b-chat-hf & greedy & 1.54E-01 & 2.19E-01 & 2.00E-04 & 4.17E-03 & \textbf{TRUE} \\ \hline
 & \textbf{C1} & {\color[HTML]{0000FF} CPT} & meditron-7b-CPT & meditron-7b & constrained & 1.19E-02 & 2.70E-02 & 2.00E-04 & 4.17E-03 & \textbf{TRUE} \\  
 & \textbf{C1} & {\color[HTML]{0000FF} CPT} & meditron-7b-CPT & meditron-7b & greedy & 4.73E-02 & 1.05E-01 & 2.00E-04 & 4.17E-03 & \textbf{TRUE} \\  
 & \textbf{C2} & {\color[HTML]{CC0000} CPT+SFT} & meditron-7b-CPT-SFT & meditron-7b & constrained & 3.78E-02 & 7.33E-02 & 2.00E-04 & 4.17E-03 & \textbf{TRUE} \\  
 & \textbf{C2} & {\color[HTML]{CC0000} CPT+SFT} & meditron-7b-CPT-SFT & meditron-7b & greedy & 6.80E-02 & 1.47E-01 & 2.00E-04 & 4.17E-03 & \textbf{TRUE} \\  
 & \textbf{C3} & {\color[HTML]{CC0000} CPT+SFT} & meditron-7b-CPT-SFT & meditron-7b-CPT & constrained & 2.20E-02 & 5.08E-02 & 2.00E-04 & 4.17E-03 & \textbf{TRUE} \\  
 & \textbf{C3} & {\color[HTML]{CC0000} CPT+SFT} & meditron-7b-CPT-SFT & meditron-7b-CPT & greedy & 1.87E-02 & 4.21E-02 & 2.00E-04 & 4.17E-03 & \textbf{TRUE} \\  
 & \textbf{C4} & {\color[HTML]{38761D} SFT} & meditron-7b-SFT & meditron-7b & constrained & 3.31E-02 & 6.48E-02 & 2.00E-04 & 4.17E-03 & \textbf{TRUE} \\  
\multirow{-8}{*}{\textbf{MEDICAL}} & \textbf{C4} & {\color[HTML]{38761D} SFT} & meditron-7b-SFT & meditron-7b & greedy & 6.23E-02 & 1.36E-01 & 2.00E-04 & 4.17E-03 & \textbf{TRUE} \\ \hline
 & \textbf{D1} & {\color[HTML]{38761D} SFT} & LLama-2-7b-hf-SFT & Llama-2-7b-chat-hf-SFT & constrained & -3.69E-02 & 1.03E-02 & 4.19E-01 & 5.56E-03 & FALSE \\  
 & \textbf{D1} & {\color[HTML]{38761D} SFT} & LLama-2-7b-hf-SFT & Llama-2-7b-chat-hf-SFT & greedy & -1.22E-01 & -6.88E-02 & 2.00E-04 & 5.56E-03 & \textbf{TRUE} \\  
 & \textbf{D2} & {\color[HTML]{38761D} SFT} & LLama-2-7b-hf-SFT & meditron-7b-SFT & constrained & -1.39E-02 & -3.59E-03 & 2.00E-04 & 5.56E-03 & \textbf{TRUE} \\  
 & \textbf{D2} & {\color[HTML]{38761D} SFT} & LLama-2-7b-hf-SFT & meditron-7b-SFT & greedy & -8.30E-03 & 1.95E-03 & 2.48E-01 & 5.56E-03 & FALSE \\  
 & \textbf{D3} & {\color[HTML]{38761D} SFT} & Llama-2-7b-chat-hf-SFT & meditron-7b-SFT & constrained & -1.92E-02 & 2.74E-02 & 9.14E-01 & 5.56E-03 & FALSE \\  
\multirow{-6}{*}{\textbf{SFT}} & \textbf{D3} & {\color[HTML]{38761D} SFT} & Llama-2-7b-chat-hf-SFT & meditron-7b-SFT & greedy & 6.34E-02 & 1.18E-01 & 2.00E-04 & 5.56E-03 & \textbf{TRUE} \\ \hline
 & \textbf{E1} & {\color[HTML]{0000FF} CPT} & Llama-2-7b-hf-CPT & Llama-2-7b-chat-hf-CPT & constrained & -2.48E-02 & 1.31E-03 & 8.28E-02 & 5.56E-03 & FALSE \\  
 & \textbf{E1} & {\color[HTML]{0000FF} CPT} & Llama-2-7b-hf-CPT & Llama-2-7b-chat-hf-CPT & greedy & -3.69E-03 & 3.09E-02 & 1.93E-01 & 5.56E-03 & FALSE \\  
 & \textbf{E2} & {\color[HTML]{0000FF} CPT} & Llama-2-7b-hf-CPT & meditron-7b-CPT & constrained & -1.69E-02 & -1.09E-03 & 2.40E-02 & 5.56E-03 & FALSE \\  
 & \textbf{E2} & {\color[HTML]{0000FF} CPT} & Llama-2-7b-hf-CPT & meditron-7b-CPT & greedy & -5.63E-02 & -2.58E-02 & 2.00E-04 & 5.56E-03 & \textbf{TRUE} \\  
 & \textbf{E3} & {\color[HTML]{0000FF} CPT} & Llama-2-7b-chat-hf-CPT & meditron-7b-CPT & constrained & -6.40E-03 & 1.11E-02 & 6.02E-01 & 5.56E-03 & FALSE \\  
\multirow{-6}{*}{\textbf{CPT}} & \textbf{E3} & {\color[HTML]{0000FF} CPT} & Llama-2-7b-chat-hf-CPT & meditron-7b-CPT & greedy & -8.45E-02 & -2.63E-02 & 2.00E-04 & 5.56E-03 & \textbf{TRUE} \\ \hline
 & \textbf{F1} & {\color[HTML]{CC0000} CPT+SFT} & LLama-2-7b-hf-CPT-SFT & Llama-2-7b-chat-hf-CPT-SFT & constrained & 1.27E-02 & 3.61E-02 & 4.00E-04 & 5.56E-03 & \textbf{TRUE} \\  
 & \textbf{F1} & {\color[HTML]{CC0000} CPT+SFT} & LLama-2-7b-hf-CPT-SFT & Llama-2-7b-chat-hf-CPT-SFT & greedy & 6.09E-02 & 1.32E-01 & 2.00E-04 & 5.56E-03 & \textbf{TRUE} \\  
 & \textbf{F2} & {\color[HTML]{CC0000} CPT+SFT} & LLama-2-7b-hf-CPT-SFT & meditron-7b-CPT-SFT & constrained & -1.83E-02 & -4.73E-03 & 1.00E-03 & 5.56E-03 & \textbf{TRUE} \\  
 & \textbf{F2} & {\color[HTML]{CC0000} CPT+SFT} & LLama-2-7b-hf-CPT-SFT & meditron-7b-CPT-SFT & greedy & -1.38E-02 & -1.53E-03 & 8.80E-03 & 5.56E-03 & FALSE \\  
 & \textbf{F3} & {\color[HTML]{CC0000} CPT+SFT} & Llama-2-7b-chat-hf-CPT-SFT & meditron-7b-CPT-SFT & constrained & -5.07E-02 & -2.24E-02 & 2.00E-04 & 5.56E-03 & \textbf{TRUE} \\  
\multirow{-6}{*}{\textbf{CPT+SFT}} & \textbf{F3} & {\color[HTML]{CC0000} CPT+SFT} & Llama-2-7b-chat-hf-CPT-SFT & meditron-7b-CPT-SFT & greedy & -1.44E-01 & -6.56E-02 & 2.00E-04 & 5.56E-03 & \textbf{TRUE} \\ \hline
\multicolumn{11}{c}{\cellcolor[HTML]{CFE2F3}\textbf{Llama 13B}} \\ \hline
 & \textbf{A1} & {\color[HTML]{0000FF} CPT} & Llama-2-13b-hf-CPT & Llama-2-13b-hf & constrained & 2.54E-03 & 2.02E-02 & 1.06E-02 & 4.17E-03 & FALSE \\  
 & \textbf{A1} & {\color[HTML]{0000FF} CPT} & Llama-2-13b-hf-CPT & Llama-2-13b-hf & greedy & 3.67E-03 & 1.98E-02 & 4.40E-03 & 4.17E-03 & FALSE \\  
 & \textbf{A2} & {\color[HTML]{CC0000} CPT+SFT} & Llama-2-13b-hf-CPT-SFT & Llama-2-13b-hf & constrained & 2.37E-02 & 5.45E-02 & 2.00E-04 & 4.17E-03 & \textbf{TRUE} \\  
 & \textbf{A2} & {\color[HTML]{CC0000} CPT+SFT} & Llama-2-13b-hf-CPT-SFT & Llama-2-13b-hf & greedy & 3.27E-02 & 5.90E-02 & 2.00E-04 & 4.17E-03 & \textbf{TRUE} \\  
 & \textbf{A3} & {\color[HTML]{CC0000} CPT+SFT} & Llama-2-13b-hf-CPT-SFT & Llama-2-13b-hf-CPT & constrained & 1.49E-02 & 3.87E-02 & 2.00E-04 & 4.17E-03 & \textbf{TRUE} \\  
 & \textbf{A3} & {\color[HTML]{CC0000} CPT+SFT} & Llama-2-13b-hf-CPT-SFT & Llama-2-13b-hf-CPT & greedy & 2.07E-02 & 4.80E-02 & 2.00E-04 & 4.17E-03 & \textbf{TRUE} \\  
 & \textbf{A4} & {\color[HTML]{CC0000} SFT} & Llama-2-13b-hf-SFT & Llama-2-13b-hf & constrained & 1.81E-02 & 4.28E-02 & 2.00E-04 & 4.17E-03 & \textbf{TRUE} \\  
\multirow{-8}{*}{\textbf{GENERAL}} & \textbf{A4} & {\color[HTML]{CC0000} SFT} & Llama-2-13b-hf-SFT & Llama-2-13b-hf & greedy & 3.19E-02 & 5.31E-02 & 2.00E-04 & 4.17E-03 & \textbf{TRUE} \\ \hline
 & \textbf{B1} & {\color[HTML]{0000FF} CPT} & Llama-2-13b-chat-hf-CPT & Llama-2-13b-chat-hf & constrained & -1.18E-03 & 4.97E-02 & 6.72E-02 & 4.17E-03 & FALSE \\  
 & \textbf{B1} & {\color[HTML]{0000FF} CPT} & Llama-2-13b-chat-hf-CPT & Llama-2-13b-chat-hf & greedy & 1.92E-02 & 3.14E-02 & 2.00E-04 & 4.17E-03 & \textbf{TRUE} \\  
 & \textbf{B2} & {\color[HTML]{CC0000} CPT+SFT} & Llama-2-13b-chat-hf-CPT-SFT & Llama-2-13b-chat-hf & constrained & 3.87E-02 & 9.30E-02 & 2.00E-04 & 4.17E-03 & \textbf{TRUE} \\  
 & \textbf{B2} & {\color[HTML]{CC0000} CPT+SFT} & Llama-2-13b-chat-hf-CPT-SFT & Llama-2-13b-chat-hf & greedy & 1.82E-01 & 2.56E-01 & 2.00E-04 & 4.17E-03 & \textbf{TRUE} \\  
 & \textbf{B3} & {\color[HTML]{CC0000} CPT+SFT} & Llama-2-13b-chat-hf-CPT-SFT & Llama-2-13b-chat-hf-CPT & constrained & 2.54E-02 & 6.15E-02 & 2.00E-04 & 4.17E-03 & \textbf{TRUE} \\  
 & \textbf{B3} & {\color[HTML]{CC0000} CPT+SFT} & Llama-2-13b-chat-hf-CPT-SFT & Llama-2-13b-chat-hf-CPT & greedy & 1.58E-01 & 2.31E-01 & 2.00E-04 & 4.17E-03 & \textbf{TRUE} \\  
 & \textbf{B4} & {\color[HTML]{38761D} SFT} & Llama-2-13b-chat-hf-SFT & Llama-2-13b-chat-hf & constrained & 3.11E-02 & 8.21E-02 & 2.00E-04 & 4.17E-03 & \textbf{TRUE} \\  
\multirow{-8}{*}{\textbf{INSTRUCT}} & \textbf{B4} & {\color[HTML]{38761D} SFT} & Llama-2-13b-chat-hf-SFT & Llama-2-13b-chat-hf & greedy & 1.72E-01 & 2.41E-01 & 2.00E-04 & 4.17E-03 & \textbf{TRUE} \\ \hline
 & \textbf{C1} & {\color[HTML]{0000FF} CPT} & MedLLaMA-13B-CPT & MedLLaMA\_13B & constrained & -2.15E-02 & 1.77E-02 & 7.73E-01 & 4.17E-03 & FALSE \\  
 & \textbf{C1} & {\color[HTML]{0000FF} CPT} & MedLLaMA-13B-CPT & MedLLaMA\_13B & greedy & -9.55E-03 & 2.34E-02 & 2.68E-01 & 4.17E-03 & FALSE \\  
 & \textbf{C2} & {\color[HTML]{CC0000} CPT+SFT} & MedLLaMA-13B-CPT-SFT & MedLLaMA\_13B & constrained & 3.11E-02 & 5.74E-02 & 2.00E-04 & 4.17E-03 & \textbf{TRUE} \\  
 & \textbf{C2} & {\color[HTML]{CC0000} CPT+SFT} & MedLLaMA-13B-CPT-SFT & MedLLaMA\_13B & greedy & 3.56E-02 & 5.96E-02 & 2.00E-04 & 4.17E-03 & \textbf{TRUE} \\  
 & \textbf{C3} & {\color[HTML]{CC0000} CPT+SFT} & MedLLaMA-13B-CPT-SFT & MedLLaMA-13B-CPT & constrained & 2.34E-02 & 7.11E-02 & 2.00E-04 & 4.17E-03 & \textbf{TRUE} \\  
 & \textbf{C3} & {\color[HTML]{CC0000} CPT+SFT} & MedLLaMA-13B-CPT-SFT & MedLLaMA-13B-CPT & greedy & 1.90E-02 & 6.57E-02 & 2.00E-04 & 4.17E-03 & \textbf{TRUE} \\  
 & \textbf{C4} & {\color[HTML]{38761D} SFT} & MedLLaMA-13B-SFT & MedLLaMA\_13B & constrained & 2.16E-02 & 4.90E-02 & 2.00E-04 & 4.17E-03 & \textbf{TRUE} \\  
\multirow{-8}{*}{\textbf{MEDICAL}} & \textbf{C4} & {\color[HTML]{38761D} SFT} & MedLLaMA-13B-SFT & MedLLaMA\_13B & greedy & 2.84E-02 & 5.80E-02 & 2.00E-04 & 4.17E-03 & \textbf{TRUE} \\ \hline
 & \textbf{D1} & {\color[HTML]{38761D} SFT} & Llama-2-13b-hf-SFT & Llama-2-13b-chat-hf-SFT & constrained & -4.30E-02 & 7.32E-03 & 2.15E-01 & 5.56E-03 & FALSE \\  
 & \textbf{D1} & {\color[HTML]{38761D} SFT} & Llama-2-13b-hf-SFT & Llama-2-13b-chat-hf-SFT & greedy & -1.36E-01 & -7.93E-02 & 2.00E-04 & 5.56E-03 & \textbf{TRUE} \\  
 & \textbf{D2} & {\color[HTML]{38761D} SFT} & Llama-2-13b-hf-SFT & MedLLaMA-13B-SFT & constrained & -3.65E-03 & 8.03E-03 & 4.70E-01 & 5.56E-03 & FALSE \\  
 & \textbf{D2} & {\color[HTML]{38761D} SFT} & Llama-2-13b-hf-SFT & MedLLaMA-13B-SFT & greedy & -4.03E-03 & 7.99E-03 & 5.32E-01 & 5.56E-03 & FALSE \\  
 & \textbf{D3} & {\color[HTML]{38761D} SFT} & Llama-2-13b-chat-hf-SFT & MedLLaMA-13B-SFT & constrained & -5.69E-03 & 4.60E-02 & 1.47E-01 & 5.56E-03 & FALSE \\  
\multirow{-6}{*}{\textbf{SFT}} & \textbf{D3} & {\color[HTML]{38761D} SFT} & Llama-2-13b-chat-hf-SFT & MedLLaMA-13B-SFT & greedy & 7.93E-02 & 1.41E-01 & 2.00E-04 & 5.56E-03 & \textbf{TRUE} \\ \hline
 & \textbf{E1} & {\color[HTML]{0000FF} CPT} & Llama-2-13b-hf-CPT & Llama-2-13b-chat-hf-CPT & constrained & -1.87E-02 & 1.02E-02 & 8.78E-01 & 5.56E-03 & FALSE \\  
 & \textbf{E1} & {\color[HTML]{0000FF} CPT} & Llama-2-13b-hf-CPT & Llama-2-13b-chat-hf-CPT & greedy & 2.31E-02 & 7.61E-02 & 2.00E-04 & 5.56E-03 & \textbf{TRUE} \\  
 & \textbf{E2} & {\color[HTML]{0000FF} CPT} & Llama-2-13b-hf-CPT & MedLLaMA-13B-CPT & constrained & 2.37E-03 & 3.99E-02 & 1.16E-02 & 5.56E-03 & FALSE \\  
 & \textbf{E2} & {\color[HTML]{0000FF} CPT} & Llama-2-13b-hf-CPT & MedLLaMA-13B-CPT & greedy & -7.36E-03 & 2.25E-02 & 7.03E-01 & 5.56E-03 & FALSE \\  
 & \textbf{E3} & {\color[HTML]{0000FF} CPT} & Llama-2-13b-chat-hf-CPT & MedLLaMA-13B-CPT & constrained & -5.92E-03 & 5.27E-02 & 2.03E-01 & 5.56E-03 & FALSE \\  
\multirow{-6}{*}{\textbf{CPT}} & \textbf{E3} & {\color[HTML]{0000FF} CPT} & Llama-2-13b-chat-hf-CPT & MedLLaMA-13B-CPT & greedy & -5.61E-02 & -2.58E-02 & 2.00E-04 & 5.56E-03 & \textbf{TRUE} \\ \hline
 & \textbf{F1} & {\color[HTML]{CC0000} CPT+SFT} & Llama-2-13b-hf-CPT-SFT & Llama-2-13b-chat-hf-CPT-SFT & constrained & -4.46E-02 & 4.48E-03 & 1.31E-01 & 5.56E-03 & FALSE \\  
 & \textbf{F1} & {\color[HTML]{CC0000} CPT+SFT} & Llama-2-13b-hf-CPT-SFT & Llama-2-13b-chat-hf-CPT-SFT & greedy & -1.48E-01 & -8.37E-02 & 2.00E-04 & 5.56E-03 & \textbf{TRUE} \\  
 & \textbf{F2} & {\color[HTML]{CC0000} CPT+SFT} & Llama-2-13b-hf-CPT-SFT & MedLLaMA-13B-CPT-SFT & constrained & -6.90E-03 & 8.26E-03 & 9.14E-01 & 5.56E-03 & FALSE \\  
 & \textbf{F2} & {\color[HTML]{CC0000} CPT+SFT} & Llama-2-13b-hf-CPT-SFT & MedLLaMA-13B-CPT-SFT & greedy & -7.39E-03 & 6.54E-03 & 7.66E-01 & 5.56E-03 & FALSE \\  
 & \textbf{F3} & {\color[HTML]{CC0000} CPT+SFT} & Llama-2-13b-chat-hf-CPT-SFT & MedLLaMA-13B-CPT-SFT & constrained & -6.57E-03 & 4.71E-02 & 1.56E-01 & 5.56E-03 & FALSE \\  
\multirow{-6}{*}{\textbf{CPT+SFT}} & \textbf{F3} & {\color[HTML]{CC0000} CPT+SFT} & Llama-2-13b-chat-hf-CPT-SFT & MedLLaMA-13B-CPT-SFT & greedy & 8.45E-02 & 1.47E-01 & 2.00E-04 & 5.56E-03 & \textbf{TRUE} \\ \hline
\end{tabular}%
}
\caption{Significance testing for MCQ/MCQU comparisons. reported separately for greedy and constrained decoding. Each row reports a paired bootstrap test between \texttt{model\_a} and \texttt{model\_b}. including the 95\% confidence interval of the mean EM difference. the two-sided $p$-value. and the Bonferroni-adjusted threshold with the resulting decision. IDs A--C compare adaptation strategies within the same model type; IDs D--F compare model initializations across types.}
\label{tab:stat-mcq}
\end{table*}

%% file: sig-stat-oeqa.tex
\begin{table*}[]
\centering
\resizebox{\textwidth}{!}{%
\begin{tabular}{llllllllll}
\hline
 & \textbf{id} & \textbf{strategy} & \textbf{model\_a} & \textbf{model\_b} & \textbf{ci95\_low} & \textbf{ci95\_high} & \textbf{p\_two\_sided} & \textbf{alpha\_Bonferoni} & \textbf{significant\_Bonferroni} \\ \hline
\multicolumn{10}{c}{\cellcolor[HTML]{CFE2F3}\textbf{Gemma 4B}} \\ \hline
 & \textbf{A1} & {\color[HTML]{0000FF} CPT} & gemma-3-4b-pt-CPT & gemma-3-4b-pt & -5.05E-02 & 3.69E-02 & 5.14E-01 & 4.17E-03 & FALSE \\  
 & \textbf{A2} & {\color[HTML]{CC0000} CPT+SFT} & gemma-3-4b-pt-CPT-SFT & gemma-3-4b-pt & -2.25E-01 & 1.65E-01 & 8.77E-01 & 4.17E-03 & FALSE \\  
 & \textbf{A3} & {\color[HTML]{CC0000} CPT+SFT} & gemma-3-4b-pt-CPT-SFT & gemma-3-4b-pt-CPT & -1.74E-01 & 1.31E-01 & 8.73E-01 & 4.17E-03 & FALSE \\  
\multirow{-4}{*}{\textbf{GENERAL}} & \textbf{A4} & {\color[HTML]{38761D} SFT} & gemma-3-4b-pt-SFT & gemma-3-4b-pt & -2.11E-01 & 1.41E-01 & 8.06E-01 & 4.17E-03 & FALSE \\ \hline
 & \textbf{B1} & {\color[HTML]{0000FF} CPT} & gemma-3-4b-it-CPT & gemma-3-4b-it & -4.58E-01 & -2.75E-01 & 2.00E-04 & 4.17E-03 & \textbf{TRUE} \\  
 & \textbf{B2} & {\color[HTML]{CC0000} CPT+SFT} & gemma-3-4b-it-CPT-SFT & gemma-3-4b-it & -4.58E-01 & -1.41E-01 & 2.00E-04 & 4.17E-03 & \textbf{TRUE} \\  
 & \textbf{B3} & {\color[HTML]{CC0000} CPT+SFT} & gemma-3-4b-it-CPT-SFT & gemma-3-4b-it-CPT & -4.94E-03 & 1.41E-01 & 6.44E-02 & 4.17E-03 & FALSE \\  
\multirow{-4}{*}{\textbf{INSTRUCT}} & \textbf{B4} & {\color[HTML]{38761D} SFT} & gemma-3-4b-it-SFT & gemma-3-4b-it & -3.81E-01 & -2.06E-01 & 2.00E-04 & 4.17E-03 & \textbf{TRUE} \\ \hline
 & \textbf{C1} & {\color[HTML]{0000FF} CPT} & medgemma-4b-pt-CPT & medgemma-4b-pt & -3.46E-01 & -9.28E-02 & 2.00E-04 & 4.17E-03 & \textbf{TRUE} \\  
 & \textbf{C2} & {\color[HTML]{CC0000} CPT+SFT} & medgemma-4b-pt-CPT-SFT & medgemma-4b-pt-CPT & 3.71E-02 & 1.63E-01 & 2.00E-04 & 4.17E-03 & \textbf{TRUE} \\  
 & \textbf{C3} & {\color[HTML]{CC0000} CPT+SFT} & medgemma-4b-pt-CPT-SFT & medgemma-4b-pt & -2.79E-01 & 3.51E-02 & 1.20E-01 & 4.17E-03 & FALSE \\  
\multirow{-4}{*}{\textbf{MEDICAL}} & \textbf{C4} & {\color[HTML]{38761D} SFT} & medgemma-4b-pt-SFT & medgemma-4b-pt & -1.88E-01 & -1.94E-04 & 3.96E-02 & 4.17E-03 & FALSE \\ \hline
 & \textbf{D1} & {\color[HTML]{38761D} SFT} & gemma-3-4b-pt-SFT & gemma-3-4b-it-SFT & -3.40E-02 & 6.17E-02 & 7.03E-01 & 5.56E-03 & FALSE \\  
 & \textbf{D2} & {\color[HTML]{38761D} SFT} & gemma-3-4b-pt-SFT & medgemma-4b-pt-SFT & 4.05E-02 & 9.53E-02 & 2.00E-04 & 5.56E-03 & \textbf{TRUE} \\  
\multirow{-3}{*}{\textbf{SFT}} & \textbf{D3} & {\color[HTML]{38761D} SFT} & gemma-3-4b-it-SFT & medgemma-4b-pt-SFT & 2.34E-02 & 9.61E-02 & 2.00E-04 & 5.56E-03 & \textbf{TRUE} \\ \hline
 & \textbf{E1} & {\color[HTML]{0000FF} CPT} & gemma-3-4b-pt-CPT & gemma-3-4b-it-CPT & 4.04E-03 & 2.43E-01 & 4.24E-02 & 5.56E-03 & FALSE \\  
 & \textbf{E2} & {\color[HTML]{0000FF} CPT} & gemma-3-4b-pt-CPT & medgemma-4b-pt-CPT & 9.44E-02 & 4.14E-01 & 2.00E-04 & 5.56E-03 & \textbf{TRUE} \\  
\multirow{-3}{*}{\textbf{CPT}} & \textbf{E3} & {\color[HTML]{0000FF} CPT} & gemma-3-4b-it-CPT & medgemma-4b-pt-CPT & 6.99E-02 & 1.63E-01 & 2.00E-04 & 5.56E-03 & \textbf{TRUE} \\ \hline
 & \textbf{F1} & {\color[HTML]{CC0000} CPT+SFT} & gemma-3-4b-pt-CPT-SFT & gemma-3-4b-it-CPT-SFT & 5.49E-03 & 7.53E-02 & 6.80E-03 & 5.56E-03 & FALSE \\  
 & \textbf{F2} & {\color[HTML]{CC0000} CPT+SFT} & gemma-3-4b-pt-CPT-SFT & medgemma-4b-pt-CPT-SFT & 7.33E-02 & 1.95E-01 & 2.00E-04 & 5.56E-03 & \textbf{TRUE} \\  
\multirow{-3}{*}{\textbf{CPT+SFT}} & \textbf{F3} & {\color[HTML]{CC0000} CPT+SFT} & gemma-3-4b-it-CPT-SFT & medgemma-4b-pt-CPT-SFT & 3.56E-02 & 1.55E-01 & 2.00E-04 & 5.56E-03 & \textbf{TRUE} \\ \hline
\multicolumn{10}{c}{\cellcolor[HTML]{CFE2F3}\textbf{Mistral 7B}} \\ \hline
 & \textbf{A1} & {\color[HTML]{0000FF} CPT} & Mistral-7B-v0.1-CPT & Mistral-7B-v0.1 & -1.74E-01 & 5.41E-02 & 6.35E-01 & 4.17E-03 & FALSE \\  
 & \textbf{A2} & {\color[HTML]{CC0000} CPT+SFT} & Mistral-7B-v0.1-CPT-SFT & Mistral-7B-v0.1 & -2.09E-01 & 1.37E-01 & 8.37E-01 & 4.17E-03 & FALSE \\  
 & \textbf{A3} & {\color[HTML]{CC0000} CPT+SFT} & Mistral-7B-v0.1-CPT-SFT & Mistral-7B-v0.1-CPT & -8.42E-02 & 9.06E-02 & 9.22E-01 & 4.17E-03 & FALSE \\  
\multirow{-4}{*}{\textbf{GENERAL}} & \textbf{A4} & {\color[HTML]{38761D} SFT} & Mistral-7B-v0.1-SFT & Mistral-7B-v0.1 & -2.32E-01 & 1.02E-01 & 6.17E-01 & 4.17E-03 & FALSE \\ \hline
 & \textbf{B1} & {\color[HTML]{0000FF} CPT} & Mistral-7B-Instruct-v0.1-CPT & Mistral-7B-Instruct-v0.1 & -4.68E-02 & 1.69E-01 & 1.48E-01 & 4.17E-03 & FALSE \\  
 & \textbf{B2} & {\color[HTML]{CC0000} CPT+SFT} & Mistral-7B-Instruct-v0.1-CPT-SFT & Mistral-7B-Instruct-v0.1-CPT & -7.19E-02 & -3.74E-02 & 2.00E-04 & 4.17E-03 & \textbf{TRUE} \\  
 & \textbf{B3} & {\color[HTML]{CC0000} CPT+SFT} & Mistral-7B-Instruct-v0.1-CPT-SFT & Mistral-7B-Instruct-v0.1 & -1.15E-01 & 1.34E-01 & 7.76E-01 & 4.17E-03 & FALSE \\  
\multirow{-4}{*}{\textbf{INSTRUCT}} & \textbf{B4} & {\color[HTML]{38761D} SFT} & Mistral-7B-Instruct-v0.1-SFT & Mistral-7B-Instruct-v0.1 & -2.64E-01 & 1.30E-02 & 1.29E-01 & 4.17E-03 & FALSE \\ \hline
 & \textbf{C1} & {\color[HTML]{0000FF} CPT} & BioMistral-7B-CPT & BioMistral-7B & -1.74E-01 & 7.18E-02 & 6.52E-01 & 4.17E-03 & FALSE \\  
 & \textbf{C2} & {\color[HTML]{CC0000} CPT+SFT} & BioMistral-7B-CPT-SFT & BioMistral-7B & -3.40E-02 & 1.03E-01 & 2.42E-01 & 4.17E-03 & FALSE \\  
 & \textbf{C3} & {\color[HTML]{CC0000} CPT+SFT} & BioMistral-7B-CPT-SFT & BioMistral-7B-CPT & 1.04E-02 & 1.31E-01 & 9.00E-03 & 4.17E-03 & FALSE \\  
\multirow{-4}{*}{\textbf{MEDICAL}} & \textbf{C4} & {\color[HTML]{38761D} SFT} & BioMistral-7B-SFT & BioMistral-7B & -1.41E-01 & 4.33E-02 & 4.24E-01 & 4.17E-03 & FALSE \\ \hline
 & \textbf{D1} & {\color[HTML]{38761D} SFT} & Mistral-7B-v0.1-SFT & Mistral-7B-Instruct-v0.1-SFT & 2.14E-02 & 9.83E-02 & 2.00E-04 & 5.56E-03 & \textbf{TRUE} \\  
 & \textbf{D2} & {\color[HTML]{38761D} SFT} & Mistral-7B-Instruct-v0.1-SFT & BioMistral-7B-SFT & -3.04E-02 & 4.71E-02 & 7.02E-01 & 5.56E-03 & FALSE \\  
\multirow{-3}{*}{\textbf{SFT}} & \textbf{D3} & {\color[HTML]{38761D} SFT} & Mistral-7B-v0.1-SFT & BioMistral-7B-SFT & 4.48E-02 & 7.93E-02 & 2.00E-04 & 5.56E-03 & \textbf{TRUE} \\ \hline
 & \textbf{E1} & {\color[HTML]{0000FF} CPT} & Mistral-7B-v0.1-CPT & Mistral-7B-Instruct-v0.1-CPT & -1.51E-01 & -9.28E-02 & 2.00E-04 & 5.56E-03 & \textbf{TRUE} \\  
 & \textbf{E2} & {\color[HTML]{0000FF} CPT} & Mistral-7B-v0.1-CPT & BioMistral-7B-CPT & -2.97E-02 & 1.67E-01 & 2.44E-01 & 5.56E-03 & FALSE \\  
\multirow{-3}{*}{\textbf{CPT}} & \textbf{E3} & {\color[HTML]{0000FF} CPT} & Mistral-7B-Instruct-v0.1-CPT & BioMistral-7B-CPT & 1.02E-01 & 3.12E-01 & 2.00E-04 & 5.56E-03 & \textbf{TRUE} \\ \hline
 & \textbf{F1} & {\color[HTML]{CC0000} CPT+SFT} & Mistral-7B-v0.1-CPT-SFT & Mistral-7B-Instruct-v0.1-CPT-SFT & -1.25E-01 & -1.04E-02 & 7.40E-03 & 5.56E-03 & FALSE \\  
 & \textbf{F2} & {\color[HTML]{CC0000} CPT+SFT} & Mistral-7B-v0.1-CPT-SFT & BioMistral-7B-CPT-SFT & -2.23E-02 & 3.51E-02 & 8.31E-01 & 5.56E-03 & FALSE \\  
\multirow{-3}{*}{\textbf{CPT+SFT}} & \textbf{F3} & {\color[HTML]{CC0000} CPT+SFT} & Mistral-7B-Instruct-v0.1-CPT-SFT & BioMistral-7B-CPT-SFT & 4.40E-02 & 1.07E-01 & 2.00E-04 & 5.56E-03 & \textbf{TRUE} \\ \hline
\multicolumn{10}{c}{\cellcolor[HTML]{CFE2F3}\textbf{LLAMA-7 FAMILY}} \\ \hline
 & \textbf{A1} & {\color[HTML]{0000FF} CPT} & Llama-2-7b-hf-CPT & Llama-2-7b-hf & -1.17E-01 & 2.53E-03 & 7.46E-02 & 4.17E-03 & FALSE \\  
 & \textbf{A2} & {\color[HTML]{CC0000} CPT+SFT} & LLama-2-7b-hf-CPT-SFT & Llama-2-7b-hf-CPT & 4.46E-02 & 1.40E-01 & 2.00E-04 & 4.17E-03 & \textbf{TRUE} \\  
 & \textbf{A3} & {\color[HTML]{CC0000} CPT+SFT} & LLama-2-7b-hf-CPT-SFT & Llama-2-7b-hf & -5.92E-02 & 9.36E-02 & 5.18E-01 & 4.17E-03 & FALSE \\  
\multirow{-4}{*}{\textbf{GENERAL}} & \textbf{A4} & {\color[HTML]{38761D} SFT} & LLama-2-7b-hf-SFT & Llama-2-7b-hf & -9.58E-02 & 5.00E-02 & 8.01E-01 & 4.17E-03 & FALSE \\ \hline
 & \textbf{B1} & {\color[HTML]{0000FF} CPT} & Llama-2-7b-chat-hf-CPT & Llama-2-7b-chat-hf & -1.05E-01 & 7.92E-02 & 7.39E-01 & 4.17E-03 & FALSE \\  
 & \textbf{B2} & {\color[HTML]{CC0000} CPT+SFT} & Llama-2-7b-chat-hf-CPT-SFT & Llama-2-7b-chat-hf & -4.91E-02 & 7.67E-02 & 4.88E-01 & 4.17E-03 & FALSE \\  
 & \textbf{B3} & {\color[HTML]{CC0000} CPT+SFT} & Llama-2-7b-chat-hf-CPT-SFT & Llama-2-7b-chat-hf-CPT & -3.13E-02 & 6.42E-02 & 3.87E-01 & 4.17E-03 & FALSE \\  
\multirow{-4}{*}{\textbf{INSTRUCT}} & \textbf{B4} & {\color[HTML]{38761D} SFT} & Llama-2-7b-chat-hf-SFT & Llama-2-7b-chat-hf & -2.67E-01 & 2.10E-03 & 8.08E-02 & 4.17E-03 & FALSE \\ \hline
 & \textbf{C1} & {\color[HTML]{0000FF} CPT} & meditron-7b-CPT & meditron-7b & -8.36E-03 & 1.95E-02 & 4.18E-01 & 4.17E-03 & FALSE \\  
 & \textbf{C2} & {\color[HTML]{CC0000} CPT+SFT} & meditron-7b-CPT-SFT & meditron-7b & 3.69E-03 & 9.40E-02 & 4.08E-02 & 4.17E-03 & FALSE \\  
 & \textbf{C3} & {\color[HTML]{CC0000} CPT+SFT} & meditron-7b-CPT-SFT & meditron-7b-CPT & -1.49E-02 & 9.99E-02 & 1.30E-01 & 4.17E-03 & FALSE \\  
\multirow{-4}{*}{\textbf{MEDICAL}} & \textbf{C4} & {\color[HTML]{38761D} SFT} & meditron-7b-SFT & meditron-7b & -9.43E-02 & 3.43E-02 & 4.08E-01 & 4.17E-03 & FALSE \\ \hline
 & \textbf{D1} & {\color[HTML]{38761D} SFT} & LLama-2-7b-hf-SFT & Llama-2-7b-chat-hf-SFT & -9.81E-02 & -2.37E-02 & 2.00E-04 & 5.56E-03 & \textbf{TRUE} \\  
 & \textbf{D2} & {\color[HTML]{38761D} SFT} & LLama-2-7b-hf-SFT & meditron-7b-SFT & -3.18E-02 & -4.98E-03 & 7.80E-03 & 5.56E-03 & FALSE \\  
\multirow{-3}{*}{\textbf{SFT}} & \textbf{D3} & {\color[HTML]{38761D} SFT} & Llama-2-7b-chat-hf-SFT & meditron-7b-SFT & -4.93E-03 & 8.30E-02 & 1.49E-01 & 5.56E-03 & FALSE \\ \hline
 & \textbf{E1} & {\color[HTML]{0000FF} CPT} & Llama-2-7b-hf-CPT & Llama-2-7b-chat-hf-CPT & -2.96E-01 & -1.31E-01 & 2.00E-04 & 5.56E-03 & \textbf{TRUE} \\  
 & \textbf{E2} & {\color[HTML]{0000FF} CPT} & Llama-2-7b-hf-CPT & meditron-7b-CPT & -1.37E-01 & -3.73E-02 & 2.00E-04 & 5.56E-03 & \textbf{TRUE} \\  
\multirow{-3}{*}{\textbf{CPT}} & \textbf{E3} & {\color[HTML]{0000FF} CPT} & Llama-2-7b-chat-hf-CPT & meditron-7b-CPT & 8.64E-02 & 1.72E-01 & 2.00E-04 & 5.56E-03 & \textbf{TRUE} \\ \hline
 & \textbf{F1} & {\color[HTML]{CC0000} CPT+SFT} & LLama-2-7b-hf-CPT-SFT & Llama-2-7b-chat-hf-CPT-SFT & -2.83E-01 & -5.47E-02 & 2.00E-04 & 5.56E-03 & \textbf{TRUE} \\  
 & \textbf{F2} & {\color[HTML]{CC0000} CPT+SFT} & LLama-2-7b-hf-CPT-SFT & meditron-7b-CPT-SFT & -7.39E-02 & -3.20E-02 & 2.00E-04 & 5.56E-03 & \textbf{TRUE} \\  
\multirow{-3}{*}{\textbf{CPT+SFT}} & \textbf{F3} & {\color[HTML]{CC0000} CPT+SFT} & Llama-2-7b-chat-hf-CPT-SFT & meditron-7b-CPT-SFT & 6.85E-03 & 2.06E-01 & 7.00E-03 & 5.56E-03 & FALSE \\ \hline
\multicolumn{10}{c}{\cellcolor[HTML]{CFE2F3}\textbf{Llama 13B}} \\ \hline
 & \textbf{A1} & {\color[HTML]{0000FF} CPT} & Llama-2-13b-hf-CPT & Llama-2-13b-hf & -9.64E-02 & -2.62E-02 & 2.00E-04 & 4.17E-03 & \textbf{TRUE} \\  
 & \textbf{A2} & {\color[HTML]{CC0000} CPT+SFT} & Llama-2-13b-hf-CPT-SFT & Llama-2-13b-hf & -1.25E-02 & 1.84E-01 & 1.19E-01 & 4.17E-03 & FALSE \\  
 & \textbf{A3} & {\color[HTML]{CC0000} CPT+SFT} & Llama-2-13b-hf-CPT-SFT & Llama-2-13b-hf-CPT & 6.32E-02 & 2.31E-01 & 2.00E-04 & 4.17E-03 & \textbf{TRUE} \\  
\multirow{-4}{*}{\textbf{GENERAL}} & \textbf{A4} & {\color[HTML]{38761D} SFT} & Llama-2-13b-hf-SFT & Llama-2-13b-hf & -3.35E-02 & 9.53E-02 & 4.87E-01 & 4.17E-03 & FALSE \\ \hline
 & \textbf{B1} & {\color[HTML]{0000FF} CPT} & Llama-2-13b-chat-hf-CPT & Llama-2-13b-chat-hf & -1.03E-02 & 1.32E-01 & 1.44E-01 & 4.17E-03 & FALSE \\  
 & \textbf{B2} & {\color[HTML]{CC0000} CPT+SFT} & Llama-2-13b-chat-hf-CPT-SFT & Llama-2-13b-chat-hf & -3.25E-01 & 6.12E-02 & 4.22E-01 & 4.17E-03 & FALSE \\  
 & \textbf{B3} & {\color[HTML]{CC0000} CPT+SFT} & Llama-2-13b-chat-hf-CPT-SFT & Llama-2-13b-chat-hf-CPT & -3.19E-01 & -6.80E-02 & 2.00E-04 & 4.17E-03 & \textbf{TRUE} \\  
\multirow{-4}{*}{\textbf{INSTRUCT}} & \textbf{B4} & {\color[HTML]{38761D} SFT} & Llama-2-13b-chat-hf-SFT & Llama-2-13b-chat-hf & -3.78E-01 & -4.44E-03 & 4.28E-02 & 4.17E-03 & FALSE \\ \hline
 & \textbf{C1} & {\color[HTML]{0000FF} CPT} & MedLLaMA-13B-CPT & MedLLaMA\_13B & 8.59E-03 & 3.82E-02 & 9.60E-03 & 4.17E-03 & FALSE \\  
 & \textbf{C2} & {\color[HTML]{CC0000} CPT+SFT} & MedLLaMA-13B-CPT-SFT & MedLLaMA\_13B & -7.65E-03 & 1.81E-01 & 1.50E-01 & 4.17E-03 & FALSE \\  
 & \textbf{C3} & {\color[HTML]{CC0000} CPT+SFT} & MedLLaMA-13B-CPT-SFT & MedLLaMA-13B-CPT & -2.00E-02 & 1.43E-01 & 1.48E-01 & 4.17E-03 & FALSE \\  
\multirow{-4}{*}{\textbf{MEDICAL}} & \textbf{C4} & {\color[HTML]{38761D} SFT} & MedLLaMA-13B-SFT & MedLLaMA\_13B & -4.70E-02 & 9.93E-02 & 3.78E-01 & 4.17E-03 & FALSE \\ \hline
 & \textbf{D1} & {\color[HTML]{38761D} SFT} & Llama-2-13b-hf-SFT & Llama-2-13b-chat-hf-SFT & 5.63E-03 & 5.57E-02 & 2.00E-04 & 5.56E-03 & \textbf{TRUE} \\  
 & \textbf{D2} & {\color[HTML]{38761D} SFT} & Llama-2-13b-hf-SFT & MedLLaMA-13B-SFT & -2.49E-02 & 4.98E-02 & 6.34E-01 & 5.56E-03 & FALSE \\  
\multirow{-3}{*}{\textbf{SFT}} & \textbf{D3} & {\color[HTML]{38761D} SFT} & Llama-2-13b-chat-hf-SFT & MedLLaMA-13B-SFT & -7.22E-02 & 1.13E-02 & 6.27E-01 & 5.56E-03 & FALSE \\ \hline
 & \textbf{E1} & {\color[HTML]{0000FF} CPT} & Llama-2-13b-hf-CPT & Llama-2-13b-chat-hf-CPT & -4.04E-01 & -2.38E-01 & 2.00E-04 & 5.56E-03 & \textbf{TRUE} \\  
 & \textbf{E2} & {\color[HTML]{0000FF} CPT} & Llama-2-13b-hf-CPT & MedLLaMA-13B-CPT & -9.08E-02 & -6.25E-02 & 2.00E-04 & 5.56E-03 & \textbf{TRUE} \\  
\multirow{-3}{*}{\textbf{CPT}} & \textbf{E3} & {\color[HTML]{0000FF} CPT} & Llama-2-13b-chat-hf-CPT & MedLLaMA-13B-CPT & 1.63E-01 & 3.21E-01 & 2.00E-04 & 5.56E-03 & \textbf{TRUE} \\ \hline
 & \textbf{F1} & {\color[HTML]{CC0000} CPT+SFT} & Llama-2-13b-hf-CPT-SFT & Llama-2-13b-chat-hf-CPT-SFT & -4.38E-02 & 4.50E-02 & 8.76E-01 & 5.56E-03 & FALSE \\  
 & \textbf{F2} & {\color[HTML]{CC0000} CPT+SFT} & Llama-2-13b-hf-CPT-SFT & MedLLaMA-13B-CPT-SFT & -2.24E-02 & 3.90E-02 & 5.73E-01 & 5.56E-03 & FALSE \\  
\multirow{-3}{*}{\textbf{CPT+SFT}} & \textbf{F3} & {\color[HTML]{CC0000} CPT+SFT} & Llama-2-13b-chat-hf-CPT-SFT & MedLLaMA-13B-CPT-SFT & -1.39E-02 & 2.73E-02 & 5.62E-01 & 5.56E-03 & FALSE \\ \hline
\end{tabular}%
}
\caption{Significance testing for OEQA comparisons. Each row reports a paired bootstrap test between \texttt{model\_a} and \texttt{model\_b}, including the 95\% confidence interval of the mean difference (\texttt{ci95\_low}, \texttt{ci95\_high}), the two-sided $p$-value, and the Bonferroni-adjusted threshold (\texttt{alpha\_Bonferoni}) with the resulting decision (\texttt{significant\_Bonferroni}). IDs A--C compare adaptation strategies within the same model type; IDs D--F compare model initializations across types.}
\label{tab:stat-oeqa}
\end{table*}

%% file: near-miss-rate.tex
\begin{table}[H]
\centering
\small
\begin{tabular}{l c c}
\hline
\textbf{Model} & \textbf{MCQ} & \textbf{MCQU} \\
\hline
Mistral                  & 0.203 & 0.202 \\
Mistral-CPT              & 0.212 & 0.203 \\
Mistral-SFT              & 0.234 & 0.204 \\
Mistral-CPT-SFT          & 0.256 & 0.203 \\
\hline
Mistral-Instruct         & 0.173 & 0.202 \\
Mistral-Instruct-CPT     & 0.205 & 0.202 \\
Mistral-Instruct-SFT     & 0.174 & 0.206 \\
Mistral-Instruct-CPT-SFT & 0.221 & 0.192 \\
\hline
BioMistral               & 0.181 & 0.205 \\
BioMistral-CPT           & 0.195 & 0.201 \\
BioMistral-SFT           & 0.220 & 0.211 \\
BioMistral-CPT-SFT       & 0.234 & 0.204 \\
\hline
\end{tabular}
\caption{Near-miss rates for MCQ and MCQU across Mistral variants. A near-miss corresponds to cases where all gold answers are ranked within the top-$k$ options but the generated answer does not match the gold label(s). Near-miss rates remain stable across model families and adaptation strategies, indicating that improvements in confidence and ranking do not directly translate into exact prediction.}
\label{tab:near-miss-MCQ-MCQu}
\end{table}

%% file: oeqa-verbosity.tex
\begin{table*}[!t]
\centering
\resizebox{\textwidth}{!}{%
\begin{tabular}{ccllllll}
\hline
\textbf{Model Type} & \textbf{Strategy} & \textit{\textbf{mean\_words}} & \textit{\textbf{std\_words}} & \textit{\textbf{median\_words}} & \textit{\textbf{mean\_chars}} & \textit{\textbf{std\_chars}} & \textit{\textbf{median\_chars}} \\ \hline
\multicolumn{8}{c}{\cellcolor[HTML]{CFE2F3}\textit{\textbf{Gemma-4B}}} \\ \hline
 & \textbf{Base} & 243,30 & 123,00 & 288,00 & 1\,616,51 & 798,74 & 1,927,00 \\
 & {\color[HTML]{0000FF} CPT} & 107,81 & 122,82 & 37,00 & 720,79 & 806,61 & 245,00 \\
 & {\color[HTML]{6AA84F} SFT} & 176,70 & \textbf{127,77} & 204,00 & 1\,182,16 & \textbf{877,20} & 1\,153,00 \\
\multirow{-4}{*}{\textbf{GENERAL}} & {\color[HTML]{FF0000} CPT+SFT} & \textbf{279,84} & 87,77 & \textbf{300,00} & \textbf{1\,884,34} & 649,64 & \textbf{2\,076,00} \\ \hline
 & \textbf{Base} & 261,62 & 67,98 & 282,00 & 1\,819,58 & 473,70 & 1\,985,00 \\
 & {\color[HTML]{0000FF} CPT} & \textbf{266,87} & 98,66 & \textbf{286,00} & 1\,763,62 & 576,08 & 1\,878,00 \\
 & {\color[HTML]{6AA84F} SFT} & 243,03 & 65,58 & 261,00 & \textbf{3\,513,80} & 827,10 & \textbf{3\,326,00} \\
\multirow{-4}{*}{\textbf{INSTRUCT}} & {\color[HTML]{FF0000} CPT+SFT} & 183,16 & \textbf{98,86} & 207,00 & 3\,497,45 & \textbf{1\,680,82} & 2\,632,00 \\ \hline
 & \textbf{Base} & 208,62 & 122,41 & 205,00 & 1\,371,82 & 793,35 & 1\,431,00 \\
 & {\color[HTML]{0000FF} CPT} & 216,10 & \textbf{142,15} & 264,00 & 1\,308,81 & \textbf{861,75} & 1\,661,00 \\
 & {\color[HTML]{6AA84F} SFT} & 271,22 & 97,47 & 292,00 & 1\,796,17 & 705,16 & \textbf{1\,982,00} \\
\multirow{-4}{*}{\textbf{MEDICAL}} & {\color[HTML]{FF0000} CPT+SFT} & \textbf{282,47} & 48,88 & \textbf{283,00} & \textbf{1\,825,36} & 519,84 & 1\,954,00 \\ \hline
\multicolumn{8}{c}{\cellcolor[HTML]{CFE2F3}\textit{\textbf{Mistral-7B}}} \\ \hline
 & \textbf{Base} & 212,56 & 58,84 & 224,00 & 1\,466,40 & 329,83 & 1\,502,00 \\
 & {\color[HTML]{0000FF} CPT} & 173,15 & 84,30 & 193,00 & 1\,102,15 & 536,40 & 1,321,50 \\
 & {\color[HTML]{6AA84F} SFT} & 130,36 & \textbf{90,43} & 138,00 & 884,67 & \textbf{603,37} & 906,00 \\
\multirow{-4}{*}{\textbf{GENERAL}} & {\color[HTML]{FF0000} CPT+SFT} & \textbf{226,60} & 31,85 & \textbf{229,00} & \textbf{1\,508,78} & 236,78 & \textbf{1\,527,00} \\ \hline
 & \textbf{Base} & 134,18 & 74,11 & 125,00 & 876,12 & 476,77 & 812,00 \\
 & {\color[HTML]{0000FF} CPT} & 67,79 & 75,91 & 37,00 & 447,32 & 481,36 & 244,00 \\
 & {\color[HTML]{6AA84F} SFT} & 19,59 & 14,19 & 19,00 & 138,75 & 100,30 & 136,00 \\
\multirow{-4}{*}{\textbf{INSTRUCT}} & {\color[HTML]{FF0000} CPT+SFT} & \textbf{168,09} & \textbf{77,91} & \textbf{199,00} & \textbf{1\,112,24} & \textbf{499,35} & \textbf{1\,314,00} \\ \hline
 & \textbf{Base} & 66,26 & 76,56 & 37,00 & 443,89 & 500,84 & 250,00 \\
 & {\color[HTML]{0000FF} CPT} & 99,03 & \textbf{102,06} & 41,00 & 651,69 & 650,72 & 279,00 \\
 & {\color[HTML]{6AA84F} SFT} & 128,83 & 98,75 & \textbf{159,00} & 855,12 & \textbf{657,31} & \textbf{950,00} \\
\multirow{-4}{*}{\textbf{MEDICAL}} & {\color[HTML]{FF0000} CPT+SFT} & \textbf{132,20} & 91,72 & 129,00 & \textbf{909,46} & 619,60 & 859,00 \\ \hline
\multicolumn{8}{c}{\cellcolor[HTML]{CFE2F3}\textit{\textbf{Llama-7B}}} \\ \hline
 & \textbf{Base} & 206,13 & 67,03 & 222,00 & 1\,358,82 & 392,59 & \textbf{1\,459,00} \\
 & {\color[HTML]{0000FF} CPT} & 41,99 & \textbf{78,06} & 8,00 & 269,34 & \textbf{490,77} & 56,00 \\
 & {\color[HTML]{6AA84F} SFT} & \textbf{219,26} & 35,68 & 220,00 & \textbf{1\,399,00} & 266,52 & 1\,441,00 \\
\multirow{-4}{*}{\textbf{GENERAL}} & {\color[HTML]{FF0000} CPT+SFT} & 217,98 & 50,08 & \textbf{224,00} & 1\,376,29 & 353,15 & 1\,442,00 \\ \hline
 & \textbf{Base} & \textbf{233,79} & 66,58 & \textbf{244,00} & \textbf{1\,513,45} & 433,69 & \textbf{1\,586,00} \\ 
 & {\color[HTML]{0000FF} CPT} & 72,29 & \textbf{87,21} & 26,00 & 483,67 & 572,37 & 180,00 \\
 & {\color[HTML]{6AA84F} SFT} & 18,24 & 14,33 & 18,00 & 126,03 & 93,60 & 127,00 \\
\multirow{-4}{*}{\textbf{INSTRUCT}} & {\color[HTML]{FF0000} CPT+SFT} & 123,65 & 86,95 & 106,00 & 859,20 & \textbf{581,43} & 784,00 \\ \hline
 & \textbf{Base} & \textbf{227,34} & 40,78 & \textbf{230,00} & \textbf{1\,498,30} & 211,08 & \textbf{1\,522,50} \\
 & {\color[HTML]{0000FF} CPT} & 130,92 & \textbf{104,72} & 128,00 & 847,07 & \textbf{671,76} & 1\,008,00 \\
 & {\color[HTML]{6AA84F} SFT} & 204,04 & 40,98 & 209,00 & 1\,350,16 & 345,45 & 1\,431,00 \\
\multirow{-4}{*}{\textbf{MEDICAL}} & {\color[HTML]{FF0000} CPT+SFT} & 216,96 & 38,84 & 220,00 & 1\,416,50 & 301,18 & 1\,466,00 \\ \hline
\multicolumn{8}{c}{\cellcolor[HTML]{CFE2F3}\textit{\textbf{Llama-13B}}} \\ \hline
 & \textbf{Base} & 168,15 & 44,63 & 146,00 & 1\,144,75 & 260,69 & 1\,020,00 \\
 & {\color[HTML]{0000FF} CPT} & 26,19 & \textbf{53,89} & 9,00 & 173,89 & \textbf{351,27} & 58,00 \\
 & {\color[HTML]{6AA84F} SFT} & 218,99 & 36,11 & 219,00 & \textbf{1\,440,23} & 300,92 & \textbf{1\,523,00} \\
\multirow{-4}{*}{\textbf{GENERAL}} & {\color[HTML]{FF0000} CPT+SFT} & \textbf{225,12} & 36,23 & \textbf{228,00} & 1\,429,25 & 280,81 & 1\,482,00 \\ \hline
 & \textbf{Base} & \textbf{226,32} & 60,72 & \textbf{235,00} & \textbf{1\,471,46} & 396,55 & \textbf{1\,529,00} \\
 & {\color[HTML]{0000FF} CPT} & 79,32 & \textbf{80,61} & 45,00 & 528,55 & \textbf{526,68} & 303,00 \\
 & {\color[HTML]{6AA84F} SFT} & 17,34 & 10,39 & 18,00 & 121,18 & 74,56 & 129,00 \\
\multirow{-4}{*}{\textbf{INSTRUCT}} & {\color[HTML]{FF0000} CPT+SFT} & 19,79 & 17,18 & 19,00 & 141,49 & 132,88 & 136,00 \\ \hline
 & \textbf{Base} & \textbf{217,49} & 47,81 & \textbf{221,00} & \textbf{1\,429,19} & 269,92 & 1\,459,00 \\
 & {\color[HTML]{0000FF} CPT} & 65,17 & \textbf{92,92} & 13,00 & 431,03 & \textbf{597,64} & 92,00 \\
 & {\color[HTML]{6AA84F} SFT} & 206,21 & 53,97 & 219,00 & 1\,394,11 & 392,06 & \textbf{1\,505,00} \\
\multirow{-4}{*}{\textbf{MEDICAL}} & {\color[HTML]{FF0000} CPT+SFT} & 215,44 & 37,48 & 220,00 & 1\,351,46 & 345,39 & 1\,402,50 \\ \hline
\end{tabular}%
}
\caption{Output length statistics for OEQA generations across model families, initialization types (GENERAL/INSTRUCT/MEDICAL), and adaptation strategies (Base, CPT, SFT, CPT+SFT). We report the mean, standard deviation, and median number of words and characters per generated answer. \textbf{Bold} values highlight, within each block, the maximum value for the corresponding statistic.}
\label{tab:oeqa-verb}
\end{table*}

%% file: en-vs-fr.tex
\begin{table}[]
\centering
\resizebox{\columnwidth}{!}{%
\begin{tabular}{cccc|cc}
\hline
 &  & \multicolumn{2}{c|}{\textbf{MCQU-FR}} & \multicolumn{2}{c}{\textbf{MCQU-EN}} \\ \cline{3-6} 
 &  & \textbf{Greedy} & \textbf{Constrained} & \textbf{Greedy} & \textbf{Constrained} \\ \cline{3-6} 
\multirow{-3}{*}{\textbf{Model Type}} & \multirow{-3}{*}{\textbf{Strategy}} & \multicolumn{2}{c|}{\textbf{EM}} & \multicolumn{2}{c}{\textbf{EM}} \\ \hline
\multicolumn{6}{c}{\cellcolor[HTML]{CFE2F3}\textit{\textbf{Gemma-4B}}} \\ \hline
 & \textbf{Base} & \textbf{5.76} & 26.63 & 1.60 & \textbf{41.22} \\ 
 & {\color[HTML]{0000FF}CPT} & 8.54 & 25.60 & \textbf{12.77} & \textbf{40.10} \\ 
 & {\color[HTML]{6AA84F}SFT} & 19.92 & 32.82 & \textbf{51.60} & \textbf{51.60} \\ 
\multirow{-4}{*}{\textbf{GENERAL}} & {\color[HTML]{FF0000}CPT+SFT} & 19.55 & 32.73 & \textbf{51.42} & \textbf{51.42} \\ \hline
 & \textbf{Base} & 29.38 & 29.76 & \textbf{47.89} & \textbf{47.94} \\ 
 & {\color[HTML]{0000FF}CPT} & 1.36 & \textbf{24.28} & \textbf{1.39} & 23.43 \\ 
 & {\color[HTML]{6AA84F}SFT} & 32.38 & 32.46 & \textbf{48.74} & \textbf{48.74} \\ 
\multirow{-4}{*}{\textbf{INSTRUCT}} & {\color[HTML]{FF0000}CPT+SFT} & 30.31 & 30.38 & \textbf{39.17} & \textbf{39.17} \\ \hline
 & \textbf{Base} & \textbf{11.50} & 26.43 & 0.04 & \textbf{32.47} \\ 
 & {\color[HTML]{0000FF}CPT} & \textbf{10.94} & \textbf{24.71} & 7.64 & 23.85 \\ 
 & {\color[HTML]{6AA84F}SFT} & 17.81 & 30.77 & \textbf{45.03} & \textbf{45.03} \\ 
\multirow{-4}{*}{\textbf{MEDICAL}} & {\color[HTML]{FF0000}CPT+SFT} & 17.41 & 30.50 & \textbf{40.14} & \textbf{40.14} \\ \hline
\multicolumn{6}{c}{\cellcolor[HTML]{CFE2F3}\textit{\textbf{Mistral-7B}}} \\ \hline
 & \textbf{Base} & \textbf{4.51} & \textbf{28.96} & 1.34 & 26.15 \\ 
 & {\color[HTML]{0000FF}CPT} & \textbf{13.84} & \textbf{27.15} & 6.00 & 25.20 \\ 
 & {\color[HTML]{6AA84F}SFT} & \textbf{19.88} & \textbf{32.98} & 7.77 & 27.00 \\ 
\multirow{-4}{*}{\textbf{GENERAL}} & {\color[HTML]{FF0000}CPT+SFT} & \textbf{19.47} & \textbf{32.22} & 9.58 & 27.39 \\ \hline
 & \textbf{Base} & \textbf{21.58} & \textbf{25.51} & 5.96 & 25.10 \\ 
 & {\color[HTML]{0000FF}CPT} & \textbf{28.65} & \textbf{29.69} & 6.90 & 25.51 \\ 
 & {\color[HTML]{6AA84F}SFT} & \textbf{31.64} & \textbf{31.74} & 7.18 & 26.38 \\ 
\multirow{-4}{*}{\textbf{INSTRUCT}} & {\color[HTML]{FF0000}CPT+SFT} & \textbf{29.94} & \textbf{30.11} & 6.75 & 25.21 \\ \hline
 & \textbf{Base} & \textbf{13.52} & \textbf{26.88} & 5.45 & 25.69 \\ 
 & {\color[HTML]{0000FF}CPT} & \textbf{12.10} & \textbf{25.49} & 6.05 & 24.32 \\ 
 & {\color[HTML]{6AA84F}SFT} & \textbf{18.45} & \textbf{31.64} & 7.10 & 26.28 \\ 
\multirow{-4}{*}{\textbf{MEDICAL}} & {\color[HTML]{FF0000}CPT+SFT} & \textbf{19.30} & \textbf{32.33} & 7.43 & 26.90 \\ \hline
\multicolumn{6}{c}{\cellcolor[HTML]{CFE2F3}\textit{\textbf{Llama-7B}}} \\ \hline
 & \textbf{Base} & \textbf{9.38} & \textbf{25.48} & 3.46 & 25.17 \\ 
 & {\color[HTML]{0000FF}CPT} & 6.00 & 25.27 & \textbf{21.14} & \textbf{28.56} \\ 
 & {\color[HTML]{6AA84F}SFT} & 15.74 & 28.61 & \textbf{32.86} & \textbf{32.87} \\ 
\multirow{-4}{*}{\textbf{GENERAL}} & {\color[HTML]{FF0000}CPT+SFT} & 16.97 & 29.77 & \textbf{39.25} & \textbf{39.25} \\ \hline
 & \textbf{Base} & 0.00 & \textbf{24.34} & 0.00 & 23.44 \\ 
 & {\color[HTML]{0000FF}CPT} & 0.00 & \textbf{24.29} & 0.00 & 23.49 \\ 
 & {\color[HTML]{6AA84F}SFT} & 29.52 & 29.58 & \textbf{38.73} & \textbf{38.73} \\ 
\multirow{-4}{*}{\textbf{INSTRUCT}} & {\color[HTML]{FF0000}CPT+SFT} & 0.00 & \textbf{24.54} & 0.00 & 23.45 \\ \hline
 & \textbf{Base} & 0.35 & \textbf{24.19} & \textbf{0.98} & 23.68 \\ 
 & {\color[HTML]{0000FF}CPT} & \textbf{11.97} & \textbf{25.14} & 0.09 & 24.96 \\ 
 & {\color[HTML]{6AA84F}SFT} & 17.38 & 30.27 & \textbf{36.80} & \textbf{36.80} \\ 
\multirow{-4}{*}{\textbf{MEDICAL}} & {\color[HTML]{FF0000}CPT+SFT} & 18.29 & 31.60 & \textbf{36.53} & \textbf{36.53} \\ \hline
\multicolumn{6}{c}{\cellcolor[HTML]{CFE2F3}\textit{\textbf{Llama-13B}}} \\ \hline
 & \textbf{Base} & 11.20 & 25.51 & \textbf{16.55} & \textbf{34.83} \\ 
 & {\color[HTML]{0000FF}CPT} & 10.68 & 26.64 & \textbf{29.79} & \textbf{37.21} \\ 
 & {\color[HTML]{6AA84F}SFT} & 17.30 & 30.18 & \textbf{43.22} & \textbf{43.22} \\ 
\multirow{-4}{*}{\textbf{GENERAL}} & {\color[HTML]{FF0000}CPT+SFT} & 18.27 & 31.41 & \textbf{43.62} & \textbf{43.62} \\ \hline
 & \textbf{Base} & 0.00 & 21.68 & 0.00 & \textbf{23.60} \\ 
 & {\color[HTML]{0000FF}CPT} & 0.00 & 24.57 & 0.00 & \textbf{24.85} \\ 
 & {\color[HTML]{6AA84F}SFT} & 30.04 & 30.10 & \textbf{46.99} & \textbf{46.99} \\ 
\multirow{-4}{*}{\textbf{INSTRUCT}} & {\color[HTML]{FF0000}CPT+SFT} & 31.42 & 31.51 & \textbf{46.57} & \textbf{46.57} \\ \hline
 & \textbf{Base} & 10.31 & 23.97 & \textbf{12.48} & \textbf{24.28} \\ 
 & {\color[HTML]{0000FF}CPT} & \textbf{10.22} & 23.37 & 10.01 & \textbf{30.87} \\ 
 & {\color[HTML]{6AA84F}SFT} & 16.73 & 29.88 & \textbf{37.71} & \textbf{37.71} \\ 
\multirow{-4}{*}{\textbf{MEDICAL}} & {\color[HTML]{FF0000}CPT+SFT} & 18.24 & 31.32 & \textbf{42.73} & \textbf{42.73} \\ \hline
\end{tabular}%
}
\caption{Cross-lingual comparison between native English MCQU benchmarks (MCQU-EN) and their French translations (MCQU-FR), reported as EM (\%). Results are shown for both greedy and constrained decoding. For each row and decoding type, bold values indicate the higher EM between MCQU-FR and MCQU-EN.}
\label{tab:en-vs-fr}
\end{table}

%% file: stat-en-vs-fr.tex
\begin{table*}[]
\centering
\resizebox{0.6\textwidth}{!}{%
\begin{tabular}{clllrrrc}
\hline
\multicolumn{1}{l}{\textbf{Model Type}} & \textbf{Strategy} & \multicolumn{1}{c}{\textbf{Model}} & \multicolumn{1}{c}{\textbf{Decoding Type}} & \multicolumn{1}{c}{\textbf{ci95\_low}} & \multicolumn{1}{c}{\textbf{ci95\_high}} & \multicolumn{1}{c}{\textbf{p\_two\_sided}} & \textbf{Significant} \\ \hline
\multicolumn{8}{c}{\cellcolor[HTML]{CFE2F3}\textbf{Gemma-4B}} \\ \hline
 & \textbf{Base} & gemma-3-4b-pt & greedy & 2,57E-02 & 6,42E-02 & 2,00E-04 & TRUE \\ 
 & \textbf{Base} & gemma-3-4b-pt & constrained & -1,97E-01 & -9,55E-02 & 2,00E-04 & TRUE \\ 
 & {\color[HTML]{0000FF} CPT} & gemma-3-4b-pt-CPT & greedy & -6,57E-02 & -1,71E-02 & 2,20E-03 & TRUE \\ 
 & {\color[HTML]{0000FF} CPT} & gemma-3-4b-pt-CPT & constrained & -1,86E-01 & -1,01E-01 & 2,00E-04 & TRUE \\ 
 & {\color[HTML]{CC0000} CPT+SFT} & gemma-3-4b-CPT-SFT & greedy & -4,01E-01 & -2,18E-01 & 2,00E-04 & TRUE \\  
 & {\color[HTML]{CC0000} CPT+SFT} & gemma-3-4b-CPT-SFT & constrained & -2,48E-01 & -1,14E-01 & 2,00E-04 & TRUE \\  
 & {\color[HTML]{38761D} SFT} & gemma-3-4b-pt-SFT & greedy & -3,98E-01 & -2,19E-01 & 2,00E-04 & TRUE \\  
\multirow{-8}{*}{\textbf{GENERAL}} & {\color[HTML]{38761D} SFT} & gemma-3-4b-pt-SFT & constrained & -2,50E-01 & -1,17E-01 & 2,00E-04 & TRUE \\ \hline
 & \textbf{Base} & gemma-3-4b-it & greedy & -2,31E-01 & -1,36E-01 & 2,00E-04 & TRUE \\  
 & \textbf{Base} & gemma-3-4b-it & constrained & -2,27E-01 & -1,33E-01 & 2,00E-04 & TRUE \\  
 & {\color[HTML]{0000FF} CPT} & gemma-3-4b-it-CPT & greedy & -8,82E-03 & 7,42E-03 & 9,70E-01 & FALSE \\  
 & {\color[HTML]{0000FF} CPT} & gemma-3-4b-it-CPT & constrained & -1,89E-02 & 4,55E-02 & 6,71E-01 & FALSE \\  
 & {\color[HTML]{CC0000} CPT+SFT} & gemma-3-4b-it-CPT-SFT & greedy & -1,14E-01 & -6,32E-02 & 2,00E-04 & TRUE \\  
 & {\color[HTML]{CC0000} CPT+SFT} & gemma-3-4b-it-CPT-SFT & constrained & -1,14E-01 & -6,21E-02 & 2,00E-04 & TRUE \\  
 & {\color[HTML]{38761D} SFT} & gemma-3-4b-it-SFT & greedy & -2,22E-01 & -9,37E-02 & 2,00E-04 & TRUE \\  
\multirow{-8}{*}{\textbf{INSTRUCT}} & {\color[HTML]{38761D} SFT} & gemma-3-4b-it-SFT & constrained & -2,20E-01 & -9,53E-02 & 2,00E-04 & TRUE \\ \hline
 & \textbf{Base} & medgemma-4b-pt & greedy & 6,05E-02 & 2,00E-01 & 2,00E-04 & TRUE \\  
 & \textbf{Base} & medgemma-4b-pt & constrained & -1,05E-01 & -6,63E-03 & 2,94E-02 & TRUE \\  
 & {\color[HTML]{0000FF} CPT} & medgemma-4b-pt-CPT & greedy & -8,53E-03 & 9,14E-02 & 1,71E-01 & FALSE \\  
 & {\color[HTML]{0000FF} CPT} & medgemma-4b-pt-CPT & constrained & -1,45E-02 & 4,11E-02 & 6,29E-01 & FALSE \\  
 & {\color[HTML]{CC0000} CPT+SFT} & medgemma-4b-pt-CPT-SFT & greedy & -2,92E-01 & -1,50E-01 & 2,00E-04 & TRUE \\  
 & {\color[HTML]{CC0000} CPT+SFT} & medgemma-4b-pt-CPT-SFT & constrained & -1,39E-01 & -4,81E-02 & 4,00E-04 & TRUE \\  
 & {\color[HTML]{38761D} SFT} & medgemma-4b-pt-SFT & greedy & -3,43E-01 & -1,90E-01 & 2,00E-04 & TRUE \\  
\multirow{-8}{*}{\textbf{MEDICAL}} & {\color[HTML]{38761D} SFT} & medgemma-4b-pt-SFT & constrained & -1,93E-01 & -8,67E-02 & 2,00E-04 & TRUE \\ \hline
\multicolumn{8}{c}{\cellcolor[HTML]{CFE2F3}\textbf{Mistral-7B}} \\ \hline
 & \textbf{Base} & Mistral-7B-v0.1 & greedy & -1,62E-02 & 1,20E-01 & 6,98E-01 & FALSE \\  
 & \textbf{Base} & Mistral-7B-v0.1 & constrained & 8,39E-02 & 1,70E-01 & 2,00E-04 & TRUE \\  
 & {\color[HTML]{0000FF} CPT} & Mistral-7B-v0.1-CPT & greedy & 3,64E-02 & 1,48E-01 & 2,00E-04 & TRUE \\  
 & {\color[HTML]{0000FF} CPT} & Mistral-7B-v0.1-CPT & constrained & 7,59E-02 & 1,43E-01 & 2,00E-04 & TRUE \\  
 & {\color[HTML]{CC0000} CPT+SFT} & Mistral-7B-v0.1-CPT-SFT & greedy & 5,94E-02 & 1,61E-01 & 2,00E-04 & TRUE \\  
 & {\color[HTML]{CC0000} CPT+SFT} & Mistral-7B-v0.1-CPT-SFT & constrained & 4,90E-02 & 1,04E-01 & 2,00E-04 & TRUE \\  
 & {\color[HTML]{38761D} SFT} & Mistral-7B-v0.1-SFT & greedy & 7,77E-02 & 1,92E-01 & 2,00E-04 & TRUE \\  
\multirow{-8}{*}{\textbf{GENERAL}} & {\color[HTML]{38761D} SFT} & Mistral-7B-v0.1-SFT & constrained & 5,74E-02 & 1,29E-01 & 2,00E-04 & TRUE \\ \hline
 & \textbf{Base} & Mistral-7B-Instruct-v0.1 & greedy & 1,30E-01 & 1,83E-01 & 2,00E-04 & TRUE \\  
 & \textbf{Base} & Mistral-7B-Instruct-v0.1 & constrained & 1,21E-01 & 1,60E-01 & 2,00E-04 & TRUE \\  
 & {\color[HTML]{0000FF} CPT} & Mistral-7B-Instruct-v0.1-CPT & greedy & 1,74E-01 & 2,62E-01 & 2,00E-04 & TRUE \\  
 & {\color[HTML]{0000FF} CPT} & Mistral-7B-Instruct-v0.1-CPT & constrained & 9,71E-02 & 1,73E-01 & 2,00E-04 & TRUE \\  
 & {\color[HTML]{CC0000} CPT+SFT} & Mistral-7B-Instruct-v0.1-CPT-SFT & greedy & 1,83E-01 & 2,94E-01 & 2,00E-04 & TRUE \\  
 & {\color[HTML]{CC0000} CPT+SFT} & Mistral-7B-Instruct-v0.1-CPT-SFT & constrained & 6,43E-02 & 1,42E-01 & 2,00E-04 & TRUE \\  
 & {\color[HTML]{38761D} SFT} & Mistral-7B-Instruct-v0.1-SFT & greedy & 1,92E-01 & 3,03E-01 & 2,00E-04 & TRUE \\  
\multirow{-8}{*}{\textbf{INSTRUCT}} & {\color[HTML]{38761D} SFT} & Mistral-7B-Instruct-v0.1-SFT & constrained & 7,81E-02 & 1,56E-01 & 2,00E-04 & TRUE \\ \hline
 & \textbf{Base} & BioMistral-7B & greedy & 4,88E-02 & 1,29E-01 & 2,00E-04 & TRUE \\  
 & \textbf{Base} & BioMistral-7B & constrained & 1,21E-01 & 1,63E-01 & 2,00E-04 & TRUE \\  
 & {\color[HTML]{0000FF} CPT} & BioMistral-7B-CPT & greedy & 3,47E-02 & 9,30E-02 & 2,00E-04 & TRUE \\  
 & {\color[HTML]{0000FF} CPT} & BioMistral-7B-CPT & constrained & 9,18E-02 & 1,20E-01 & 2,00E-04 & TRUE \\  
 & {\color[HTML]{CC0000} CPT+SFT} & BioMistral-7B-CPT-SFT & greedy & 6,82E-02 & 2,06E-01 & 2,00E-04 & TRUE \\  
 & {\color[HTML]{CC0000} CPT+SFT} & BioMistral-7B-CPT-SFT & constrained & 8,89E-02 & 1,88E-01 & 2,00E-04 & TRUE \\  
 & {\color[HTML]{38761D} SFT} & BioMistral-7B-SFT & greedy & 6,71E-02 & 1,94E-01 & 2,00E-04 & TRUE \\  
\multirow{-8}{*}{\textbf{MEDICAL}} & {\color[HTML]{38761D} SFT} & BioMistral-7B-SFT & constrained & 8,33E-02 & 1,68E-01 & 2,00E-04 & TRUE \\ \hline
\multicolumn{8}{c}{\cellcolor[HTML]{CFE2F3}\textbf{Llama-7B}} \\ \hline
 & \textbf{Base} & Llama-2-7b-hf & greedy & 1,22E-02 & 1,24E-01 & 6,60E-03 & TRUE \\  
 & \textbf{Base} & Llama-2-7b-hf & constrained & -2,49E-02 & 3,53E-02 & 8,46E-01 & FALSE \\  
 & {\color[HTML]{0000FF} CPT} & Llama-2-7b-hf-CPT & greedy & -2,23E-01 & -6,14E-02 & 4,00E-03 & TRUE \\  
 & {\color[HTML]{0000FF} CPT} & Llama-2-7b-hf-CPT & constrained & -2,27E-02 & 1,19E-01 & 3,12E-01 & FALSE \\  
 & {\color[HTML]{CC0000} CPT+SFT} & LLama-2-7b-hf-CPT-SFT & greedy & -2,93E-01 & -1,33E-01 & 2,00E-04 & TRUE \\  
 & {\color[HTML]{CC0000} CPT+SFT} & LLama-2-7b-hf-CPT-SFT & constrained & -1,49E-01 & -3,08E-02 & 4,20E-03 & TRUE \\  
 & {\color[HTML]{38761D} SFT} & LLama-2-7b-hf-SFT & greedy & -2,27E-01 & -9,34E-02 & 4,00E-04 & TRUE \\  
\multirow{-8}{*}{\textbf{GENERAL}} & {\color[HTML]{38761D} SFT} & LLama-2-7b-hf-SFT & constrained & -8,38E-02 & 7,10E-03 & 8,72E-02 & FALSE \\ \hline
 & \textbf{Base} & Llama-2-7b-chat-hf & greedy & 0,00E+00 & 0,00E+00 & 1,00E+00 & FALSE \\  
 & \textbf{Base} & Llama-2-7b-chat-hf & constrained & 1,84E-01 & 3,09E-01 & 2,00E-04 & TRUE \\  
 & {\color[HTML]{0000FF} CPT} & Llama-2-7b-chat-hf-CPT & greedy & 0,00E+00 & 1,02E-04 & 7,23E-01 & FALSE \\  
 & {\color[HTML]{0000FF} CPT} & Llama-2-7b-chat-hf-CPT & constrained & 5,27E-02 & 2,03E-01 & 2,00E-04 & TRUE \\  
 & {\color[HTML]{CC0000} CPT+SFT} & Llama-2-7b-chat-hf-CPT-SFT & greedy & 0,00E+00 & 6,80E-05 & 7,01E-01 & FALSE \\  
 & {\color[HTML]{CC0000} CPT+SFT} & Llama-2-7b-chat-hf-CPT-SFT & constrained & 8,66E-02 & 1,40E-01 & 2,00E-04 & TRUE \\  
 & {\color[HTML]{38761D} SFT} & Llama-2-7b-chat-hf-SFT & greedy & -1,32E-01 & -3,94E-02 & 1,20E-03 & TRUE \\  
\multirow{-8}{*}{\textbf{INSTRUCT}} & {\color[HTML]{38761D} SFT} & Llama-2-7b-chat-hf-SFT & constrained & -1,31E-01 & -3,80E-02 & 1,20E-03 & TRUE \\ \hline
 & \textbf{Base} & meditron-7b & greedy & -1,03E-02 & -2,51E-03 & 1,00E-03 & TRUE \\  
 & \textbf{Base} & meditron-7b & constrained & -1,42E-02 & 3,19E-02 & 5,00E-01 & FALSE \\  
 & {\color[HTML]{0000FF} CPT} & meditron-7b-CPT & greedy & -6,55E-03 & 1,08E-01 & 3,98E-01 & FALSE \\  
 & {\color[HTML]{0000FF} CPT} & meditron-7b-CPT & constrained & 2,15E-03 & 6,87E-02 & 3,58E-02 & TRUE \\  
 & {\color[HTML]{CC0000} CPT+SFT} & meditron-7b-CPT-SFT & greedy & -2,61E-01 & -7,38E-02 & 1,80E-03 & TRUE \\  
 & {\color[HTML]{CC0000} CPT+SFT} & meditron-7b-CPT-SFT & constrained & -1,05E-01 & 2,86E-02 & 1,86E-01 & FALSE \\  
 & {\color[HTML]{38761D} SFT} & meditron-7b-SFT & greedy & -2,64E-01 & -1,02E-01 & 2,00E-04 & TRUE \\  
\multirow{-8}{*}{\textbf{MEDICAL}} & {\color[HTML]{38761D} SFT} & meditron-7b-SFT & constrained & -1,15E-01 & -3,59E-03 & 4,00E-02 & TRUE \\ \hline
\multicolumn{8}{c}{\cellcolor[HTML]{CFE2F3}\textbf{Llama-13B}} \\ \hline
 & \textbf{Base} & Llama-2-13b-hf & greedy & -1,31E-01 & 5,40E-02 & 2,74E-01 & FALSE \\  
 & \textbf{Base} & Llama-2-13b-hf & constrained & -1,35E-01 & -4,79E-02 & 4,00E-04 & TRUE \\  
 & {\color[HTML]{0000FF} CPT} & Llama-2-13b-hf-CPT & greedy & -2,65E-01 & -1,00E-01 & 2,00E-04 & TRUE \\  
 & {\color[HTML]{0000FF} CPT} & Llama-2-13b-hf-CPT & constrained & -1,09E-01 & 2,30E-02 & 1,58E-01 & FALSE \\  
 & {\color[HTML]{CC0000} CPT+SFT} & Llama-2-13b-hf-CPT-SFT & greedy & -3,31E-01 & -1,52E-01 & 2,00E-04 & TRUE \\  
 & {\color[HTML]{CC0000} CPT+SFT} & Llama-2-13b-hf-CPT-SFT & constrained & -1,82E-01 & -5,08E-02 & 1,60E-03 & TRUE \\  
 & {\color[HTML]{38761D} SFT} & Llama-2-13b-hf-SFT & greedy & -3,30E-01 & -1,68E-01 & 2,00E-04 & TRUE \\  
\multirow{-8}{*}{\textbf{GENERAL}} & {\color[HTML]{38761D} SFT} & Llama-2-13b-hf-SFT & constrained & -1,87E-01 & -6,61E-02 & 2,00E-04 & TRUE \\ \hline
 & \textbf{Base} & Llama-2-13b-chat-hf & greedy & 0,00E+00 & 0,00E+00 & 1,00E+00 & FALSE \\  
 & \textbf{Base} & Llama-2-13b-chat-hf & constrained & -4,79E-02 & 2,61E-03 & 1,06E-01 & FALSE \\  
 & {\color[HTML]{0000FF} CPT} & Llama-2-13b-chat-hf-CPT & greedy & 0,00E+00 & 0,00E+00 & 1,00E+00 & FALSE \\  
 & {\color[HTML]{0000FF} CPT} & Llama-2-13b-chat-hf-CPT & constrained & -2,29E-02 & 2,51E-02 & 7,65E-01 & FALSE \\  
 & {\color[HTML]{CC0000} CPT+SFT} & Llama-2-13b-chat-hf-CPT-SFT & greedy & -2,09E-01 & -7,94E-02 & 4,00E-04 & TRUE \\  
 & {\color[HTML]{CC0000} CPT+SFT} & Llama-2-13b-chat-hf-CPT-SFT & constrained & -2,06E-01 & -7,95E-02 & 2,00E-04 & TRUE \\  
 & {\color[HTML]{38761D} SFT} & Llama-2-13b-chat-hf-SFT & greedy & -2,21E-01 & -1,15E-01 & 2,00E-04 & TRUE \\  
\multirow{-8}{*}{\textbf{INSTRUCT}} & {\color[HTML]{38761D} SFT} & Llama-2-13b-chat-hf-SFT & constrained & -2,20E-01 & -1,16E-01 & 2,00E-04 & TRUE \\ \hline
 & \textbf{Base} & MedLLaMA\_13B & greedy & -8,31E-02 & 4,03E-02 & 4,90E-01 & FALSE \\  
 & \textbf{Base} & MedLLaMA\_13B & constrained & -2,46E-02 & 1,98E-02 & 7,65E-01 & FALSE \\  
 & {\color[HTML]{0000FF} CPT} & MedLLaMA-13B-CPT & greedy & -3,43E-02 & 3,73E-02 & 8,95E-01 & FALSE \\  
 & {\color[HTML]{0000FF} CPT} & MedLLaMA-13B-CPT & constrained & -1,10E-01 & -4,51E-02 & 2,00E-04 & TRUE \\  
 & {\color[HTML]{CC0000} CPT+SFT} & MedLLaMA-13B-CPT-SFT & greedy & -3,14E-01 & -1,60E-01 & 2,00E-04 & TRUE \\  
 & {\color[HTML]{CC0000} CPT+SFT} & MedLLaMA-13B-CPT-SFT & constrained & -1,68E-01 & -5,56E-02 & 1,40E-03 & TRUE \\  
 & {\color[HTML]{38761D} SFT} & MedLLaMA-13B-SFT & greedy & -2,76E-01 & -1,29E-01 & 2,00E-04 & TRUE \\  
\multirow{-8}{*}{\textbf{MEDICAL}} & {\color[HTML]{38761D} SFT} & MedLLaMA-13B-SFT & constrained & -1,27E-01 & -2,23E-02 & 7,20E-03 & TRUE \\ \hline
\end{tabular}%
}
\caption{Paired significance testing between MCQU-EN and MCQU-FR for each model configuration (model, strategy, and decoding type). Reported values are the 95\% confidence interval of the mean EM difference and the corresponding two-sided two-sided $p$-value; \texttt{Significant} indicates whether the difference is statistically significant. We define the difference as $(\mathrm{FR}-\mathrm{EN})$, such that positive values indicate higher performance in French.}
\label{tab:stat-en-vs-fr}
\end{table*}

%% file: gpu-conso.tex
\begin{table*}[]
\centering
\resizebox{\textwidth}{!}{%
\begin{tabular}{ccccccccccc}
\hline
\textbf{Model Size} &  \textbf{Strategy} & \textbf{\begin{tabular}[c]{@{}c@{}}Dataset size\\ (KB)\end{tabular}} & \textbf{Epochs} & \textbf{Batch-size} & \textbf{Type of GPU} & \textbf{\begin{tabular}[c]{@{}c@{}}Memory per\\ GPU (GB)\end{tabular}} & \textbf{\begin{tabular}[c]{@{}c@{}}Number of\\ GPUs\end{tabular}} & \textbf{\begin{tabular}[c]{@{}c@{}}Training\\ time (hours)\end{tabular}} & \textbf{\begin{tabular}[c]{@{}c@{}}Emissions\\ (g CO2e)\end{tabular}} & \textbf{Cost (USD)} \\ \hline
 & {\color[HTML]{0000FF} \textbf{CPT}} & 4\,000\,000 & 3 & 4 & NVIDIA A100 & 80 & 24 & 80 & 49\,344 & 1\,824.62 \\  
\multirow{-2}{*}{\textbf{4B}} & {\color[HTML]{6AA84F} \textbf{SFT}} & 369 & 10 & 4 & NVIDIA H100 & 80 & 3 & 146 & 11\,256.6 & 832.48 \\ \hline
 & {\color[HTML]{0000FF} \textbf{CPT}} & 4\,000\,000 & 3 & 2 & NVIDIA A100 & 80 & 32 & 40 & 32\,896 & 1\,216.42 \\  
\multirow{-2}{*}{\textbf{7B}} & {\color[HTML]{6AA84F} \textbf{SFT}} & 369 & 10 & 4 & NVIDIA H100 & 80 & 1 & 190 & 4\,883 & 361.12 \\ \hline
 & {\color[HTML]{0000FF} \textbf{CPT}} & 4\,000\,000 & 3 & 2 & NVIDIA H100 & 80 & 32 & 100 & 82\,240 & 6\,082.08 \\  
\multirow{-2}{*}{\textbf{13B}} & {\color[HTML]{6AA84F} \textbf{SFT}} & 369 & 10 & 4 & NVIDIA H100 & 80 & 6 & 122 & 18\,812.4 & 1\,391.27 \\ \hline
\end{tabular}%
}
\caption{Summary of computational resources and environmental impact for different adaptation strategies, aggregated by model size. Reported values correspond to a representative training configuration per strategy. CPT+SFT costs are obtained by summing CPT and SFT.}

\label{tab:conso}
\end{table*}